\documentclass{article}
\usepackage{iclr2025_conference,times}
\iclrfinalcopy %for arxiv

\usepackage[utf8]{inputenc} % allow utf-8 input
\usepackage[T1]{fontenc}    % use 8-bit T1 fonts
\usepackage{hyperref}       % hyperlinks
\usepackage{url}            % simple URL typesetting
\usepackage{booktabs}       % professional-quality tables
\usepackage{amsfonts}       % blackboard math symbols
\usepackage{nicefrac}       % compact symbols for 1/2, etc.
\usepackage{microtype}      % microtypography
\usepackage{xcolor}         % colors

\usepackage{etoolbox}
\providetoggle{todos}
\providetoggle{long}
\settoggle{todos}{false}
\settoggle{long}{false}
\providetoggle{arxiv}
\settoggle{arxiv}{true}
\providetoggle{iclr}
\settoggle{iclr}{false}

\providetoggle{unitcircle}
\settoggle{unitcircle}{false}
\usepackage{tikz}
\usetikzlibrary{patterns}
\usetikzlibrary{topaths, shapes ,arrows ,positioning}
\tikzset{bend angle=15,
}
\usepackage{pgfplots}
\newtoks\mycoords
\usepackage{xparse,fp,xfp}
\usetikzlibrary{decorations.pathreplacing}
\usepackage{microtype}
\usepackage{graphicx}
\usepackage{booktabs} % for professional tables
\usepackage{amsmath}
\usepackage{amsfonts}
\usepackage{lastpage}
\usepackage{multirow}
\usepackage{tabu}
\usepackage{lipsum}
\usepackage{clrscode}
\usepackage{caption}
\usepackage{subfig}
\usepackage{float}
\usepackage{appendix}
\usepackage{mathrsfs}
\usepackage{indentfirst}

\newcommand{\lasso}{Lasso} %vs lasso vs LASSO
\newcommand{\params}{\theta} %full parameters
\newcommand{\paramspace}{\Theta} %\params \in \paramspace
\newcommand{\regparams}{\theta_w} %regularized parameters
\newcommand{\nn}[2]{f_{#1}\left({#2};\params\right)} 
\newcommand{\Wul}[2]{\mathbf{W}^{\left(#1,#2\right)}} % #1 is unit u, #2 is layer l
\newcommand{\bul}[2]{\mathbf{b}^{\left(#1,#2\right)}} % #1 is unit u, #2 is layer l
\newcommand{\sWul}[2]{w^{\left(#1,#2\right)}} % #1 is unit u, #2 is layer l. small w (lowercase)
\newcommand{\sbul}[2]{b^{\left(#1,#2\right)}} % 
\newcommand{\lossfunc}{\mathcal{L}_{\mathbf{y}}}
\newcommand{\regfunc}{r}

\newcommand{\Xul}[2]{\mathbf{X}^{\left(#1,#2\right)}} % #1 is unit u, #2 is layer l
\newcommand{\extbias}{\xi}

\newcommand{\Amat}{\mathbf{A}}
\newcommand{\nc}{non-convex}
\newcommand{\nd}[1]{{#1}-D}
\newcommand{\deepnarrow}{deep narrow}
\newcommand{\deepReLU}[1]{\deepnarrow{} network}

\newcommand{\kink}{breakplane}
\newcommand{\kinkpoint}{breakpoint}

\newcommand{\reflect}[2]{R_{\left({#1},{#2}\right)}}

\newcommand{\adam}{Adam}
\newcommand{\wl}[1]{\mathbf{W}^{\left({#1}\right)}}
\newcommand{\xl}[1]{\mathbf{X}^{\left({#1}\right)}}
\newcommand{\bl}[1]{\mathbf{b}^{\left({#1}\right)}}
\newcommand{\sbl}[1]{b^{\left({#1}\right)}}

\newcommand\scalemath[2]{\scalebox{#1}{\mbox{\ensuremath{\displaystyle #2}}}} %https://tex.stackexchange.com/questions/43065/matrices-too-big-to-fit-the-width-of-a-page

\usepackage{amssymb} %in jmlr2e.sty
\usepackage{mathtools}
\usepackage{amsthm}
\usepackage[capitalize,noabbrev]{cleveref}
\usepackage{bbm}

\newtheorem{theorem}{Theorem}[section]

\newtheorem{lemma}[theorem]{Lemma}

\newtheorem{definition}[theorem]{Definition}

\newtheorem{remark}[theorem]{Remark}
{% begin code
\par\noindent
  \textbf{#1 (Informal)} \begin{itshape}%
}%
{\end{itshape}\par\vspace{\baselineskip}}% end code
\newcommand{\reflectnn}{deep narrow network}

\newcommand{\xprime}{\mathbf{x}'_{j_1}}
\newcommand{\lratiopenalty}{r}
\newcommand{\featurefunc}{f_j}
\newcommand{\sublibrary}{data feature sub-library}
\newcommand{\regplane}{\mathcal{H}}
\newcommand{\reflectplane}{\mathcal{H}_R}
\newcommand{\avgplane}{\mathcal{H}_A}
\usepackage{comment}

\usepackage[textsize=small]{todonotes}

\author{
  Emi Zeger and Mert Pilanci \\
 Department of Electrical Engineering \\
  Stanford University \\
}

\title{
Black Boxes and Looking Glasses: Multilevel Symmetries, Reflection Planes, and Convex\\ Optimization in Deep Networks}
\begin{document}

\maketitle

\begin{abstract}
We show that training deep neural networks (DNNs) with absolute value activation and arbitrary input dimension can be formulated as equivalent convex Lasso problems with novel features expressed using geometric algebra. This formulation reveals geometric structures encoding symmetry in neural networks. Using the equivalent Lasso form of DNNs, we formally prove a fundamental distinction between deep and shallow networks: deep networks inherently favor symmetric structures in their fitted functions, with greater depth enabling multilevel symmetries, i.e., symmetries within symmetries.  Moreover, Lasso features represent distances to hyperplanes that are reflected across training points. These reflection hyperplanes are spanned by training data and are orthogonal to optimal weight vectors. Numerical experiments support theory and demonstrate theoretically predicted features when training networks using embeddings generated by Large Language Models.
\end{abstract}

%9 pages including appendix, excluding references and acknowledgements
\section{Introduction}
Recent advancements have demonstrated that deep neural networks are powerful models that can perform tasks including natural language processing, synthetic data and image generation, classification, and regression. However, research literature still lacks in intuitively understanding why deep networks are so powerful: what they "look for" in data, or in other words, how each layer extracts features.
%%%from simod paper
%Convex analysis of neural networks was developed in several prior works. As an example, infinite-width neural networks enable the convexification of the overall model \citep{bach2017breaking,bengio2005convex,fang2019convex}. However, due to the infinite-width assumption, these results do not reflect finite-width neural networks in practice.
%%%%%%%
%
%The dictionary matrix size is $N \times 2N$ for $N$ one-dimensional samples in a $2$-layer neural net with a variety of activation functions. We also give dictionaries for sign activation in deeper networks.
 %
 We are interested in the following question:
\begin{quote}
\textbf{\textit{Is there a fundamental difference in the nature of functions learned by deep networks, as opposed to shallow networks?}}
\end{quote}
We answer this question by transforming \nc{} training problems into convex formulations and analyzing their structure. \textbf{We show that deep nets favor symmetric structures, with greater depth representing multilevel symmetries encoded via reflections of reflections of data points.}

A \textit{neural network} of depth $L$
%and $m_l$ neurons in each layer 
is a parameterized function $ \nn{L}{ \cdot \:} : \mathbb{R}^{1\times d} \to \mathbb{R}$ of the form % with input $\mathbf{x} \in \mathbb{R}^d$:
\begin{equation}\label{eq:nn_nonrecursive}
    \nn{L}{\mathbf{x}} {=} \sum_{i=1}^{m_L}\sigma\left({\cdots} \left(\sigma\left(\sigma\left(\mathbf{x}\Wul{i}{1}+\bul{i}{1}\right)\Wul{i}{2} {+} \bul{i}{2}\right)\cdots \right)\Wul{i}{L{-}1} {+} \bul{i}{L{-}1} \right)\alpha_i {+} \extbias.
\end{equation}
The trainable parameters are the outermost weights $\alpha_i {\in} \mathbb{R}$, an external bias $\extbias {\in} \mathbb{R}$; and the inner weights $ \Wul{i}{l} {\in} \mathbb{R}^{m_{l{-}1}'\times m_l'}$ and inner biases $\bul{i}{l} {\in} \mathbb{R}^{1 \times m_l'}$ for each layer $l$. There are $m_l {=} m_L m_l'$ neurons in each layer.  The \textit{activation} function is
$\sigma: \mathbb{R} {\to} \mathbb{R}$. Note that $m_{0}'{=}d$, which is the input data dimension, and $m_{L{-}1}'{=}1$. 
Let $N$ be the number of training samples and $\mathbf{X} {\in} \mathbb{R}^{N \times d}$ be the \textit{training data matrix}, where each row of $\mathbf{X}$ is a training sample $\mathbf{x}_n {\in} \mathbb{R}^{1 {\times} d}$. The neural network extends to matrix inputs row-wise. 
We consider regression problems, although the results can also be applied to classification problems. The \textit{target} or \textit{label} vector is $\mathbf{y} {\in} \mathbb{R}^N$.
The neural network \textit{training problem} is
\begin{equation}\label{eq:trainingprob_explicit}
    \min_{\params \in \paramspace} \frac{1}{2}\left \| \nn{L}{\mathbf{X}}-\mathbf{y} \right\|_2^2 + \frac{\beta}{L}\sum_{i=1}^{m} \left( \left| \alpha_i \right|^L + \sum_{l=1}^{L-1} \big\|\Wul{i}{l}\big\|^L_1 \right)  
\end{equation}
where $\params {=} \left\{\Wul{i}{l}, \bul{i}{l}, \alpha_i, \extbias : i {\in} [m_L], l {\in} [L{-}1] \right\}$ denotes the set of trainable parameters and $\paramspace$ is the parameter space. %A $l_1$ regularization penalizes the weights of the network, and 
The \textit{regularization parameter} is $\beta{>}0$, which is fixed. The $l_1$ regularization penalty in \eqref{eq:trainingprob_explicit} is used for analytical tractability and it is known that $l_1$ regularization can approximate $l_2$ regularization \cite{Pilanci:2023}. Our results generalize to other convex loss functions in addition to $l_2$ loss, and also applies to the case when $\beta {\to} 0$ by using the minimum norm problem \eqref{rm_gen_lasso_nobeta}. Neural networks \eqref{eq:nn_nonrecursive} and their training problems \eqref{eq:trainingprob_explicit} are \nc{}, which can make understanding and training even shallow neural networks challenging. We provide a foundational analysis that isolates the effect of neural network depth by focusing on the \reflectnn{}. The \emph{\reflectnn{}} is a special case of \eqref{eq:nn_nonrecursive} with absolute value activation $\sigma(x){=}|x|$, arbitrary depth $L$, and a width $m_1{=}{\cdots}{=}m_L{=}m$ that is arbitrary but constant across layers:
%name this
\begin{equation}\label{eq:deepnarrowabsvalnetwork}
    \nn{L}{\mathbf{x}} = \sum_{i=1}^{m}\left | \rule{0cm}{0.8cm} \cdots \Bigg |  \bigg| \Big|\mathbf{x}\Wul{i}{1}+\sbul{i}{1}\Big|\sWul{i}{2} + \sbul{i}{2}\bigg |\cdots \Bigg | \sWul{i}{L-1} + \sbul{i}{L-1} \right| \alpha_i + \extbias.
\end{equation}
Except for $\Wul{i}{1} {\in} \mathbb{R}^{d}$, all trainable parameters in the \reflectnn{} are scalars, as denoted by the lowercase, non-boldface text in \eqref{eq:deepnarrowabsvalnetwork}. The total number of neurons across each of the $L$ layers is $m$. We show that increasing the number of layers in a \reflectnn{} even with the constant width $m_1{=}{\cdots}{=}m_{L{-}1}{=}m$ enables the network to learn richer features.
%\subsection{Why use absolute value activations?}

The \reflectnn{} uses absolute value activation instead of the more standard ReLU. The absolute value activation is a symmetric function and this allows for tractable, simpler formulations of neural network features.
The absolute value activation is equivalent to the ReLU activation in $2$-layer networks with skip connections \citep{1DNN}. 
Moreover, neural networks with absolute value activation have been shown to perform comparably to ReLU and have advantages such as no exploding or vanishing gradients, making them suitable for deep networks \citep{Berngardt:2023} and outperforming other activations in computer vision \citep{Perez:2023}. %to have comparable performance as those with the more traditionally used ReLU activation, and in some cases they outperform ReLU (\textcolor{red}{CITE}). 
Our results suggest that absolute value activations are especially primed to capture reflections and symmetries in the data, leading to their representation power. Extensions to ReLU networks are discussed in \Cref{ReLUcomparison}.
%We discuss extensions to other activations and wider, more general architectures in \Cref{sec:extensions}. 
%We show that increasing the number of layers even without making the network wider causes reflection features. 

Recently, \citep{pilanci2023complexity} described neural network features using geometric algebra, also known as Clifford algebra. Geometric algebra is an extension of vector algebra that centers around geometric operations such as translation and reflections \citep{perwass:2009, Berger:2009, Chisholm:2012}. Geometric algebra provides a unified framework for expressing laws of physics \citep{doran:2003, hestenes:2003} and has applications in computer graphics \citep{vince:2008}.
%While many geometric notions that this work uses, such as generalized cross products and their relationship to volume, can be expressed using traditional vector notation without geometric algebra, 
Geometric algebra generalizes vector operations such as cross products, inner products, and outer products from a geometric perspective \cite{Berger:2009}. The use of geometric algebra, which is well-studied, also opens doors to potentially deep connections or applications to physics. In this work, we extend \citep{pilanci2023complexity} and leverage geometric algebra to examine the connection between deep neural networks and \lasso{} problems.
A \textit{\lasso{} problem} is a convex optimization problem of the form
\begin{equation}\label{eq:lasso}
\min_{\mathbf{z}, \extbias}  \frac{1}{2} \| \Amat{} \mathbf{z} + \extbias \mathbf{1} - \mathbf{y} \|^2_2 + \beta \|\mathbf{z}\|_1
\end{equation}
where $\mathbf{A}$ is a \textit{dictionary matrix}, and its columns $\mathbf{A}_i {\in} \mathbb{R}^N$ are \textit{feature vectors}. We will show that the \lasso{} problem and the \reflectnn{} training problem are equivalent (\Cref{lemma:deep}). The \lasso{} problem, its solution set, and solvers for it are well-studied \citep{efron2004least,Tibshirani:1996,  tibshirani2013unique}. Thus the \lasso{} model sheds new light on the training and geometric interpretability of neural networks. 
For training, the convexity of the \lasso{} problem implies that all of its stationary points are optimal, which prevents convergence to sub-optimal local minima that can occur with the \nc{} training problem \eqref{eq:trainingprob_explicit}. Moreover, there are efficient and interpretable algorithms such as Least Angle Regression (LARS) for solving the \lasso{} problem \citep{efron2004least}. 

In terms of geometric interpretability, the $l_1$ regularization in the \lasso{} problem \eqref{eq:lasso} favors a sparse solution for $\mathbf{z}$. The dictionary columns $\mathbf{A}_i$ where $z_i {\neq} 0$ constitute a sparse set of feature vectors that we will show correspond to \textit{feature} functions defining an optimal neural network. We will also show that the features can be expressed using volumes and wedge products defined in geometric algebra, and that they measure distances to \emph{reflection planes} spanned by subsets of training data. For $\mathbf{a},\mathbf{b} {\in} \mathbb{R}^d$, the \emph{reflection of $\mathbf{a}$ about $\mathbf{b}$} is $\reflect{\mathbf{a}}{\mathbf{b}} {=} 2\mathbf{b}{-}\mathbf{a}$. With respect to $\mathbf{b}$, $\mathbf{a}$ and $\reflect{\mathbf{a}}{\mathbf{b}}$ have a \emph{single-level symmetry}. Higher-order reflections can be defined to create deeper levels of symmetry (\cref{section:deeper}). Neural networks encode reflection planes that are based at training data and their reflections about each other, and which are orthogonal to neuron weights.

\textbf{Significance:} The \lasso{} model elucidates a quantitative difference between shallow and deep networks. Deep networks exploit multilevel symmetries through reflections that capture increasingly more complex relations between data as networks deepen. Every layer in a network creates more layers of reflective symmetry (see  \Cref{fig:absvalfeat}).
%\begin{definition}[reflection of vectors]\label{def:reflection} %move this earlie
%Neuron weights are also expressed using geometric algebra, and they are orthogonal to the reflection planes.
%However, there is a paucity of research on \nc{} neural networks. While there has been prior work on infinite-width networks, the literature lacks for finite-width architectures. In this paper, we focus on a simple, finite-width network, a \reflectnn, to introduce an analytical framework with potential to extend more general networks. 
%
%The \lasso{} problem's convexity makes training avoid local minima, which also reduces the need for hyperparameter tuning. In addition the \lasso{} model makes the problem interpretable. The \lasso{} dictionary reveals geometric symmetries in the network, and its elements as well as optimal weights can be expressed using wedge products defined within Clifford's Algebra.
%
%\subsection{Why use Geometric Algebra?}

%would it help to also write reflections using geom alg if it does apply?

\subsection{Related Work}\label{section:relatedwork}
Convexifying neural networks has been studied in \citep{bach2017breaking,bengio2005convex,fang2019convex}. However, these works assume that the network has infinite width, limiting their practical application.
Recently, the training problem for finite-width, shallow ReLU networks has been shown to be equivalent to a convex reformulation with group norm regularization \citep{ergen2020aistats, ergen2021convex,pilanci2020neural}. 

{The use of geometric algebra for  neuron operations is explored in \citep{brehmer:2023, Ruhe:2023}. In contrast, our work uses conventional, real-valued neuron weights but interprets the features using geometric algebra.}
\cite{pilanci2023complexity} uses geometric algebra to formulate $2$ and $3$-layer ReLU neural network training problems as convex \lasso{} problems with features that represent high-dimensional volumes of parallelotopes spanned by training data. From the \lasso{} problem, an optimal network can be reconstructed, and its weights are orthogonal to the training data. The input data is of arbitrary dimension.

\cite{1DNN} shows that for \nd{1} input data and a variety of activation functions and arbitrarily deep networks, the training problem can be written as a \lasso{} problem. Moreover, for absolute value activations, simple networks with \nd{1} input exhibit features that represent reflections of training data, demonstrating that neural networks learn geometric features.

In this paper, we extend \cite{Pilanci:2023} and \cite{1DNN} to prove that neural networks with absolute value activation and arbitrary input dimension and depth can be reformulated as equivalent \lasso{} problems with reflection features, using geometric algebra. The \lasso{} model suggests that larger models learn features related to symmetries in the data. This work differs from \cite{Pilanci:2023} by considering absolute value instead of ReLU activation, which introduces new reflection features, and differs from \cite{1DNN} by considering arbitrary d-dimensional input instead of \nd{1} data. This work provides foundational analysis of networks with arbitrary dimensional data as functions that represent higher levels of reflections as networks deepen.

Unlike prior work, this work also reveals a \emph{sparsity factor}, which is the ratio of $l_2$ and $l_1$ norms of a vector, in a network's equivalent \lasso{} model, which encourages parsimonious solutions.
%shows how the $l_1$ regularization in the training problem \eqref{eq:trainingprob_explicit} facilitates parsimonious networks, both through a \lasso{} problem and through a "sparsity factor" in the features. 
This sparsity factor has also been studied in \citep{Yin:2014,Xu:2021}. Additionally, this paper proves that the reflection complexity grows linearly with the number of layers in a \reflectnn{}.
%Geometric algebra can be used to train and polish neural networks computationally efficiently, which can be adapted to our work \cite{Wang:2024, pilanci2023complexity}.
The reflection features suggest that networks such as Large Language Models (LLMs) may learn analogous structures and concepts in language.
The presence of geometric structures in LLMs has been studied in \citep{Park:2024}. Neural networks trained using LLM embeddings appear to learn the features predicted in \lasso{} models (\Cref{section:numresults}). 

\subsection{Contributions}
We show the following:
\begin{itemize}
    \item Training \reflectnn{}s of arbitrary depth is equivalent to solving convex \lasso{} problems with a finite set of features representing multilevel symmetries via reflections of increasing complexity (\Cref{lemma:deep}, \Cref{absvalsublib}).
    \item There are explicit \lasso{} dictionaries for $3$-layer networks, and reconstructions of optimal networks from corresponding \lasso{} models. The reconstructed, optimal first-layer weight is orthogonal to a subset of augmented training data (\Cref{thrm:3layersimpler}). 
    \item Networks with $3$ layers can be described in terms of geometric algebra. They measure volumes of paralleletopes spanned by training data and distances to reflection planes (\Cref{thrm:3layersimpler}).
    %parallel hyperplanes reflected over data points
    \item Neural networks trained using LLM text vector embeddings learn similar features as \lasso{} models (\Cref{section:numresults}, \Cref{app:num_results}).
\end{itemize}

\subsection{Notation}\label{section:notation}
Let $e_i {\in} \mathbb{R}^d$ denote the $i^\text{th}$ canonical basis vector. 
%Let $\mathcal{E} {=} \{e_1, {\cdots}, e_d \}$. 
Let $[d]{=}\{1,{\cdots},d\}$. The number of non-zero elements in a vector $\mathbf{z}$ is $\|\mathbf{z}\|_0$. Neural network inputs, parameters, and outputs are all real-valued. $\text{Vol}(\mathbf{v}_1,{\cdots},\mathbf{v}_d)$ is the unsigned volume of the paralleletope spanned by $\mathbf{v}_1,{\cdots},\mathbf{v}_d$.
%This paper is organized as follows. \textcolor{red}{fill in.}

\section{Background: Geometric Algebra}\label{section:geoalgebra}

%\textcolor{red}{give example instead here and move some of this to appendix?}

Consider the Euclidean vector space $\mathbb{R}^d$ with a scalar product (also called a inner or dot product for vector inputs). A \textit{geometric algebra over $\mathbb{R}^d$} is a vector space $\mathbb{G}^d$ (over the scalar field $\mathbb{R}$) of elements called \textit{multivectors}, equipped with an associative binary operation called a \textit{geometric product} that satisfies the following: (i) $\mathbb{G}^d$ contains $\mathbb{R}^d$ and is closed under the geometric product, (ii) geometric product is bilinear and distributes over addition, (iii) scalar multiplication is associative and commutative with the geometric product,  (iv) the geometric product of a vector with itself is equal to the scalar product of the vector with itself \cite{perwass:2009}. 

The geometric product of $A,B {\in} \mathbb{G}^d$ is denoted as $AB$. It can be shown that the geometric product of the basis vectors $e_i, e_j$ of $\mathbb{R}^d {\subset} \mathbb{G}^d$ satisfy $e_i e_i {=} 1$ and $e_i e_j {=} {-}e_j e_i$ \cite{perwass:2009}. Every multivector is a linear combination of $\textit{basis blades}$, which are multivectors that are geometric products of the form $E {=} e_{\mathcal{G}[1]} {\cdots} e_{\mathcal{G}[|\mathcal{G}|]}{=}\prod_{i\in\mathcal{G}}e_{i}$, where $\mathcal{G} {\subset} [d]$ is an ordered set with no repeated elements and $\mathcal{G}[i]$ is its $i^{\text{th}}$ element. A basis blade $E$ has \textit{grade} $g(E){=}|\mathcal{G}|$. Scalars are defined to have grade $0$. A dot product $A{\cdot} B$ and wedge product $A \wedge B$ are defined for multivectors in a way that generalizes the vector inner product and cross product, respectively (\Cref{section:geomalgebra}).  For vectors $A, B {\in} \mathbb{R}^d {\subset }\mathbb{G}^d$, $AB {=} A{\cdot} B {+} A {\wedge} B$ where the dot product represents the projection of $A$ onto $B$, and the wedge product represents the signed (oriented) area of the triangle spanned by $A$ and $B$.

%Clifford's algebra consists of a set of vector-like elements such as k-blades, which generalize vectors, and operations between them, including the wedge and geometric product, which generalize dot and cross products. 

%We consider a Clifford algebra over the Euclidean vector space $\mathbb{R}^d$ defined as a set of elements containing $e_1,\cdots, e_d$ which are called basis elements, that is closed under linear combinations and a binary operation called the geometric product.  
%A \textit{vector} is a linear combination of basis elements. The geometric product of two basis elements is also in the algebra and is denoted by their concatenation and defined by the rules that $e_ie_i = 1$ and $e_i e_j = -e_j e_i$, and extends associatively and linearly. For $n \geq 1$, a $n$-vector is a geometric product of $n$ basis elements. The \textit{dot} (or inner) product and \textit{wedge} (or outer) product are defined as \textcolor{red}{definition}. The dot product for vectors is consistent with the usual definition. The dot product gives is a scalar and represents the projection operation. The dot and wedge product extend linearly, and the wedge product is associative. %check
%For vectors, $\mathbf{v}_1 \mathbf{v}_2 = \mathbf{v}_1 \cdot \mathbf{v}_2 + \mathbf{v}_1 \wedge \mathbf{v}_2$.
%is this definition correct?
%
The \textit{generalized cross product} \cite{Berger:2009}, \cite{Pilanci:2023} of $\mathbf{v}_1,{\cdots},\mathbf{v}_{d{-}1} {\in} \mathbb{R}^d$ is defined as
\begin{equation}\label{eq:gencrosprod}
    \begin{aligned}
        \times_{i=1}^{d-1} \mathbf{v}_i = \sum_{i}^{d-1} (-1)^{i{-}1}|V_i| e_i 
    \end{aligned}
\end{equation}
where $V_i {=} \left[\mathbf{v}_1,{\cdots},\mathbf{v}_{i{-}1},\mathbf{v}_{i{+}1},{\cdots}, \mathbf{v}_{d{-}1}\right]$ is a $(d{-}1) {\times} (d{-} 1)$ square matrix and $e_i$ is the $i^{\text{th}}$ canonical basis vector. It can be shown that the generalized cross product is a vector that is orthogonal to $\mathbf{v}_1,{\cdots},\mathbf{v}_{d{-}1}$.
%By simplifying using the properties 
Since $e_i e_i{=}1$ and $e_i e_j {=} {-}e_j e_i$, the grade of all multivectors in $\mathbb{G}^d$ is at most $d$. A \textit{pseudoscalar} is a multivector of grade $d$. The \textit{unit pseudoscalar} is $\mathbf{I} {=} e_1 {\cdots}, e_d$.  
The \textit{unit pseudoscalar inverse}  is $\mathbf{I}^{-1}{=}e_d{\cdots} e_1$ which satisfies $\mathbf{I}^{-1}\mathbf{I}{=}1{=}\mathbf{I}\mathbf{I}^{-1}$. The \textit{Hodge star operator} $\star: \mathbb{G}^d {\to} \mathbb{G}^d$ is defined as $\star(\mathbf{v}){=}\star \mathbf{v} {=} \mathbf{v} \mathbf{I}^{-1}$. If $\mathbf{v}$ is a wedge product of vectors, $\star \mathbf{v}$ represents the orthogonal complement of their span.
It can be shown that the generalized cross product \eqref{eq:gencrosprod} can be equivalently written as
\begin{equation}\label{eq:cross_wedge}
    \begin{aligned}
        \times_{i=1}^{d-1} \mathbf{v}_i 
        &= \star \wedge_{i=1}^{d-1} \mathbf{v}_i.
        %&= \left( \wedge_{i=1}^{d-1} \sum_{j=1}^{d} {v_{i}}_j e_j \right) e_{d} \cdots e_1 \\
    \end{aligned}
\end{equation}
%todo include figure?
% The wedge product formulation of the generalized cross product gives a more compact expression and highlights that the generalized cross product is in the orthogonal complement of $\mathbf{v}_1,\cdots, \mathbf{v}_{d-1}$.
It can be shown that the $l_2$ norm of 
the generalized cross product of $\mathbf{v}_1,{\cdots},\mathbf{v}_{d{-}1}$ gives the $d{-}1$ volume of the parallelotope spanned by $\mathbf{v}_1,\cdots,\mathbf{v}_{d{-}1}$, while 
the inner product of $\times_{i=1}^{d-1} \mathbf{v}_i $ with a vector $\mathbf{v}_d$ gives the volume of the parallelotope spanned by $\mathbf{v}_1,\cdots,\mathbf{v}_d$ \citep{Berger:2009}. %todo be more precise/picture?

%-weights can be written as wedge products of differences of training data and training data with reflections -> picture?

%we can write reflection as geom prod? does this help?

\section{Prior work on shallow networks}\label{sectin:prior}
This section introduces the equivalences established in prior work between simple $2$-layer networks and \lasso{} problems. The \lasso{} problems have features representing volumes of parallelotopes spanned by training data.
The convex \lasso{} problem is \textit{equivalent} to the \nc{} training problem: they have the same optimal value and a network that is optimal in the training problem can be reconstructed from an optimal \lasso{} solution. \citep{1DNN} shows that

\begin{theorem}[\citep{1DNN}]\label{thrm:2layer1D}
     The training problem for a $2$-layer \reflectnn{} and \nd{1} data is equivalent to a \lasso{} problem with dictionary elements
     \begin{equation}\label{eq:2layer1D}
           \mathbf{A}_{i,j} = \left| x_i-x_{j} \right|  
     \end{equation}
     provided that $m \geq \|\mathbf{z}^*\|_0$, where $\mathbf{z}^*$ is a solution to the \lasso{} problem.
\end{theorem}

The dictionary elements $|x_i-x_j|$ in \Cref{thrm:2layer1D} measure the distance between training samples $x_i$ and $x_j$, or equivalently, their \nd{1} volume. \citep{pilanci2023complexity} extends this volume interpretation to higher dimensions as 

\begin{theorem}[\citep{pilanci2023complexity}]\label{thrm:2layer0bias}
     The training problem for a $2$-layer ReLU network with bias parameters set to $0$ and data of arbitrary dimension is equivalent to a \lasso{} problem with dictionary elements
     \begin{equation}\label{eq:2layerND}
           \mathbf{A}_{i,j} = \frac{\mathrm{Vol}_+(\mathbf{x}_i,\mathbf{x}'_{j_1},\cdots,\mathbf{x}'_{j_{d{-}1}})}{\big\| \mathbf{x}'_{j_1}\wedge\cdots\wedge\mathbf{x}'_{j_{d{-}1}}\big\|_1} 
           = \frac{ \big(\star ( \mathbf{x}_i \wedge \mathbf{x}'_{j_1}\wedge \cdots\wedge \mathbf{x}'_{j_{d{-}1}})\big)_+}{\big\|\mathbf{x}'_{j_1}\wedge\cdots\wedge\mathbf{x}'_{j_{d{-}1}}\big\|_1}
     \end{equation}
     where the multi-index $j{=}(j_1,{\cdots},j_{d-1})$ indexes over all combinations of $d{-}1$ samples of $\mathbf{x}_1,{\cdots},\mathbf{x}_N,e_1,{\cdots},e_d$.
    % provided that $m \geq \|\mathbf{z}^*\|_0$.
\end{theorem}

In \Cref{thrm:2layer0bias}, $\text{Vol}_+$ refers to the positive part of the signed volume. Similar results holds for other $2$-layer networks, including those with nonzero bias, and with absolute value activation.
This work extends the above results to deeper layers by analyzing \reflectnn{}s.

\section{Main Results}
In this section, we state our main results on the equivalence between training deep neural networks and \lasso{} problems with geometric features. In contrast to shallow networks (\Cref{sectin:prior}), the deeper layers enable features to represent reflections of training data.

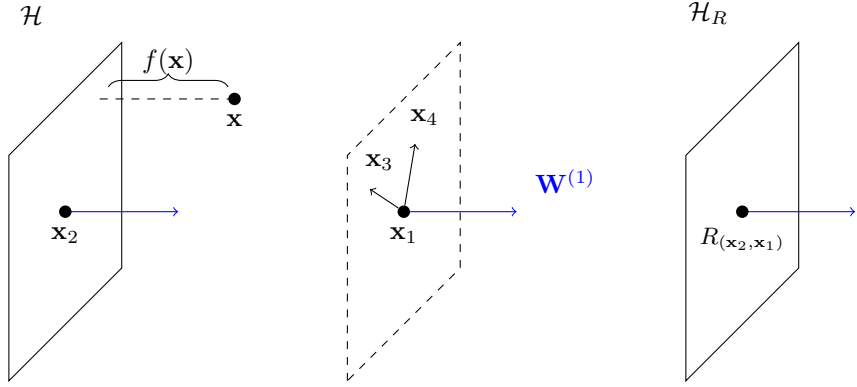
\begin{figure}
    \centering
    \begin{tikzpicture}[scale=1.5]
     %\draw[->]        (0,0) -- (\bx,\by) node[above, fill=white,label=above:$\mathbf{b}$] (b){};
        %\draw[] (0,0) -- (1,1) node[above, label=above:$l_1$] () {};
        %\draw[dashed] (1,0) -- (2,1);
        %\draw[] (2,0) -- (3,1) node[above, label=above:$l_2$](){};
        \draw[] (-3,0) -- (-3,2) -- (-2,3) -- (-2,1) -- cycle;
        \draw[] (-2.8,3) node[label=$\regplane$] () {};
        \draw[blue,->] (-2.5,1.5) node[fill=black, circle, scale=0.5, label=below:\textcolor{black}{$\mathbf{x}_{2}$}] () {} -- (-1.5,1.5) ;
        \draw[dashed] (0,0) -- (0,2) -- (1,3) -- (1,1) -- cycle; %j_0 = n^{(2)}
        \draw[blue,->] (0.5,1.5) node[ fill=black, circle, scale=0.5, label=below:\textcolor{black}{$\mathbf{x}_1$}] (a){} -- (1.5,1.5) node[ label=above right:\textcolor{blue}{$\wl{1}$}] () {};
        \draw[] (3,0) -- (3,2) -- (4,3) -- (4,1) -- cycle;
        \draw[] (3.2,3) node[label=$\reflectplane$] () {};
        \draw[blue, ->] (3.5,1.5) node[fill=black, circle, scale=0.5, label=below:\small\textcolor{black}{ $\reflect{\mathbf{x}_{2}}{\mathbf{x}_1}$}] () {} -- (4.5,1.5) ;
        \draw[->] (a) -- (0.2,1.7) node[right, label=$\mathbf{x}_3$] (){};
        \draw[->] (a) -- (0.6,2.1) node[right, label=${\mathbf{x}}_4$] (){};
        \draw[dashed] (-1,2.5) node[fill=black, circle, scale=0.5, label=below:\textcolor{black}{$\mathbf{x}$}] (xstart) {} --  (-2.2,2.5) node[] (xend){};
        \draw [decorate,decoration={brace,amplitude=5pt,mirror,raise=1ex}]
  (xstart) -- (xend) node[midway,yshift= 2pt, label=above:$f(\mathbf{x})$]{};
    \end{tikzpicture}
    \caption{Example of reflection planes for $3$-layer neural networks with data in $\mathbb{R}^3$.}
    \label{fig:3lyr_reflection_planes}
    %todo: label as l1, l2 and drop heights
\end{figure}

\subsection{$3$-layer neural networks}
We show that $3$-layer \reflectnn{}s are equivalent to \lasso{} problems with discrete, explicit dictionaries. Proofs in this section are deferred to \Cref{app:maintheorems}.
The next result gives the equivalence between the training problem \eqref{eq:trainingprob_explicit} for $3$-layer \reflectnn{}s \eqref{eq:deepnarrowabsvalnetwork} and \lasso{} models \eqref{eq:lasso}, using the language of geometric algebra (\Cref{section:geoalgebra}). Let $\text{Vol}(\mathbf{v}_1,{\cdots},\mathbf{v}_d)$ denote the volume of the parallelotope spanned by $\mathbf{v}_1,{\cdots},\mathbf{v}_d$.

\begin{theorem}\label{thrm:3layersimpler}
    The training problem for a $3$-layer \reflectnn{} and $d$-dimensional data is equivalent to a \lasso{} problem with dictionary elements 
    $ \mathbf{A}_{i,j} = \featurefunc(\mathbf{x}_i)$ defined as
    %$ \mathbf{A}_{i,j} = A_{\mathbf{x}_{j_0},\mathbf{x}'_{j_1},\mathbf{x}_{j_2},\mathbf{x}_{j_3}',\cdots,\mathbf{x}'_{j_{2d-1}}}(\mathbf{x}_i)$ where
    \begin{equation} \label{eq:3layerdicsimpler} 
        \begin{aligned}
        %A_{\mathbf{x}_{j_0},\mathbf{x}'_{j_1},\mathbf{x}_{j_2},\mathbf{x}_{j_3}',{\cdots},\mathbf{x}'_{j_{2d-1}}}(\mathbf{x}) 
        \featurefunc(\mathbf{x}) 
        &{=} \frac{\left|\mathrm{Vol}\left(\mathbf{x}{-}\mathbf{x}'_{j_1},\mathbf{x}_{j_2}{-}\mathbf{x}'_{j_3},{\cdots},\mathbf{x}_{j_{2(d{-}1)}}{-}\mathbf{x}'_{j_{2d{-}1}}\right){-}
            \mathrm{Vol}\left(\mathbf{x}_{j_0}{-}\mathbf{x}'_{j_1},\mathbf{x}_{j_2}{-}\mathbf{x}'_{j_3},{\cdots},\mathbf{x}_{j_{2(d{-}1)}}{-}\mathbf{x}'_{j_{2d{-}1}}\right)\right|}{\left\| 
            \left(\mathbf{x}_{j_2}{-}\mathbf{x}'_{j_3}\right){\times} {\cdots} {\times} \left(\mathbf{x}_{j_{2(d-1)}}{-}\mathbf{x}'_{j_{2d{-}1}}\right)
           \right \|_1}
        \end{aligned}
    \end{equation}
 provided that $m {\geq} \|\mathbf{z}^*\|_0$, where $\mathbf{z}^*, \xi^*$ is a \lasso{} solution. The multi-index $j{=}(j_{-1},j_0,j_1,j_2,{\cdots},j_{2d{-}1})$ indexes over all $j_{-1}, j_{2k} {\in} [N], \xprime {\in} \left\{\frac{\mathbf{x}_{j_{-1}}{+}\mathbf{x}_{j_0}}{2},\mathbf{x}_{j_{-1}}\right\}$ and 
% $\mathbf{x}_{j_{2k}} {\in}\left\{\mathbf{x}_1{\cdots},\mathbf{x}_{N}\right\}$ and $\mathbf{x}'_{j_{2k+1}}{\in }\mathcal{X}'{=}\left\{\frac{\mathbf{x}_k+\mathbf{x}_l}{2}\right\}_{k,l\in[N]} \cup \left\{\reflect{\mathbf{x}_k}{\mathbf{x}_l}\right\}_{k,l\in[N]} \cup \{\mathbf{x}_k-e_l\}_{k\in[N],l\in[d]}$,
$\mathbf{x}'_{j_{2k+1}}{\in }\left\{\mathbf{x}'_{j_1}, \reflect{\mathbf{x}_{j_0}}{\mathbf{x}'_{j_1}}\right\} {\cup} \{\mathbf{x}_{j_{2k}}{-}e_l\}_{l\in[d]}$ for $k{\geq}1$.
 An optimal neural network can be reconstructed as $\nn{3}{\mathbf{x}}{=}\sum_{i}{z}^*_i \featurefunc(\mathbf{x}){+} \xi^*$, and the corresponding optimal $\wl{1}$ is orthogonal to $\mathbf{x}_{j_2}{-}\mathbf{x}'_{j_3}, {\cdots} , \mathbf{x}_{j_{2(d-1)}}{-}\mathbf{x}'_{j_{2d{-}1}}$. %$\mathcal{A}$ and $R$ refer to the average and reflection operators, respectively.
\end{theorem}

\Cref{thrm:3layersimpler} gives a novel characterization of optimal neural networks as \lasso{} solutions.
\Cref{thrm:3layersimpler} states that we can train a globally optimal network by constructing a dictionary matrix, solving the \lasso{} problem \eqref{eq:lasso} using well-known techniques \citep{efron2004least}, and then using a \lasso{} solution to reconstruct an optimal network. The dictionary is constructed from differences of points in an \emph{augmented} training data set composed of reflections and averages of training data. For shorthand, we may refer to these augmented data points as the training data. Taking averages of training data has been used in effective data augmentation to improve the performance of training \citep{zhang:2018}. This \lasso{} equivalence may suggest theoretical reasons underlying this phenomenon. 
%
%The functions $\featurefunc(\mathbf{x})$ are features that interpolate the dictionary elements.
%
%featurefunc_j check no jj
%
We now discuss several advantages of using the \lasso{} problem to train networks and address various aspects of the model equivalence. 
%The reconstruction allows us to train neural networks using the \lasso{} model instead, which is convex.

\textbf{Training benefits:}
The convexity of the \lasso{} problem ensures global optimality. In \nc{} problems, training can get stuck in local, suboptimal optima depending on the initialization. The convex \lasso{} model avoids this problem. Additionally,
using the \lasso{} model reduces the need to tune hyperparameters such as the number of neurons $m$, as $m{=}\|\mathbf{z}^*\|_0$ suffices. 

\textbf{Uniqueness:} The training problem and \lasso{} problems may not have unique solutions. \citep{1DNN} discusses the relationship between the solution set of the training problem with \nd{1} data and the set of neural networks reconstructed from all \lasso{} solutions. Analysis of the \lasso{} solution sets and stationary points in the training problem for higher dimensional data is an area for future work. 

%n x d lasso complexity is n^2 d, linear in number of features, so making dictionary dominates

\textbf{Complexity:} The complexity of finding a \lasso{} solution with a dictionary of size $N{\times} F$ is $O(N^2 F)$ \citep{efron2004least}, which is linear in the number of feature vectors $F$. $F$ is finite and can be bounded. 
\begin{lemma}\label{lemma:Asize}
    The $3$-layer \lasso{} dictionary $\mathbf{A}$ in \Cref{thrm:3layersimpler} consists of $O\left((Nd)^d\right)$ feature vectors. The exponential complexity in $d$ can not be avoided for global optimization of ReLU networks unless $\mathbf{P}{=}\mathbf{NP}$. 
\end{lemma}
%todo update this lemma
%
\Cref{lemma:Asize} is an overestimate of the dictionary size, since the \lasso{} dictionary contains repeated columns and the presence of linearly dependent vectors $\mathbf{x}_{j_k}{-}\mathbf{x}_{j_{k{+}1}}'$ \eqref{eq:3layerdicsimpler} spanning a parallelotope will make its volume $0$, resulting in a $\mathbf{0}$ vector that does not contribute to the \lasso{} model. To reduce the computational overhead of creating the dictionary, the dictionary features can be subsampled \citep{Wang:2024}. \citep{wang2021hidden} suggests that subsampling hyperplanes corresponds to arriving at a local optima. Another practical usage of the \lasso{} model is to use it in a polishing step instead of using it to train the entire model from scratch. Polishing consists of partially training a network with the \nc{} training problem, extracting the \kink{}s from the neurons to estimate a \lasso{} dictionary, and then using the \lasso{} estimate to fine-tune the network \citep{pilanci2023complexity}. 
%Finally, our \lasso{} model can be adapted to convolutional networks, which have been shown to have significantly reduced complexity for training their convexifications \citep{pilanci2020neural}.

%There are promising approaches that can be adapted to reduce the complexity of the \lasso{} problem \citep{Wang:2024} by subsampling hyperplanes. 
%Adapting our approach to convolutional networks also reduces the complexity to polynomial in networks in \citep{pilanci2020neural}, which can be adapted. 
%The convexification approach can also be used instead of training the entire network, as a polishing tool using a limited number of features \citep{pilanci2023complexity}. %check is that how it works?
%Additionally, the training problem can be written as an equivalent \lasso{} problem with a smaller dictionary. See Appendix for details.

\textbf{Comparison to ReLU}: \label{ReLUcomparison}
% \citep{pilanci2023complexity} analyzes $3$-layer ReLU networks for arbitrary input dimension. In contrast to the ReLU activation, the absolute value activation allows for reflection features in the \reflectnn{} architecture. Intuitively, this is due to the $2$ symmetric "branches" $\{(x,\sigma(x){=}-x):x{\leq} 0\}$ and $\{(x,\sigma(x){=}x):x{\geq} 0\}$ of the piecewise linear absolute value activation, while the ReLU activation has an all-zero branch $\{(x,\sigma(x){=}0):x{\leq} 0\}$. 
\citep{1DNN} shows that $2$-layer networks with absolute value activation are equivalent to those with ReLU when the network uses a skip connection, and increasing the width of ReLU networks with \nd{1} input data introduces reflection features.
\Cref{fig:relu} illustrates a reflection feature occurring in a network trained on \nd{2} data. This suggests that wider ReLU architectures with higher-dimensional data will have reflection features, and is an area for future analysis. 

\textbf{Interpretibility:}  The $l_1$ regularization in the \lasso{} problem selects a minimal number of feature vectors which are discrete samples of \emph{features} $\featurefunc(\mathbf{x})$.  Intuitively, the network \emph{learns} these features, as a linear combination of feature functions corresponds to a parsimonious and optimal network. The features interpolate subsets of the training data. 
%Intuitively, a network \emph{learns} a fixed, finite set of functions called \textit{features} if there is an optimal network that is a linear combination of those features. 
%The functions $A_{j}(\mathbf{x})$ are features. T
The mapping between the features and optimal weights is in the Appendix (\Cref{def:reconstruct}).
%The reconstruction formula (\Cref{def:reconstruct}) simply makes the network a linear combination of the features \eqref{eq:feature} corresponding to each column of the dictionary matrix, scaled by the solution $z^*_i$. 
\Cref{thrm:3layersimpler} gives a volumetric interpretation of features. The features also have geometric algebraic and a distance-based interpretations. 

A vector $\mathbf{x}$ is \emph{sparse} if it has few non-zero elements. The \emph{sparsity factor} of $\mathbf{x}{\neq}\mathbf{0}$ is $\lratiopenalty(\mathbf{x}){=}\frac{\|\mathbf{x}\|_2}{\|\mathbf{x}\|_1} {\in} \left[\frac{1}{\sqrt{N}},1\right]$. The sparsity factor is a measure of a vector's sparseness. If $\mathbf{x}$ has one non-zero element, then $\lratiopenalty(\mathbf{x}){=}1$ and if $\mathbf{x}$ is parallel to $\mathbf{1}$, then $\lratiopenalty(\mathbf{x}){=}\frac{1}{\sqrt{N}}$. The more sparse $\mathbf{x}$ is, the larger its sparsity factor.  %The $3$-layer feature function \eqref{eq:3layerfeature} has several geometric interpretations. 
Next, given $j_{{-}1},j_0 {\in} [N], \mathbf{w} {\in} \mathbb{R}^d$, define the hyperplanes $\regplane{=} \{\mathbf{x} {\in} \mathbb{R}^d{:} (\mathbf{x}{-}\mathbf{x}_{j_{{-}1}})\mathbf{w}{=}0\}, \mathcal{H}_{0}{=} \{\mathbf{x}{\in} \mathbb{R}^d{:} (\mathbf{x}{-}\mathbf{x}_{j_0})\mathbf{w}{=}0\}$. The \emph{average} of the hyperplanes $\regplane{}$ and  $\mathcal{H}_{0}$ is the hyperplane $\avgplane{=}\left\{\mathbf{x}{\in} \mathbb{R}^d: \left(\mathbf{x}{-}\frac{\mathbf{x}_{j_{{-}1}}+\mathbf{x}_{j_0}}{2}\right)\mathbf{w}{=}0\right\}$. The \emph{reflection} of the hyperplanes $\regplane$ across $\mathcal{H}_0$ is  $\reflectplane{=}\left\{\mathbf{x}{\in} \mathbb{R}^d: \left(\mathbf{x}{-}\reflect{\mathcal{H}_{j_{-1}}}{\mathcal{H}_{j_0}}\right)\mathbf{w}{=}0\right\}$. These definitions will be used in the next result to describe network features.
%The average and reflection operators output the average and reflection of hyperplanes, respectively. These operators characterize the features of a $3$-layer network. 
%simplify H to make j, w implicit?

\begin{figure}
\begin{tikzpicture}[scale=0.65, font = \normalsize, declare function={
 f(\x,\y)=abs(abs((\x-\ax)*\wx+(\y-\ay)*\wy)-abs((\jx-\ax)*\wx+(\jy-\ay)*\wy));
}]
%choose
\pgfmathsetmacro{\ix}{3} %x_n1 = x_i (1st coord)
\pgfmathsetmacro{\iy}{3} %x_n1 = x_i (2nd coord)
\pgfmathsetmacro{\jx}{1} %x_n2 = x_j (1st coord)
\pgfmathsetmacro{\jy}{2} %x_n2 = x_j (2nd coord)
\pgfmathsetmacro{\kx}{2} %1  %x_n3 = x_k used to get w^(1) (1st coord).
\pgfmathsetmacro{\ky}{1}  %2.5 x_n3 (2nd coord)

%determined
\pgfmathsetmacro{\ax}{\ix} %a=x_i=x_j1 to get reflections. (1st coord)
\pgfmathsetmacro{\ay}{\iy} % a (2nd coord)
\pgfmathsetmacro{\wx}{-(\ky-\ay)} %1. w^(1) is orthogonal to diff of data point and a. (1st coord)
\pgfmathsetmacro{\wy}{\kx-\ax} %0.5 w^(2) (2nd coord)
\pgfmathsetmacro{\xmin}{-2}
\pgfmathsetmacro{\xmax}{6}
\pgfmathsetmacro{\ymin}{-4}
\pgfmathsetmacro{\ymax}{6}
%reflection points
\pgfmathsetmacro{\rx}{2*\ax-\jx}
\pgfmathsetmacro{\ry}{2*\ay-\jy}
%reflection plane H1
\pgfmathsetmacro{\HtwoMaxx}{\jx+2.6*(\kx-\ix)}
\pgfmathsetmacro{\HtwoMaxy}{\jy+2.6*(\ky-\iy)}
\pgfmathsetmacro{\HtwoMinx}{\jx-2*(\kx-\ix)}
\pgfmathsetmacro{\HtwoMiny}{\jy-2*(\ky-\iy)}
%reflection plane HR
\pgfmathsetmacro{\HrMaxx}{\rx+3.7*(\kx-\ix)}
\pgfmathsetmacro{\HrMaxy}{\ry+3.7*(\ky-\iy)}
\pgfmathsetmacro{\HrMinx}{\rx-1*(\kx-\ix)}
\pgfmathsetmacro{\HrMiny}{\ry-1*(\ky-\iy)}

\begin{axis}[
view={-30}{60}, %view={-5}{10}. %view={0}{90} gives flat
colormap/viridis,
hide axis,
xmin=\xmin, xmax=\xmax,
ymin=\ymin, ymax=\ymax,
zmin=0, zmax=6,
clip=false,
]
\addplot3[
domain=\xmin:\xmax,
domain y=\ymin:\ymax,
samples=50,
%samples y=40,
surf,
]{f(\x,\y)};
%reflection plane H2
\addplot3[dashed,red] coordinates {(\HtwoMinx,\HtwoMiny,0)(\HtwoMaxx,\HtwoMaxy,0) };
\addplot3[red, mark=none, point meta=explicit symbolic, nodes near coords] coordinates {(\HtwoMaxx+0.3,\HtwoMaxy-1.2,0)  [$\regplane$]}; %label
%reflection plane Hr
\addplot3[dashed,red] coordinates {(\HrMinx,\HrMiny,0)(\HrMaxx,\HrMaxy,0) };
\addplot3[red, mark=none, point meta=explicit symbolic, nodes near coords] coordinates {(\HrMaxx-0.3,\HrMaxy-2,0)  [$\reflectplane$]};
%data points
%\node at (rel axis cs:4,4,0) [below] {$x$};
\addplot3[mark=*,red,point meta=explicit symbolic,nodes near coords] 
coordinates {(\ix,\iy,0)[$\mathbf{x}_{1}$]}; %[\footnotesize$\mathbf{x}_{j_1}{=}\mathbf{a}$]
\addplot3[mark=*, red,point meta=explicit symbolic,nodes near coords] 
coordinates {(\jx,\jy,0)};%[\footnotesize$\mathbf{x}_{j_2}$]};
\addplot3[red, mark=none, point meta=explicit symbolic, nodes near coords] coordinates {(\jx-0.5,\jy-0.1,0)  [$\mathbf{x}_{2}$]};
%
%\addplot3[mark=*,red,point meta=explicit symbolic,nodes near coords] coordinates {(2*\jx-\ax,2*\jy-\ay,0)  [\footnotesize$\reflect{\mathbf{x}_{j_1}}{\mathbf{x}_{j_2}}$]};
\addplot3[mark=*,red,point meta=explicit symbolic,nodes near coords] coordinates {(\rx,\ry,0)  };%[\footnotesize$\reflect{\mathbf{x}_{j_2}}{\mathbf{x}_{j_1}}$]};
\addplot3[red, mark=none, point meta=explicit symbolic, nodes near coords] coordinates {(\rx+1,\ry-1,0)  [$\reflect{\mathbf{x}_{2}}{\mathbf{x}_{1}}$]};
\addplot3[mark=*,red,point meta=explicit symbolic,nodes near coords] coordinates {(\kx,\ky,0)};  %[\footnotesize$\mathbf{x}_{j_3}$]};
\addplot3[red, mark=none, point meta=explicit symbolic, nodes near coords] coordinates {(\kx-0.4,\ky-1.6,0)  [$\mathbf{x}_{3}$]};

%\fill (0,0) circle (2pt) node[above right] {above right};
%\fill (canvas cs:x=4,y=0)    circle (2pt);
%\node[red, anchor=east] coordinates {(2*\ix-\jx,2*\iy-\jy,0) [$\reflect{\mathbf{x}_{j_2}}{\mathbf{x}_{j_1}}$]};
%draw[->, thick, red] (0,0,0) -- (\wx,\wy,0) node[above] {$\wl{1}$};
\draw[->, red, thick] (axis cs: \ax,\ay) -- (axis cs: \ax+\wx,\ay+\wy) node[right] {\small$\wl{1}$};
\draw[->, red, thick] (axis cs: \ax,\ay) -- (axis cs: \kx,\ky) node[below] {}; % {\footnotesize$\mathbf{x}_{j_3}{-}\mathbf{a}$};
\end{axis}
\end{tikzpicture}
%%%%%%%%%%%%%
%4 layer with just single reflections
\begin{tikzpicture}[scale=0.65, font = \normalsize, declare function={
 f(\x,\y)=abs(abs(abs((\x-\ax)*\wx+(\y-\ay)*\wy)-abs((\jx-\ax)*\wx+(\jy-\ay)*\wy))-abs(abs((\hx-\ax)*\wx+(\hy-\ay)*\wy)-abs((\jx-\ax)*\wx+(\jy-\ay)*\wy)));
 %f(\x,\y)=abs(abs((\x-\ax)*\wx+(\y-\ay)*\wy)-abs((\jx-\ax)*\wx+(\jy-\ay)*\wy));
}]
%choose
\pgfmathsetmacro{\ix}{3} %x_n1 = x_i (1st coord)
\pgfmathsetmacro{\iy}{3} %x_n1 = x_i (2nd coord)
\pgfmathsetmacro{\jx}{1} %x_n2 = x_j (1st coord)
\pgfmathsetmacro{\jy}{2} %x_n2 = x_j (2nd coord)
\pgfmathsetmacro{\kx}{2} %1  %x_n3 = x_k used to get w^(1) (1st coord).
\pgfmathsetmacro{\ky}{1}  %2.5 x_n3 (2nd coord)
\pgfmathsetmacro{\hx}{5} %make h big enough
\pgfmathsetmacro{\hy}{0}

%determined
\pgfmathsetmacro{\ax}{\ix} %a=x_i=x_j1 to get reflections. (1st coord)
\pgfmathsetmacro{\ay}{\iy} % a (2nd coord)
\pgfmathsetmacro{\wx}{-(\ky-\ay)} %1. w^(1) is orthogonal to diff of data point and a. (1st coord)
\pgfmathsetmacro{\wy}{\kx-\ax} %0.5 w^(2) (2nd coord)
\pgfmathsetmacro{\xmin}{0}%-2
\pgfmathsetmacro{\xmax}{8}%6
\pgfmathsetmacro{\ymin}{-2}%-4
\pgfmathsetmacro{\ymax}{10}%6
%reflections
\pgfmathsetmacro{\rx}{2*\ax-\hx}
\pgfmathsetmacro{\ry}{2*\ay-\hy}
%reflection plane H1
\pgfmathsetmacro{\HtwoMaxx}{\ix+2.5*(\kx-\ix)}
\pgfmathsetmacro{\HtwoMaxy}{\iy+2.5*(\ky-\iy)}
\pgfmathsetmacro{\HtwoMinx}{\ix-4*(\kx-\ix)}
\pgfmathsetmacro{\HtwoMiny}{\iy-4*(\ky-\iy)}
%reflection plane H4
\pgfmathsetmacro{\HfourMaxx}{\hx+1*(\kx-\ix)}
\pgfmathsetmacro{\HfourMaxy}{\hy+1*(\ky-\iy)}
\pgfmathsetmacro{\HfourMinx}{\hx-3*(\kx-\ix)}
\pgfmathsetmacro{\HfourMiny}{\hy-3*(\ky-\iy)}
%reflection plane HR
\pgfmathsetmacro{\HrMaxx}{\rx+1*(\kx-\ix)}
\pgfmathsetmacro{\HrMaxy}{\ry+1*(\ky-\iy)}
\pgfmathsetmacro{\HrMinx}{\rx-1.8*(\kx-\ix)}
\pgfmathsetmacro{\HrMiny}{\ry-1.8*(\ky-\iy)}

\begin{axis}[
view={-30}{60}, %view={-30}{60}. %view={0}{90} gives flat
colormap/viridis,
hide axis,
xmin=\xmin, xmax=\xmax,
ymin=\ymin, ymax=\ymax,
zmin=0, %zmax=6,
clip=false,
]
\addplot3[
domain=\xmin:\xmax,
domain y=\ymin:\ymax,
samples=50, %60 in previous
surf,
]{f(\x,\y)};
%reflection plane H2
\addplot3[dashed,red] coordinates {(\HtwoMinx,\HtwoMiny,0)(\HtwoMaxx,\HtwoMaxy,0) };
\addplot3[red, mark=none, point meta=explicit symbolic, nodes near coords] coordinates {(\HtwoMinx-0.3,\HtwoMiny+0.5,0)  [$\regplane$]}; 
%reflection plane H4
\addplot3[dashed,red] coordinates {(\HfourMinx,\HfourMiny,0)(\HfourMaxx,\HfourMaxy,0) };
\addplot3[red, mark=none, point meta=explicit symbolic, nodes near coords] coordinates {(\HfourMinx+0.3,\HfourMiny+0.5,0)  [${\regplane}'$]}; 
%reflection plane Hr
\addplot3[dashed,red] coordinates {(\HrMinx,\HrMiny,0)(\HrMaxx,\HrMaxy,0) };
\addplot3[red, mark=none, point meta=explicit symbolic, nodes near coords] coordinates {(\HrMinx+0.3,\HrMiny+0.8,0)  [$\reflectplane$]};
%data
%x1
\addplot3[mark=*,red,point meta=explicit symbolic,nodes near coords] 
coordinates {(\ix,\iy,0)[]}; %
\addplot3[red, mark=none, point meta=explicit symbolic, nodes near coords] coordinates {(\ix-0.5,\iy,0)  [$\mathbf{x}_{1}$]};
%x2
\addplot3[mark=*,red,point meta=explicit symbolic,nodes near coords] 
coordinates {(\jx,\jy-0.1,0)[$\mathbf{x}_{2}$]};
%x3
\addplot3[mark=*,red,point meta=explicit symbolic,nodes near coords] coordinates {(\kx,\ky,0)};
\addplot3[red, mark=none, point meta=explicit symbolic, nodes near coords] coordinates {(\kx+0.5,\ky-0.5,0)  [$\mathbf{x}_{3}$]};
%x4
\addplot3[mark=*,red,point meta=explicit symbolic,nodes near coords] 
coordinates {(\hx,\hy,0)}; %
\addplot3[mark=none,red,point meta=explicit symbolic,nodes near coords] 
coordinates {(\hx-0.7,\hy-0.5,0)[$\mathbf{x}_{4}$]};
%R(x2,x1)
\addplot3[mark=*,red,point meta=explicit symbolic,nodes near coords] coordinates {(2*\ax-\jx,2*\ay-\jy,0)  [$\reflect{\mathbf{x}_{2}}{\mathbf{x}_{1}}$]};
%R(x4,x1)
\addplot3[mark=*,red,point meta=explicit symbolic,nodes near coords] coordinates {(\rx,\ry,0)  []};
\addplot3[mark=none,red,point meta=explicit symbolic,nodes near coords] 
coordinates {(\rx+1.3,\ry-0.5,0)[$\reflect{\mathbf{x}_{4}}{\mathbf{x}_{1}}$]};
%vectors w
\draw[->, red, thick] (axis cs: \ax,\ay) -- (axis cs: \ax+\wx,\ay+\wy) node[right] {$\wl{1}$};
\draw[->, red, thick] (axis cs: \ax,\ay) -- (axis cs: \kx,\ky) node[below] {};
\end{axis}
\end{tikzpicture}
%
%%%%%%%%%%
%4layer with double reflections
\begin{tikzpicture}[scale=0.65, font = \normalsize, declare function={
 f(\x,\y)=abs(abs(abs((\x-\ax)*\wx+(\y-\ay)*\wy)-abs((\jx-\ax)*\wx+(\jy-\ay)*\wy))-abs(abs((\hx-\ax)*\wx+(\hy-\ay)*\wy)-abs((\jx-\ax)*\wx+(\jy-\ay)*\wy)));
 %f(\x,\y)=abs(abs((\x-\ax)*\wx+(\y-\ay)*\wy)-abs((\jx-\ax)*\wx+(\jy-\ay)*\wy));
}]
%choose
\pgfmathsetmacro{\ix}{3} %x_n1 = x_i (1st coord)
\pgfmathsetmacro{\iy}{3} %x_n1 = x_i (2nd coord)
\pgfmathsetmacro{\jx}{1} %x_n2 = x_j (1st coord)
\pgfmathsetmacro{\jy}{2} %x_n2 = x_j (2nd coord)
\pgfmathsetmacro{\kx}{2} %1  %x_n3 = x_k used to get w^(1) (1st coord).
\pgfmathsetmacro{\ky}{1}  %2.5 x_n3 (2nd coord)
\pgfmathsetmacro{\hx}{4}
\pgfmathsetmacro{\hy}{0}

%determined
\pgfmathsetmacro{\ax}{\ix} %a=x_i=x_j1 to get reflections. (1st coord)
\pgfmathsetmacro{\ay}{\iy} % a (2nd coord)
\pgfmathsetmacro{\wx}{-(\ky-\ay)} %1. w^(1) is orthogonal to diff of data point and a. (1st coord)
\pgfmathsetmacro{\wy}{\kx-\ax} %0.5 w^(2) (2nd coord)
\pgfmathsetmacro{\xmin}{0}%-2
\pgfmathsetmacro{\xmax}{8}%6
\pgfmathsetmacro{\ymin}{-2}%-4
\pgfmathsetmacro{\ymax}{10}%6
%single reflection
\pgfmathsetmacro{\rx}{2*\ix-\hx}
\pgfmathsetmacro{\ry}{2*\iy-\hy}
%doube reflection
\pgfmathsetmacro{\rrx}{2*(2*\ax-\jx)-\hx}
\pgfmathsetmacro{\rry}{2*(2*\ay-\jy)-\hy}
%other double reflection R(R(x1,x4),x2)
\pgfmathsetmacro{\rrrx}{2*\jx-\rx}
\pgfmathsetmacro{\rrry}{2*\jy-\ry}
%reflection plane H4
\pgfmathsetmacro{\HfourMaxx}{\hx+1*(\kx-\ix)}
\pgfmathsetmacro{\HfourMaxy}{\hy+1*(\ky-\iy)}
\pgfmathsetmacro{\HfourMinx}{\hx-4*(\kx-\ix)}
\pgfmathsetmacro{\HfourMiny}{\hy-4*(\ky-\iy)}
%reflection plane double
\pgfmathsetmacro{\HrrMaxx}{\rrx+4.7*(\kx-\ix)}
\pgfmathsetmacro{\HrrMaxy}{\rry+4.7*(\ky-\iy)}
\pgfmathsetmacro{\HrrMinx}{\rrx-1*(\kx-\ix)}
\pgfmathsetmacro{\HrrMiny}{\rry-1*(\ky-\iy)}
%reflection plane single
\pgfmathsetmacro{\HrMaxx}{\rx+1.7*(\kx-\ix)}
\pgfmathsetmacro{\HrMaxy}{\ry+1.7*(\ky-\iy)}
\pgfmathsetmacro{\HrMinx}{\rx-2*(\kx-\ix)}
\pgfmathsetmacro{\HrMiny}{\ry-2*(\ky-\iy)}
%other double reflection plane
\pgfmathsetmacro{\HrrrMaxx}{\rrrx+0.5*(\kx-\ix)}
\pgfmathsetmacro{\HrrrMaxy}{\rrry+0.5*(\ky-\iy)}
\pgfmathsetmacro{\HrrrMinx}{\rrrx-5.8*(\kx-\ix)}
\pgfmathsetmacro{\HrrrMiny}{\rrry-5.8*(\ky-\iy)}

\begin{axis}[
view={-30}{60}, %view={-30}{60}. %view={0}{90} gives flat
colormap/viridis,
hide axis,
xmin=\xmin, xmax=\xmax,
ymin=\ymin, ymax=\ymax,
zmin=0, %zmax=6,
clip=false,
]
\addplot3[
domain=\xmin:\xmax,
domain y=\ymin:\ymax,
samples=50,
surf,
]{f(\x,\y)};
%reflection plane H2
\addplot3[dashed,red] coordinates {(\HfourMinx,\HfourMiny,0)(\HfourMaxx,\HfourMaxy,0) };
\addplot3[red, mark=none, point meta=explicit symbolic, nodes near coords] coordinates {(\HfourMinx+0.5,\HfourMiny+0.5,0)  [$\regplane$]}; 
%Double reflection plane
\addplot3[dashed,red] coordinates {(\HrrMinx,\HrrMiny,0)(\HrrMaxx,\HrrMaxy,0) };
\addplot3[red, mark=none, point meta=explicit symbolic, nodes near coords] coordinates {(\HrrMaxx-1,\HrrMaxy-3,0)  [${{\reflectplane}_R}'$]}; 
%Single reflection plane
\addplot3[dashed,red] coordinates {(\HrMinx,\HrMiny,0)(\HrMaxx,\HrMaxy,0) };
\addplot3[red, mark=none, point meta=explicit symbolic, nodes near coords] coordinates {(\HrMinx,\HrMiny+1,0)  [${\reflectplane}$]}; 
%Other Double reflection plane
\addplot3[dashed,red] coordinates {(\HrrrMinx,\HrrrMiny,0)(\HrrrMaxx,\HrrrMaxy,0) };
\addplot3[red, mark=none, point meta=explicit symbolic, nodes near coords] coordinates {(\HrrrMinx,\HrrrMiny+1,0)  [${{\reflectplane}_R}$]}; 
%R(x4,x1)
\addplot3[mark=*,red,point meta=explicit symbolic,nodes near coords] coordinates {(\rx,\ry,0)  []};
\addplot3[red, mark=none, point meta=explicit symbolic, nodes near coords] coordinates {(\rx-0.45,\ry+1,0)  [$\reflect{\mathbf{x}_{1}}{\mathbf{x}_{4}}$]}; 
\addplot3[mark=*,red,point meta=explicit symbolic,nodes near coords] 
coordinates {(\ix,\iy,0)[$\mathbf{x}_{1}$]}; %
\addplot3[mark=*,red,point meta=explicit symbolic,nodes near coords] 
coordinates {(\jx,\jy-0.1,0)[$\mathbf{x}_{2}$]};
\addplot3[mark=*,red,point meta=explicit symbolic,nodes near coords] coordinates {(\kx,\ky,0)};
\addplot3[red, mark=none, point meta=explicit symbolic, nodes near coords] coordinates {(\kx-0.8,\ky-1.8,0)  [$\mathbf{x}_{3}$]};
%x4
\addplot3[mark=*,red,point meta=explicit symbolic,nodes near coords] 
coordinates {(\hx,\hy,0)[]}; %
\addplot3[mark=none,red,point meta=explicit symbolic,nodes near coords] 
coordinates {(\hx-0.7,\hy-0.5,0)[$\mathbf{x}_{4}$]};
%R(x2,x1)
\addplot3[mark=*,red,point meta=explicit symbolic,nodes near coords] coordinates {(2*\ax-\jx,2*\ay-\jy,0)  [$\reflect{\mathbf{x}_{2}}{\mathbf{x}_{1}}$]};
%double reflection
\addplot3[mark=*,red,point meta=explicit symbolic,nodes near coords] coordinates {(2*(2*\ax-\jx)-\hx,2*(2*\ay-\jy)-\hy,0)  [$\reflect{\mathbf{x}_{4}}{\reflect{\mathbf{x}_{2}}{\mathbf{x}_{1}}}$]};
%other double reflection
\addplot3[mark=*,red,point meta=explicit symbolic,nodes near coords] coordinates {(\rrrx,\rrry,0)  []};
\addplot3[red, mark=none, point meta=explicit symbolic, nodes near coords] coordinates {(\rrrx-2.2,\rrry-0.5,0)  [$\reflect{\reflect{\mathbf{x}_1}{\mathbf{x}_4}}{\mathbf{x}_2}$]}; 
\draw[->, red, thick] (axis cs: \ax,\ay) -- (axis cs: \ax+\wx,\ay+\wy) node[right] {$\wl{1}$};
\draw[->, red, thick] (axis cs: \ax,\ay) -- (axis cs: \kx,\ky) node[below] {};
\end{axis}
\end{tikzpicture}
\caption{Examples of $3$-layer (left) and $4$-layer (middle, right) \reflectnn{} features for \nd{2} data. Red dashed lines indicate reflection planes (lines in $\mathbb{R}^2$). %Observe that $\wl{1}$ is perpendicular to $\mathbf{x}_{j_3}-\mathbf{x}_{j_1}$ and is the direction in which the feature changes.% The \kink{}s occur along $\left(\mathbf{x}-\mathbf{x}_{j_1}\right)\wl{1}=0, \left(\mathbf{x}-\mathbf{x}_{j_2}\right)\wl{1}=0$ and $ \left(\mathbf{x}-\reflect{x_{j_2}}{\mathbf{x}_{j_1}}\right)\wl{1}=0$. Here, $\mathbf{a}=\mathbf{x}_{j_1}$.
}
 \label{fig:absvalfeat}
\end{figure}
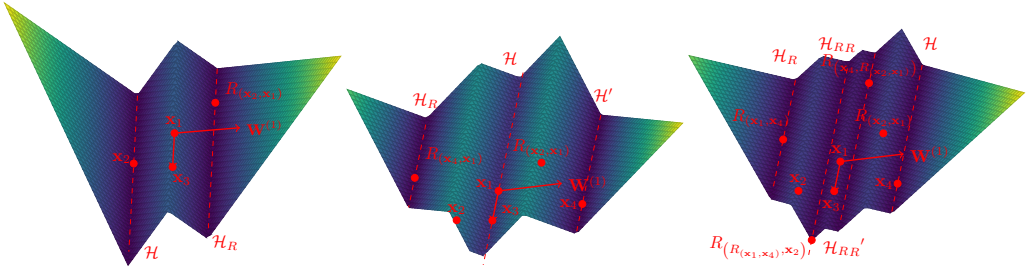
\begin{theorem}\label{theorem:geometricInterp}
The $3$-layer \reflectnn{} features in \Cref{thrm:3layersimpler} can be written as
    \begin{equation}\label{eq:f_as_distance_wedge}
    \scalemath{0.9}{
        \begin{aligned}
             \featurefunc(\mathbf{x})&{=} \frac{\biggl|\left|{\star}\left(\mathbf{x}{-}\xprime {\wedge} \mathbf{x}_{j_2}{-}\mathbf{x}'_{j_3}{\wedge}{\cdots} \wedge \mathbf{x}_{j_{2(d{-}1)}}{-}\mathbf{x}'_{j_{2d{-}1}}\right)\right|{-}\left|{\star}\left(\mathbf{x}_{j_0}{-}\xprime {\wedge} \mathbf{x}_{j_2}{-}\mathbf{x}'_{j_3}{\wedge}{\cdots} {\wedge }\mathbf{x}_{j_{2(d-1)}}{-}\mathbf{x}'_{j_{2d{-}1}}\right)\right|\biggr|}{\big\| 
            (\mathbf{x}_{j_2}{-}\mathbf{x}'_{j_3})\wedge \cdots \wedge (\mathbf{x}_{j_{2(d-1)}}{-}\mathbf{x}'_{j_{2d{-}1}})
           \big \|_1}\\
           &=    \lratiopenalty\left(\mathbf{w}\right)
           \begin{cases}
              \mathrm{Dist}\left(\mathbf{x},\regplane\cup\reflectplane\right) &\mbox{ if } \xprime = \mathbf{x}_{j_{-1}} \\
            %\mathrm{Dist}\left(\mathbf{x},\mathcal{H}_{(j_1,j_{2:2d-1})}\cup\mathcal{H}_{(\frac{\mathbf{x}_{j_1}+\mathbf{x}_{j_0}}{2},j_{2:2d-1})}\right)
            \mathrm{Dist}\left(\mathbf{x},\regplane\cup\avgplane\right)
            &\mbox{ else } %\mathbf{x}_{j_1}'=\frac{\mathbf{x}_{j_1}+\mathbf{x}_{j_0}}{2}\\
           \end{cases}
        \end{aligned}
        }
    \end{equation}
     where $\mathbf{w}{=}\left((\mathbf{x}_{j_2}{-}\mathbf{x}_{j_3}') {\times} {\cdots} {\times} (\mathbf{x}'_{j_{2(d-1)}}{-}\mathbf{x}'_{j_{2d-1}})\right)^T$. %$\xplane_{(j_1,j_{2:2d-1})} {=} \{\mathbf{x}{:} (\mathbf{x}{-}\mathbf{x}_{j_1})\mathbf{w}{=}0\}$. 
     In the reconstructed optimal network, $\wl{1}$ is a scalar multiple of $\mathbf{w}$.
\end{theorem}
\Cref{theorem:geometricInterp} shows that neural network features represent distances to parallel planes containing averages and reflections of training data. Moreover, these planes are orthogonal to $\wl{1}$ and are spanned by a subset of training data. \Cref{theorem:geometricInterp} also describes the features as wedge products, and gives a formula to explicitly compute the dictionary elements using wedge products or generalized cross products \eqref{eq:gencrosprod}, \eqref{eq:cross_wedge}. 

%ontaining a data point or a reflection, scaled by the sparsity factor of the wedge product of training data. 
\Cref{theorem:geometricInterp} shows that the \lasso{} dictionary includes \emph{reflection features} that measure the minimum distance to two parallel \emph{reflection planes}: $\regplane$, which contains $\mathbf{x}_{j_{-1}}$, and $\reflectplane$, which is the reflection of $\regplane$
 across the hyperplane $\mathcal{H}_{0}$ containing $\mathbf{x}_{j_0}$. Since distance is an unsigned value, the network induces a symmetry about the two planes.  
%
%The second equality of \Cref{theorem:geometricInterp} interprets the features as distances to hyperplanes spanned by training data. Geometric algebra enabled interpretation of features from a geometric lens.
%
%The reconstruction allows us to train neural networks using the \lasso{} model instead, which is convex.
%
%\Cref{thrm:3layersimpler} states that the reconstructed optimal first-layer weight vector $\wl{1}$ is orthogonal to the differences of training samples. 
%
%which has also been studied in \citep{Yin:2014,Xu:2021}. 
%The hyperplanes are points in $\mathbb{R}$ and lines in $\mathbb{R}^2$. 
%
\begin{figure}
    \centering
    \begin{minipage}[t]{0.3\textwidth}
    \includegraphics[width=\linewidth]{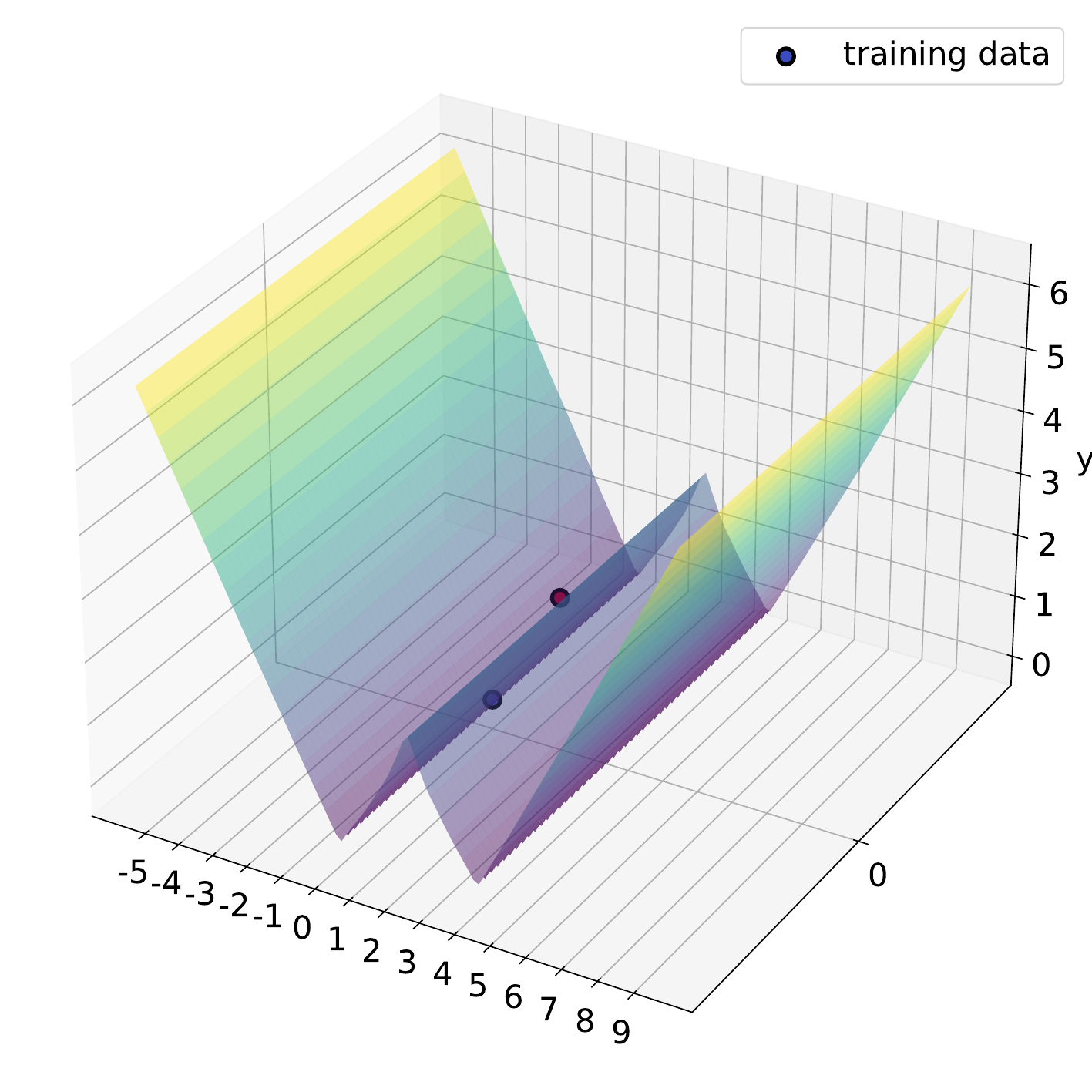}
    \end{minipage}
    \begin{minipage}[t]{0.3\textwidth}
    \includegraphics[width=\linewidth]{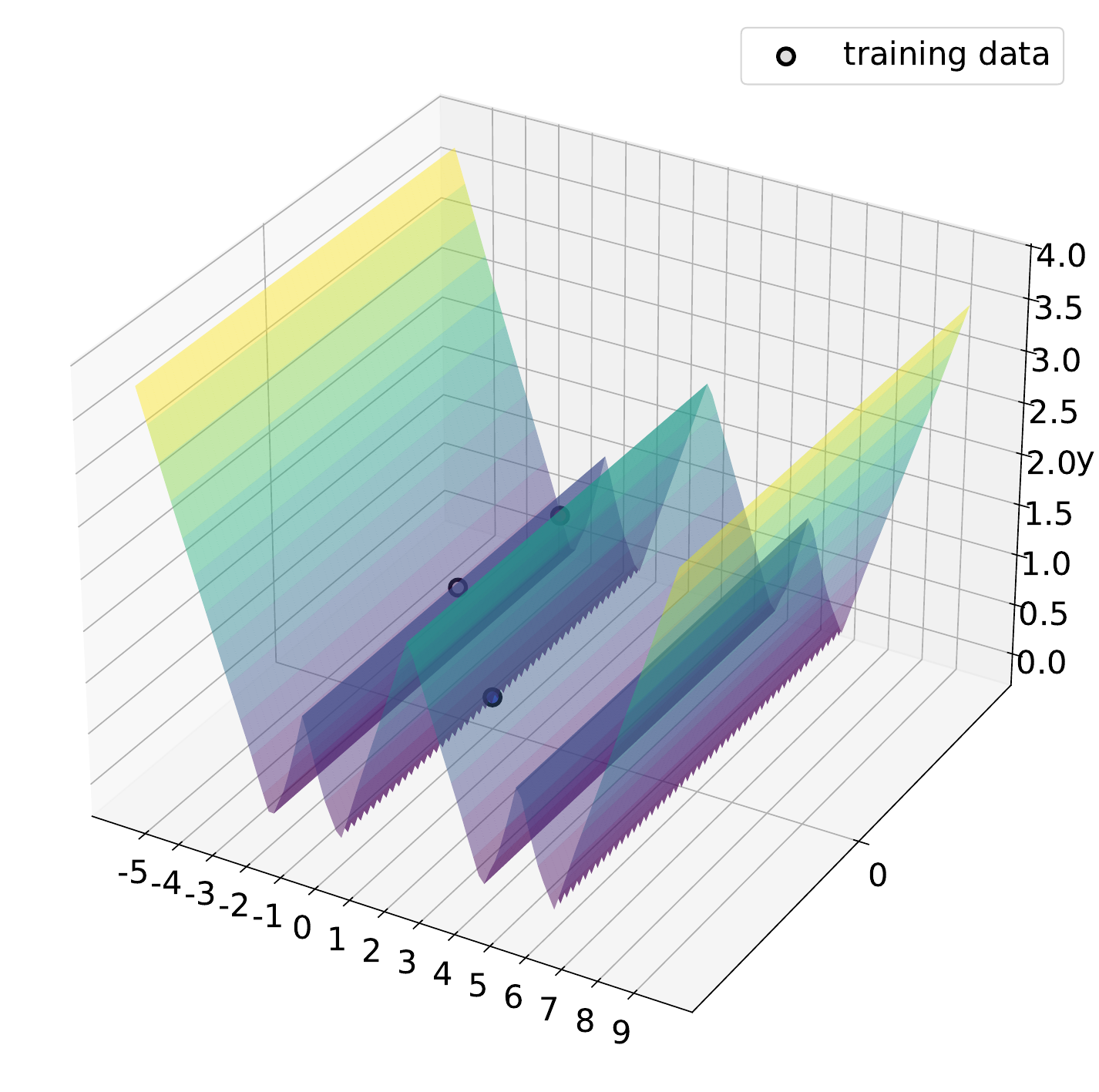}
    \end{minipage}
    \begin{minipage}[t]{0.3\textwidth}
    \includegraphics[width=\linewidth]{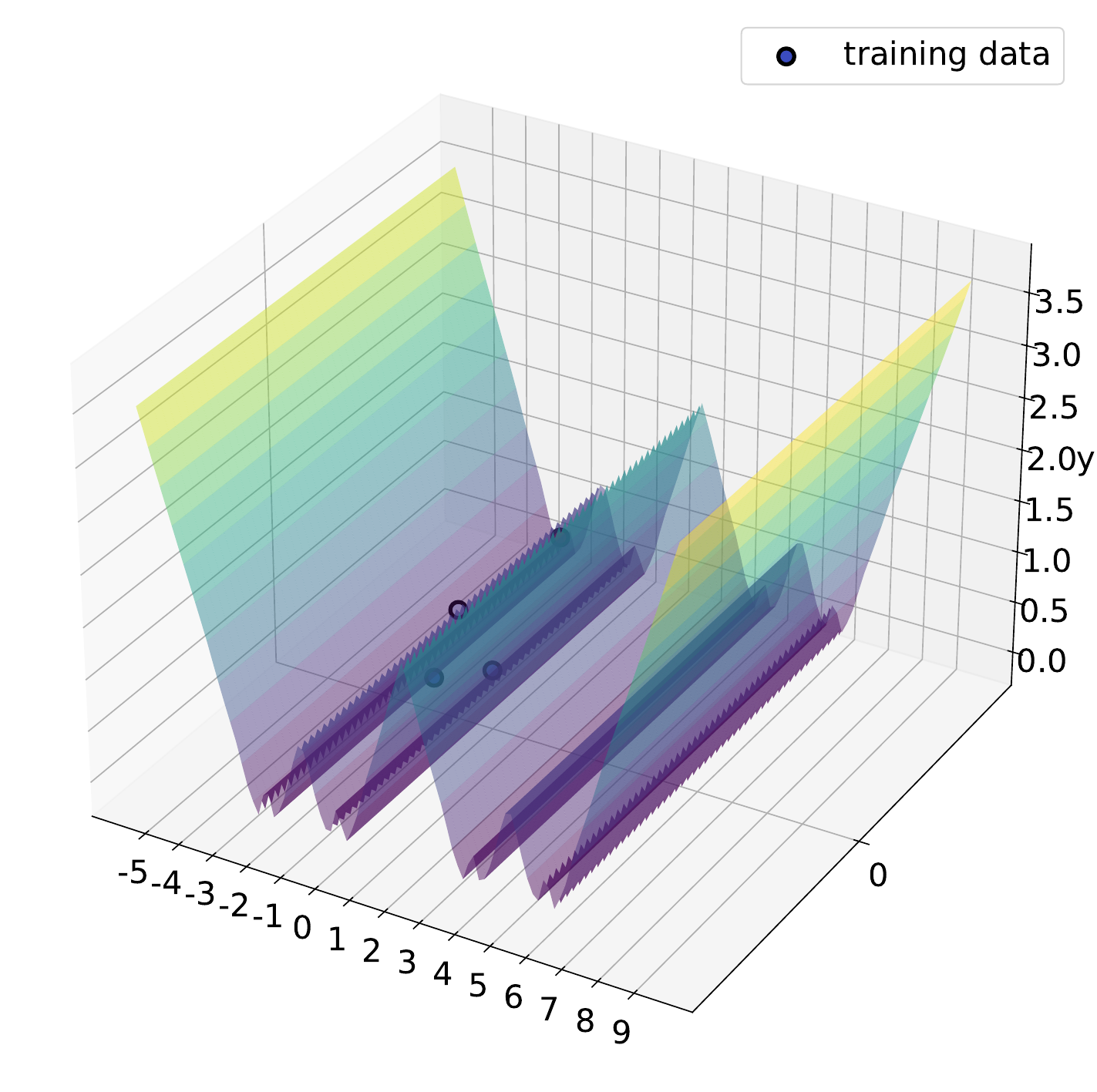}
    \end{minipage}
    \caption{\adam{}-trained neural net with absolute value activation and \nd{2} training data, of depth $3$ (left), $4$ (middle), and $5$ (right).
     There are \kink{}s at reflections of up to order $L{-}2$ of data points. The $3$ and $4$-layer networks match the corresponding features shown in the left and right-most plots of \Cref{fig:absvalfeat}, respectively.}
    \label{fig:2dabsval}
\end{figure}
\Cref{fig:3lyr_reflection_planes} illustrates an example of reflection planes $\regplane, \reflectplane$ in $\mathbb{R}^3$. The right plot of \Cref{fig:absvalfeat} graphs an example feature $\featurefunc(\mathbf{x})$ in $\mathbb{R}^2$. The feature outputs the distance between the reflection "planes" $\regplane, \reflectplane$, which are lines in $\mathbb{R}^2$ and are plotted as red, dashed lines. These are axes of symmetry encoded in $3$-layer features. 
In the notation used in \Cref{thrm:3layersimpler}, \Cref{fig:3lyr_reflection_planes} and the right plot of \Cref{fig:absvalfeat} both graph the case where $\xprime{=}\mathbf{x}_{1}, \mathbf{x}_{j_0}{=}\mathbf{x}_2,\mathbf{x}_{j_2}{=}\mathbf{x}_3,\mathbf{x}_{j_3}'{=}\mathbf{x}_1$, (and additionally $\mathbf{x}_{j_4}{=}\mathbf{x}_4,\mathbf{x}'_{5}{=}\mathbf{x}_1$ for \Cref{fig:3lyr_reflection_planes}). 

We can interpret reflection planes as \emph{concepts}. The concepts collect training data into planes, and the neural network measures the distance, or similarity, between inputs and concepts. The concepts are based at a training data point, and a point that is not necessarily in the training set: a reflection of a data point about another data point. In other words, the neural network predicts single-level reflective symmetries in the data. Intuitively, these symmetries can act as a translation between analogous concepts across different domains, with reflections representing shifts in structure while preserving core functionality. 
%A reflection translates the \kink{}, which represents an analogy.

Another way to view the geometry of a feature is by analysing its \kink{}s. A piecewise linear function $f:\mathbb{R} {\to} \mathbb{R}$ has a \emph{\kinkpoint{}} at $\mathbf{x}$ if $f$ changes slope at $\mathbf{x}$. A function $g:\mathbb{R}^{1\times d} \to \mathbb{R}$ defined by $g(\mathbf{x}){=}f(\mathbf{x}\mathbf{w})$ where $\mathbf{w}{\in}\mathbb{R}^d$ has a \emph{\kink{}} along the plane $\{\mathbf{x} \in \mathbb{R}^{1\times d}:\mathbf{x}\mathbf{w}{+}b{=}0\}$ if $f$ has a \kinkpoint{} at $\mathbf{xw}$.
%\end{definition}
%
A \kink{} of a function is a "kink" in its graph. As seen above, the reflection hyperplanes are \kink{}s of $3$-layer networks and create axes of symmetry.
The reflection \kink{}s in $3$-layer networks generalize to deeper networks, as shown next.

\subsection{Deeper networks}\label{section:deeper}

Here, we extend a \lasso{} equivalence to deeper networks.
%Let $ \wl{1} = \frac{{\mathbf{x}}_{j_2}-\mathbf{x}'_{j_{3}} \times \cdots \times {\mathbf{x}}_{j_{2(d-1)}}-\mathbf{x}'_{j_{2d-1}} }{\left \| {\mathbf{x}}_{j_2}-\mathbf{x}'_{j_{3}} \times \cdots \times {\mathbf{x}}_{j_{2(d-1)}}-\mathbf{x}'_{j_{2d-1}} \right \|_1}$ as defined in \Cref{thrm:3layersimpler}. 
For $L\geq 2$, the $L$-layer \textit{feature function} is a parameterized  function $\featurefunc:\mathbb{R}^d {\to} \mathbb{R}$ defined as
\begin{equation}\label{eq:feature}
    \featurefunc(\mathbf{x}) = \left | \rule{0cm}{0.8cm} \cdots \Bigg |  \bigg| \Big|\mathbf{x}\wl{1}+\sbl{1}\Big| + \sbl{2}\bigg |\cdots \Bigg | + \sbl{L-1} \right|.
\end{equation}
A feature function \eqref{eq:feature} can be viewed as a basic unit of a \reflectnn{} \eqref{eq:deepnarrowabsvalnetwork}, or a \reflectnn{}  where $m{=}1, \alpha{=}1, \xi{=}0$ and $\sWul{i}{l}{=}1$ for $l{>}1$. 
A feature function has \textit{data feature biases} if for $ n^{(1)}{=}\xprime, n^{(2)}, {\cdots}, n^{(L{-}1)} {\in} [N]$, all of its bias parameters are defined recursively as

%deeper other is not eactly this
 \begin{equation}\label{eq:optbias}
    \begin{aligned}
        \sbl{1} {=}-\mathbf{x}'_{j_1}\wl{1}, \quad %~
       % \sbl{2}&= -\left|\mathbf{x}_{n^{(2)}}\wl{1}+\sbl{1}\right| = -\left|\left(\mathbf{x}_{n^{(2)}}-\mathbf{a}\right)\wl{1}\right|\\
      %  &\vdots \\
         \sbl{l} {=} -\Bigg | \rule{0cm}{0.8cm} \cdots \bigg |  \Big| \big|\mathbf{x}_{n^{(l)}}\wl{1}+\sbl{1}\big| + \sbl{2}\Big |\cdots \bigg | + \sbl{l-1} \Bigg |,
    \end{aligned}
\end{equation}

for $l {\in} [L{-}1]$. The \emph{\sublibrary} is a set of feature functions with data feature biases. The next result states that \reflectnn{}s of \emph{any} depth are equivalent to \lasso{} problems.

\begin{theorem}\label{lemma:deep}
    The training problems for \reflectnn{}s with arbitrary depth and input dimension are equivalent to \lasso{} problems with finite, discrete dictionaries. The set of feature vectors contains a \sublibrary{} and consists of feature functions \eqref{eq:feature} sampled at training data points. An optimal network is reconstructed as $\sum_i z^*_i {\featurefunc}(\mathbf{x})+\xi^*$. 
    %where $\mathbf{z}^*, \xi^*$ are \lasso{} solutions, provided $m \geq \|\mathbf{z}^*\|_0$. 
\end{theorem}
%matrix M will be more complicated for deeper networks, so optimal network might not be orthogonal to same points as 3-layer
%
%The dictionary columns correspond to feature functions.
%where n1=j0
%
%but reconstruction depends on b without simplifying, need to know where b came from 
%The \lasso{} equivalence extends to deeper networks as well.
The proof is in \Cref{app:maintheorems}. 
\Cref{def:reconstruct} describes a mapping between features and optimal weights. Explicit expressions of features written as \eqref{eq:feature} for $2$ and $3$-layer networks is in \Cref{rk:explicit_features}. The \sublibrary{} is a set of novel features representing reflections of increasing complexity, or \emph{order}. The order of a reflection is defined recursively as follows.
%While there is no unique \lasso{} dictionary that gives a \lasso{} problem equivalent to the training problem, the data feature sub-library is an essential sub-dictionary that directly derives from the training problem's dual constraint and contain novel reflection features. %does u alway give reflections?
%
%Additionally, there is not a \lasso{} problem with a unique dictionary that is equivalent to the training problem. The dictionary can be expanded in a way without changing the optimal value (see Appendix). 
%
%\end{definition}
%The sub-library is a subset of the library ignoring the midpoint feature bias parameters.
%Neural networks learn features in the deep library. 
%
%Analyzing the complete library of networks with more than $3$ layers is an area of future work.
%As suggested above, the features contain important geometric information. They demonstrate \kink{}s at reflections of data points.
%defined next.
%\begin{definition}[\kink{}]
   %
%We generalize this to a sublibrary of deeper networks. 
%To analyze the \sublibrary{}, we generalize the definition of a reflection. The definition is recursive.
%
\begin{figure}
    \centering
    \begin{tikzpicture}[scale=0.6,declare function={
    f(\a,\b)=2*\b-\a;
     }]
    \pgfmathsetmacro{\ax}{4} 
    \pgfmathsetmacro{\ay}{1} 
    \pgfmathsetmacro{\bx}{1} 
    \pgfmathsetmacro{\by}{2} 
    \pgfmathsetmacro{\cx}{-3} 
    \pgfmathsetmacro{\cy}{1.5}  
    \pgfmathsetmacro{\rabx}{f(\ax,\bx)}
    \pgfmathsetmacro{\raby}{f(\ay,\by)}
    \pgfmathsetmacro{\rabrcx}{f(\rabx,\cx)}
    \pgfmathsetmacro{\rabrcy}{f(\raby,\cy)}
    \pgfmathsetmacro{\rcrabx}{f(\cx,\rabx)}
    \pgfmathsetmacro{\rcraby}{f(\cy,\raby)}
    \pgfmathsetmacro{\crabxmid}{(\cx+\rabx)/2}
    \pgfmathsetmacro{\crabymid}{(\cy+\raby)/2}
    \pgfmathsetmacro{\crabrcxmid}{(\cx+\rabrcx)/2}
    \pgfmathsetmacro{\crabrcymid}{(\cy+\rabrcy)/2}
    \pgfmathsetmacro{\rabrabrcxmid}{(\rabx+\rcrabx)/2}
    \pgfmathsetmacro{\rabrabrcymid}{(\raby+\rcraby)/2}
    \pgfmathsetmacro{\abxmid}{(\ax+\bx)/2}
    \pgfmathsetmacro{\abymid}{(\ay+\by)/2}
    \pgfmathsetmacro{\rabbxmid}{(\rabx+\bx)/2}
    \pgfmathsetmacro{\rabbymid}{(\raby+\by)/2}
    %r(a,b) = (-2,3)
    %abc
    \draw[->]        (0,0) -- (\ax,\ay) node[right, fill=white,label=right:$\mathbf{a}$] (a) {};
    \draw[->]        (0,0) -- (\bx,\by) node[above, fill=white,label=above:$\mathbf{b}$] (b){};
    \draw[->]        (0,0) -- (\cx,\cy) node[left, fill=white,label=left:$\mathbf{c}$] (c) {};
    %reflections
    \draw[blue,   ->]        (0,0) -- (\rabx,\raby) node[above, fill=white,label= left :$\reflect{\mathbf{a}}{\mathbf{b}}$] (Rab){};
    \draw[red, ->]        (0,0) -- (\rabrcx,\rabrcy) node[left,fill=white, label=left:$\reflect{\reflect{\mathbf{a}}{\mathbf{b}}}{\mathbf{c}}$] (Rabc) {};
    \draw[red, ->]        (0,0) -- (\rcrabx,\rcraby) node[above,fill=white,label=above:$\reflect{\mathbf{c}}{\reflect{\mathbf{a}}{\mathbf{b}}}$] (Rcab) {};
    %lines between
    \draw[dashed] (a) -- (Rab);
    \draw[dashed] (Rcab) -- (Rabc);
    %(0.5,0) node {$|$} (0.5,1) node {$|$}
    %(0,0.5) node[rotate=70] {$||$} (1,.5) node[rotate=70] {$||$}
    \draw (\crabxmid,\crabymid) node[left, rotate=60] {$|$};
    \draw (\crabrcxmid,\crabrcymid) node[rotate=60] {$|$};
    \draw (\rabrabrcxmid,\rabrabrcymid) node[rotate=60] {$|$};
    \draw (\abxmid,\abymid) node[right, rotate=-20]{$||$};
    \draw (\rabbxmid,\rabbymid) node[right, rotate=-20]{$||$};
    \end{tikzpicture}
    \caption{Reflections in $\mathbb{R}^2$ of different orders. Order $0$: $\mathbf{a},\mathbf{b},\mathbf{c}$. Order $1$: $ \reflect{\mathbf{a}}{\mathbf{b}}$. Order $2$: $ \reflect{\reflect{\mathbf{a}}{\mathbf{b}}}{\mathbf{c}}, \reflect{\mathbf{c}}{\reflect{\mathbf{a}}{\mathbf{b}}}$ .}
    \label{fig:reflections}
\end{figure}
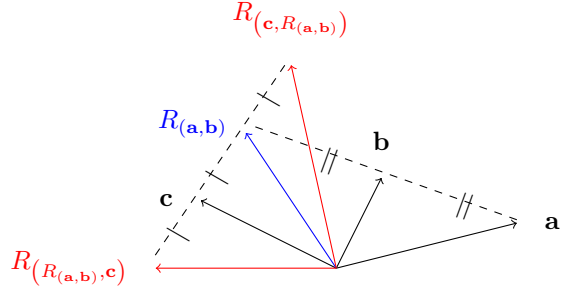
%
%\begin{definition}[higher order reflections]\label{def:reflections}
    An \emph{order-$0$} reflection of a point $\mathbf{x}_0 {\in} \mathbb{R}^d$ is simply $R_0\left(\mathbf{x}_0\right){=}\mathbf{x}_0$. Given points $\mathbf{x}_0,\mathbf{x}_1$, the \emph{standard reflections} $\reflect{\mathbf{x}_0}{\mathbf{x}_1}$ and $\reflect{\mathbf{x}_1}{\mathbf{x}_0}$ are \emph{order-$1$} reflections, which create a single-level symmetry of the point being reflected and its reflection. An \emph{order-$2$} reflection is a reflection of a reflection, which induces a \emph{second-level symmetry} around the reflection.
    %$\reflect{R_1}{\mathbf{x}_0}$ of a reflection $R_1$ about another point $x$, or a reflection $\reflect{x}{R_1}$ of a point $x$ about another reflection $R_1$. 
    In general for $k{>}0$, the \emph{order-$k$ reflection} of $\mathbf{x}_0, {\cdots}, \mathbf{x}_k {\in} \mathbb{R}^d$ is of the form $R_k(\mathbf{x}_0, {\cdots}, \mathbf{x}_k) {\in} \left\{ \reflect{R_{k{-}1}\left(\mathbf{x}_0, \cdots, \mathbf{x}_{k-1}\right)}{\mathbf{x}_{k}}, \reflect{\mathbf{x}_{k}}{R_{k{-}1}\left(\mathbf{x}_0, \cdots, \mathbf{x}_{k-1}\right)} \right\}$.
%\end{definition}
%
\Cref{fig:reflections} plots examples of reflections of order $0,1$ and $2$ in $\mathbb{R}^2$. As seen in  \Cref{fig:reflections}, in $\mathbb{R}^2$, $\reflect{\mathbf{a}}{\mathbf{b}}$ is the point on the line between $\mathbf{a}$ and $\mathbf{b}$, whose distance from $\mathbf{b}$ is the same as the distance from $\mathbf{b}$ to $\mathbf{a}$, creating a symmetry.
%Higher order reflections are found by continuing to "flip" points across the line connecting them.

%A $0$-order reflection will often not be called a reflection. 
%A first-order reflection may be called a standard reflection. 

%The features are exactly the feature vectors with $\mathbf{x}_i$ replaced by a general input variable $\mathbf{x}$. 
%For $3$-layer \reflectnn{}s, the features measure distances to parallel hyperplanes and
%$\xl{L}(\mathbf{x})$ 
%are w-shaped piecewise folded planes. 
%or $\mathbf{a} \in \mathcal{K}_{n^{(1)},n^{(2)}}$, $\xl{3}(\mathbf{x})=\left|\left|\left(\mathbf{x}-\mathbf{x}_{j_1}'\right)\wl{1} \right| - \left| \left(\mathbf{x}_{n^{(2)}}-\mathbf{x}_{j_1}'\right)\wl{1}\right| \right|$ which has \kink{}s when $\left(\mathbf{x}-\mathbf{x}_{j_1}'\right)\wl{1}=0$ or $\left(\mathbf{x}-\mathbf{x}_{n^{(2)}}\right)\wl{1}=0$, or $\left(\mathbf{x}-\reflect{\mathbf{x}_{n^{(2)}}}{\mathbf{x}_{j_1}'}\right)\wl{1}=0$.
\Cref{theorem:geometricInterp} states that the features for a $3$-layer network measure distances to reflection planes, which include \kink{}s at first-order reflections.
%These are parallel lines reflected across each other. 
The left plot of \Cref{fig:absvalfeat} illustrates an example of a $3$-layer feature. The weight $\wl{1}$ is orthogonal to $\mathbf{x}_{3}{-}\mathbf{x}_{1}$ and the \kink{}s occur along $\mathcal{H}_1{=}\left\{\left(\mathbf{x}{-}\mathbf{x}_{1}\right)\wl{1}{=}0\right\}, \regplane{=}\left\{\left(\mathbf{x}{-}\mathbf{x}_{2}\right)\wl{1}{=}0\right\}$ and $\reflectplane=\left\{ \left(\mathbf{x}{-}\reflect{x_{2}}{\mathbf{x}_{1}}\right)\wl{1}{=}0\right\}$. The height of the graph is the minimum of the distance from the reflection planes $\regplane$ and $\reflectplane$. Similarly, $4$-layer features can have \kink{}s that represent first or second-order reflections, as shown respectively in the middle and right plots of \Cref{fig:absvalfeat}, which plots examples of \sublibrary{} features. The peaks and troughs of the features plotted in \Cref{fig:absvalfeat} are \kink{}s. The troughs are indicated by dashed red lines and depict higher-level symmetries and reflection planes. Deeper networks have the capacity for learning increasingly complex reflection features and multilevel symmetries. %This increasingly complex reflections in \kink{}s generalizes for deeper layers. 
%We first informally state this result.

%The features measure distances to reflection planes located along \kink{}s.

\begin{theorem}[Multilevel symmetries]\label{absvalsublib}
    A \reflectnn{} of arbitrary input dimension and depth  $L$ is equivalent to a \lasso{} problem with a dictionary containing features with no reflections for $L=2$, standard reflections for $L{=}3$, and up to order-$2(L{-}3)$ reflections for $L>3$.
\end{theorem}

%An absolute value network is equivalent to a \lasso{} problem whose library is equal to the deep library for $L \in [3]$ and which contains the data feature sub-library for deeper $L$. 
%We formally describe the data feature sub-library appearing in deep \reflectnn{}s. 

%\begin{theorem}\label{absvalsublib}
%    The data feature sub-library of a $L$-layer, \deepnarrow{}, absolute value network contains features with reflections of order at most $L-2$ for $L\leq 4$ and $2(L-3)$ for $L\geq 4$.
%\end{theorem}
The proof is in \Cref{app:deep}. \Cref{absvalsublib} shows that each additional layer in a network induces higher order reflections and deeper levels of symmetry.  \Cref{absvalsublib} describes the \sublibrary{} in \Cref{lemma:deep} and shows that the maximum order of reflections in features grows linearly with depth.
A full analysis of the entire dictionary for networks with more than $3$ layers is an area for future work. Nonetheless, our approaches and proof techniques demonstrate the principle of how to find all of the dictionary elements for deeper networks, as discussed in the proof of \Cref{lemma:deep}.
%w^1 might change reflection a bit as a breakline, just note

%%%%%%alternative%%%%%%%%%%
%\begin{figure}%scale=2

%\caption{Example of a feature for \nd{2} data and $4$-layer %\reflectnn{}. 
%Observe that $\wl{1}$ is perpendicular to $\mathbf{x}_{j_3}-\mathbf{x}_{j_1}$ and is the direction in which the feature changes. 
%The \kink{}lines occur along $\left(\mathbf{x}-\mathbf{x}_{j_1}\right)\wl{1}=0, \left(\mathbf{x}-\mathbf{x}_{j_2}\right)\wl{1}=0$ and $ \left(\mathbf{x}-\reflect{x_{j_2}}{\mathbf{x}_{j_1}}\right)\wl{1}=0$. 
%Here, $\mathbf{a}=\mathbf{x}_{j_1}$.
%}
% \label{fig:absvalfeatL4alt}
%\end{figure}
\section{Numerical Results}\label{section:numresults}

We perform experiments training the \textit{standard} network
\begin{equation}\label{eq:standardnet}
    \nn{L}{\mathbf{x}} {=} \sigma\left({\cdots} \left(\sigma\left(\sigma\left(\mathbf{x}\wl{1}+\bl{1}\right)\wl{2} {+} \bl{2}\right)\cdots \right)\wl{L{-}1}{+} \bl{L{-}1} \right)\boldsymbol{\alpha} {+} \extbias.
\end{equation}

where $\wl{1} {\in} \mathbb{R}^{d \times 1}, \bl{1} {\in} \mathbb{R}, {\cdots}, \wl{l} {\in} \mathbb{R}, \bl{l} {\in} \mathbb{R}, \wl{L{-}1} {\in} \mathbb{R}^{1 {\times} m}, \bl{L{-}1} {\in} \mathbb{R}^{1 {\times} m}, \xi {\in} \mathbb{R}, \boldsymbol{\alpha} {\in} \mathbb{R}^m$. 
 For $2$-layer networks, the standard network \eqref{eq:standardnet} is equivalent to the network \eqref{eq:nn_nonrecursive}, and \eqref{eq:nn_nonrecursive} can be converted into a standard architecture \citep{1DNN}. The standard architecture is more traditional, and we perform experiments on the standard architecture to demonstrate that the \lasso{} model can be useful for this architecture as well. 

\subsection{Simulated Data}
In \Cref{fig:2dabsval}, $3,4,$ and $5$-layer networks \eqref{eq:standardnet} are trained with \adam{}. The second coordinate is $0$ in all samples. The data is given in \Cref{app:num_results}.
We first project the \nd{2} data to \nd{1} along the first coordinate  and solve the \lasso{} problem for the \nd{1} data as $\beta {\to} 0$.
The \textit{minimum ($l_1$) norm subject to interpolation} version of the \lasso{} problem is 
\begin{equation}\label{rm_gen_lasso_nobeta}
		\min_\mathbf{z, \xi} \|\mathbf{z}\|_1 \mbox { s.t. } \Amat{} \mathbf{z}+\xi \mathbf{1} = \mathbf{y}.
	\end{equation}
Loosely speaking, as $\beta{\to}0$, if $\Amat{}$ has full column rank, the \lasso{} problem approaches the minimum norm problem \eqref{rm_gen_lasso_nobeta}. For \nd{1} data, an optimal solution to \eqref{rm_gen_lasso_nobeta} for certain simple sets of training data is known \citep{1DNN}. After solving the \lasso{} problem, the \nc{} model \eqref{eq:trainingprob_explicit} with the orignal \nd{2} data is trained with $\beta{=}10^{-7} {\approx} 0$, where 
a minimum norm problem solution is used to pre-initialize a subset of the neurons. All other weights (we use $m{=}100$) are initialized randomly according to Pytorch defaults. Only one neuron per layer is pre-initialized, and for the neuron weight in the first layer, the second coordinate is left to random initialization.  
%The \adam{}-trained networks closely match the \lasso{} solutions. Moreover, the $5$-layer networks suggest that features continue to gain more complex reflections with depth.
 A learning rate of $5(10^{-3})$ and weight decay of $10^{-4}$ are used. The data is trained over $10^3$ epochs. A similar experiment is performed with a ReLU network, which also demonstrates \kink{}s at reflections of training data (\Cref{app:num_results} \Cref{fig:relu}).
%When $\beta{\to}0$, the \lasso{} problem approaches the  minimum norm problem \eqref{rm_gen_lasso_nobeta}. 
%For each $L {\in} {3,4,5}$, the neural net reconstructed from the single feature in the left plot with corresponding \lasso{} parameter $z^*_i=1$ and $\xi^*=0$ is optimal in the minimum norm problem \eqref{rm_gen_lasso_nobeta} by \Cref{lemma:minNormCert} and \Cref{lemma:minNormCert2neurons}. 
%The figures show that the networks trained with the \nc{} problem closely match the \lasso{} solutions. 
As shown in \Cref{fig:2dabsval}, the networks exhibit \kink{}s at data points and their reflections which are not among the training data.
%The standard architecture in the \nc{} model shows the applicability of the \lasso{} formulation. 
Additionally, the neural networks exhibit features matching the shapes of those shown in the right and left plots of \Cref{fig:absvalfeat} for $3$ and $4$ layers, respectively. The $5$-layer network shows that features gain more complex reflections with depth.
%SGD gives similar results as \adam{}. 

\subsection{Large Language Model Embeddings}\label{section:LLM}
We also train neural networks using Adam to perform binary $(y_n{=}\pm1)$ classification of text and observe multi-level symmetries. Models from OpenAI GPT4 and Bert are trained to embed text as vectors, which \reflectnn{}s are then trained on as input to classify the corresponding text. The \reflectnn{}s are trained using the \nc{} regression problem \eqref{eq:trainingprob_explicit}. Positive network outputs are interpreted as $y_n{=}1$ and negative outputs as $y_n{=}{-}1$.

\begin{figure}[t]
    \centering
    \begin{minipage}[t]{0.25\textwidth}
     \includegraphics[width=\linewidth,]{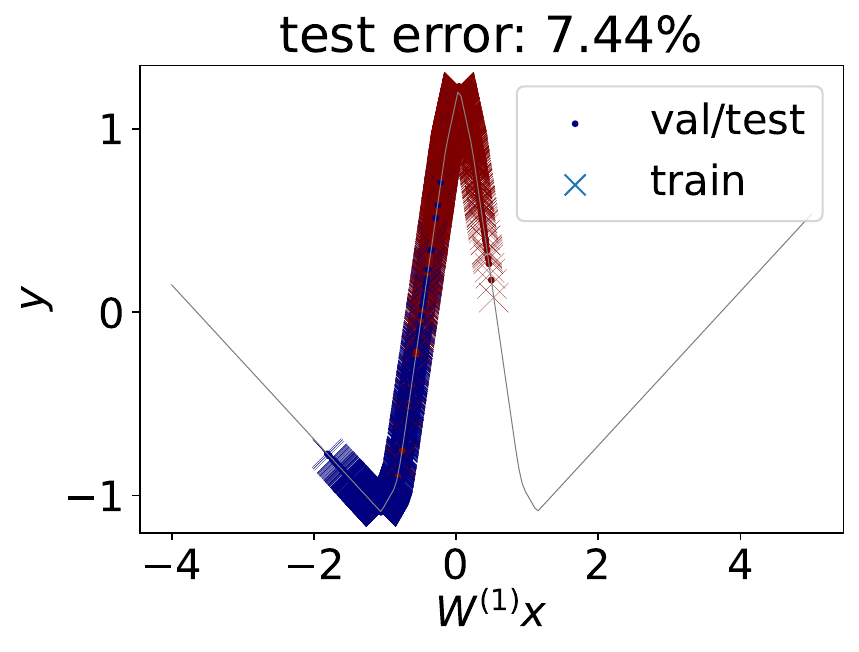}
    \end{minipage}
    \begin{minipage}[t]{0.24\textwidth}
    \includegraphics[width=\linewidth,trim= 40 0 0 0,clip]{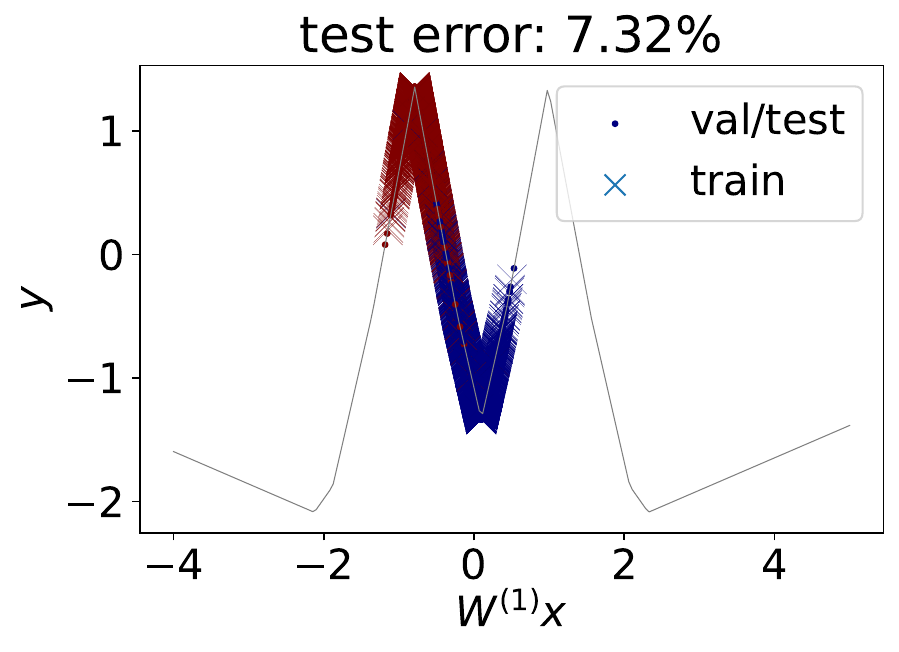}
    \end{minipage}
    \begin{minipage}[t]{0.24\textwidth}
    \includegraphics[width=\linewidth,trim= 40 0 0 0,clip]{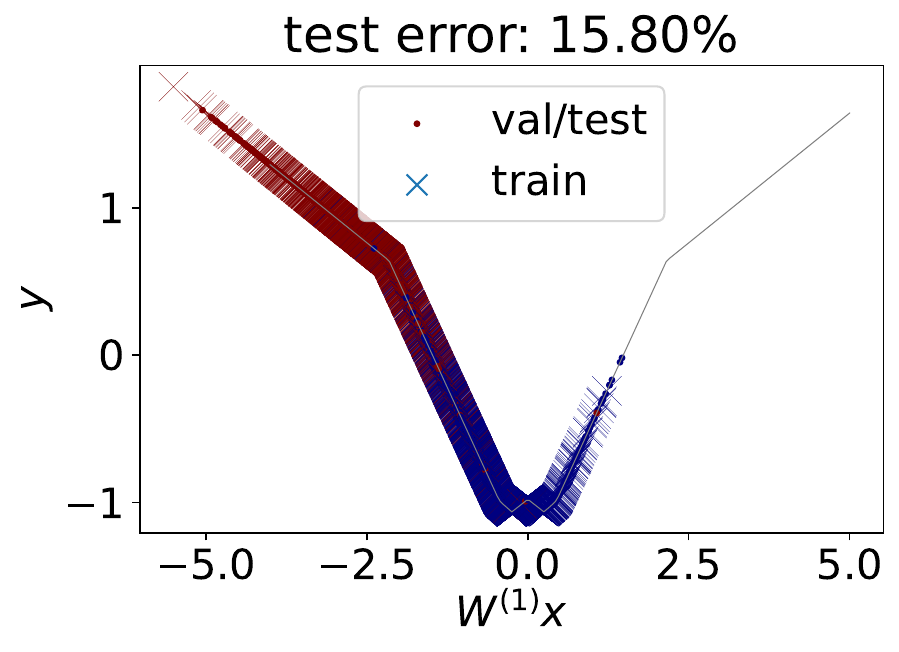}
    \end{minipage}
    \begin{minipage}[t]{0.24\textwidth}
    \includegraphics[width=\linewidth,trim= 40 0 0 0,clip]{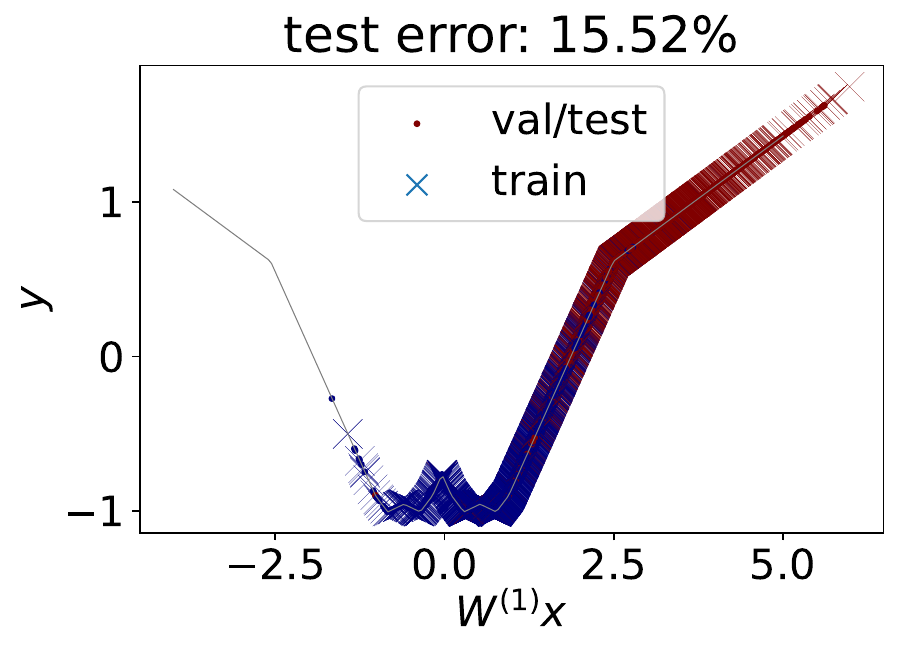}
    \end{minipage}
    \caption{$3$-layer, followed by $4$-layer (right)  \reflectnn{}s trained on text embeddings of IMDB reviews outputted by OpenAI GPT4 (left-most $2$ plots) and Bert (right-most $2$ plots). %Bert IMDB
 The network's predictions are plotted as a function of the input $\mathbf{x}$ projected to \nd{1} as $\mathbf{W}^{(1)}\mathbf{x}$. True labels of $1$ and ${-}1$ are indicated by red and blue, respectively. The networks exhibit multi-level symmetry.}
    \label{fig:imdb_openaiL=3absval}
\end{figure}

\Cref{fig:imdb_openaiL=3absval} plots predictions of neural nets that are trained to perform sentiment analysis, rating IMDB reviews as having positive or negative sentiments \citep{IMDB}. \Cref{fig:imdb_openaiL=3absval} plots the points $\left(\mathbf{W}^{(1)}\mathbf{x}_n,\hat{y}_n\right)$ in red and blue, where $\hat{y}_n{=}\nn{L}{\mathbf{x}_n}$ is the network's prediction on the training sample $\mathbf{x}_n$ and the color corresponds to the training labels as red for $y_n{=}1$ and blue for $y_n{=}{-}1$. The network prediction on all points $\mathbf{x}$ is plotted in gray. For any constant $c{\in}\mathbb{R}$, a \lasso{} feature \eqref{eq:feature} has constant value along $\{\mathbf{x}:\wl{1}\mathbf{x}{=}c\}$, so projecting the input along $\wl{1}$ can be viewed as a cross-section of the feature. 

In general, there are (possibly sub-optimal) weights and biases that can make a \reflectnn{} be an asymmetrical function with many \kink{}s that do not contain training samples. However, the trained networks in \Cref{fig:imdb_openaiL=3absval} appear multilevel symmetric with \kink{}s at data samples and resembles the shapes of the $3$ and $4$-layer reflection features illustrated in \Cref{fig:absvalfeat}. In particular, the $3$-layer networks shown in the first and third plots from the left in  \Cref{fig:imdb_openaiL=3absval} resemble the $3$-layer network in left plot of \Cref{fig:absvalfeat}, while the $4$-layer network shown in the second and fourth plots in \Cref{fig:imdb_openaiL=3absval} resemble the $4$-layer network shown in the middle and right plots, respectively, of \Cref{fig:absvalfeat}. This is consistent with the sparse selection of features in the \lasso{} problem. \Cref{app:num_results} contains training details and additional results. %THe errors are...
Code from the github repository for \cite{Wang:2024} was used to generate the embeddings.
%We used a standard architecture:

\section{Conclusion}
We prove an equivalence between neural networks and \lasso{} problems with novel geometric features. 
Our convexification approach can be extended to other piecewise linear activation functions \cite{1DNN}. 
%While the absolute value activation learns reflection features with just a simple architecture, \cite{1DNN} provides methods for larger architectures with \nd{1} input and ReLU to learn reflections, and these can be similarly adopted to higher dimensional input data. 
A limitation of this work is the choice of the activation function and the $\ell_1$ regularization of the weights needed for analytic tractability. We believe that multi-level symmetries hold for standard ReLU networks, which is left for future work.

\newpage

\iftoggle{arxiv}{
\section{Acknowledgments}
This work was supported in part by National Science Foundation (NSF) under Grant DMS-2134248; in part by the NSF CAREER Award under Grant CCF-2236829; in part by the U.S. Army Research Office Early Career Award under Grant W911NF-21-1-0242; in part by the Office of Naval Research under Grant N00014-24-1-2164, in part by the National Science Foundation Graduate Research Fellowship under Grant No. DGE-1656518.
}

\iftoggle{iclr}{
\section{Reproducibility Statement}

Proofs of results proved in prior work:

\begin{itemize}
     \item \Cref{thrm:2layer1D}: proof in \citep{1DNN}\\
    \item \Cref{thrm:2layer0bias}: proof in \citep{pilanci2023complexity}
\end{itemize}

Proof of results proved in this work:

\begin{itemize}
    \item \Cref{thrm:3layersimpler}: proof in \Cref{proof_3lyrsimpler}\\
    \item \Cref{lemma:Asize}: proof in \Cref{proof_of_lemmaAsize}\\
    \item \Cref{theorem:geometricInterp}: proof in \Cref{proof:thrm_geom_interp}\\
    \item \Cref{lemma:deep}: proof in \Cref{proof_lemmadeep}\\
    \item \Cref{absvalsublib}: proof in \Cref{proof:absvalsublib}
\end{itemize}

Code is in the supplementary file. Please run "run.ipynb" to generate figures referenced in \Cref{section:numresults} and \Cref{app:num_results}.
}
%A $l_2$ loss function is used to quantify the error of the neural network. 

%neural tangent \cite{pilanci2020neural}.

%%%%%%%%%%%
\newpage
\bibliography{references} % Entries are in the refs.bib file
\bibliographystyle{iclr2025_conference}

\clearpage
\newpage
\appendix
\section{Appendix}

\section{Geometric Algebra}\label{section:geomalgebra}
\textbf{Wedge and inner product}:
A dot product of multivectors, unlike a dot product of vectors, can have multivector outputs. The \textit{dot} or \textit{inner product} of arbitrary basis blades $E_1 {=} \prod_{i \in \mathcal{G}^{(1)}}e_{i}$ and $E_2{=}\prod_{i \in \mathcal{G}^{(2)}}e_{i}$ is $E_1 {\cdot} E_2 {=} E_1 E_2$ if $g(E_1 E_2){=}|g(E_1){-}g(E_2)|$, and $0$ otherwise. In other words, if  $G^{(1)} {\subset} G^{(2)}$, then $E_1 E_2$ is composed of basis blades in $\mathcal{G}^{(2)}{-}\mathcal{G}^{(1)}$, and $E_1 E_2{=}0$ otherwise. (Similarly if $G^{(2)} {\subset} G^{(1)}$).
The \textit{wedge} or \textit{outer product} is defined by $E_1 {\wedge} E_2 {=} E_1 E_2$ if $E_1 E_2$ has grade $g(E_1 E_2) {=} g(E_1) +g(E_2)$, and $0$ otherwise. So if $\mathcal{G}^{(i)} {\cap} \mathcal{G}^{(j)}{=} \emptyset$, then $E_1 E_2$ is composed of basis blades in $\mathcal{G}^{(1)} {\cup} \mathcal{G}^{(2)}$, and otherwise, $E_1 E_2{=}0$. 
%it is plus or minus the geometric product of basis vectors.
The dot and wedge product extend to multivectors linearly.  A \textit{vector} is a special case of a multivector 
$\mathbf{v}{\in} \mathbb{R}^d {\subset} \mathbb{G}^d$ of grade $1$. For more detailed background, see \cite{perwass:2009}.

Note that linearly distributing terms and applying $e_i e_i =1$ shows that $\wedge_{i=1}^{d-1} \mathbf{v}_i=\wedge_{i=1}^{d-1} \sum_{j=1}^{d} {v_{i}}_j e_j $ consists of a sum of $d-1$-vectors, so their product with $e_d \cdots e_1$, as implied by the Hodge star operator in \eqref{eq:cross_wedge} is a vector, which is consistent with \eqref{eq:gencrosprod}.

\begin{remark}\label{rk:feature_as_wedge}
    It holds that $\star \left(\mathbf{u} \wedge \left( \wedge_{i=1}^{d-1} \mathbf{v}_i \right) \right) = \mathbf{u}^T \left(\times_{i=1}^{d-1} \mathbf{v}_i \right)\in \mathbb{R}$ and $\|\wedge_{i=1}^{d-1}\mathbf{v}_i\|_2 = \left \|\times_{i=1}^{d-1}\mathbf{v}_i \right\|_2$. This is because

%First, we write the feature function in terms of the Hodge dual and wedge products (\Cref{section:geoalgebra}).

\begin{equation}\label{eq:wedge_cross}
   \begin{aligned}
        \wedge_{i=1}^{d-1}\mathbf{v}_i &= \sum_{k=1}^d \left(\left( \wedge_{i=1}^{d-1} \mathbf{v}_i \right)^T \left(e_1 \cdots e_{k-1}e_{k+1}\cdots e_d  \right)\right) e_1 \cdots e_{k-1}e_{k+1}\cdots e_d \\
        &= \sum_{k=1}^d \left( \times_{i=1}^{d-1} \mathbf{v}_i \right)_k  e_1 \cdots e_{k-1}e_{k+1}\cdots e_d 
   \end{aligned}
\end{equation}

and $e_j {\wedge} e_k {=} 0$ if $j {\neq} k$, 
we have $\mathbf{u} \wedge \left( \wedge_{i{=}1}^{d{-}1} \mathbf{v}_i \right)  
{=}   \sum_{j{=}1}^d u_j e_j  \wedge \left( \wedge_{i{=}1}^{d-1} \mathbf{v}_i \right) 
{=}  \sum_{j=1}^d u_j e_j \wedge  \left(\left( \times_{i{=}1}^{d-1} \mathbf{v}_i \right)_k \right) e_1 {\cdots} e_{j{-}1}e_{j{+}1}{\cdots} e_d $. So $\star \left(\mathbf{u} \wedge \left( \wedge_{i=1}^{d-1} \mathbf{v}_i \right) \right) = \mathbf{u} \wedge \left( \wedge_{i=1}^{d-1} \mathbf{v}_i \right) e_d \cdots e_1 = \mathbf{u}^T \left(\times_{i=1}^{d-1} \mathbf{v}_i \right)\in \mathbb{R}$. From \eqref{eq:wedge_cross} it also follows that $\|\wedge_{i=1}^{d-1}\mathbf{v}_i\|_2 = \left \|\times_{i=1}^{d-1}\mathbf{v}_i \right\|_2$. 

\end{remark}
%Therefore,

%%todo be consistent with row vs column vector for x's
%\begin{equation}\label{eq:feature_wedge}
%    \begin{aligned}
%        f(\mathbf{x}) &=\lratiopenalty\left(\times_{i=1}^{d-1} \Tilde{\mathbf{x}}_i\right) \left| \left| \frac{\star((\mathbf{x}-\mathbf{a}) \wedge \tilde{\mathbf{x}}_1 \wedge \cdots \tilde{\mathbf{x}}_{d-1} )}{\|\wedge_{i=1}^{d-1} \tilde{\mathbf{x}}_i \|_2} \right|- \left| \frac{\star((\mathbf{x}_{n^{(2)}}-\mathbf{a}) \wedge \tilde{\mathbf{x}}_1 \wedge \cdots \tilde{\mathbf{x}}_{d-1} )}{\|\wedge_{i=1}^{d-1} \tilde{\mathbf{x}}_i \|_2} \right| \right|.
%    \end{aligned}
%\end{equation}

%\Cref{eq:feature_wedge} shows that neural networks can be expressed by leveraging operations from geometric algebra. The sparsity factor makes the feature function at every point larger in magnitude when $\wl{1} \propto \times_{i=1}^{d-1}\tilde{\mathbf{x}}_i$ is more sparse, making it more likely to be selected in the \lasso{} problem as a feature (compared to feature functions with similar \kink{}s but smaller magnitudes). Moreover, \eqref{eq:feature_wedge} has a geometric interpretation. 

\begin{remark}\label{rk:distance_volume}
It holds that
\begin{equation}\label{eq:volume}
  \left| \star\left( \mathbf{v}_1 \wedge \cdots \wedge \mathbf{v}_d\right) \right|= \mathrm{Vol}(\mathbf{v}_1,\cdots,\mathbf{v}_d)
\end{equation}

and

\begin{equation}\label{eq:distance}
    \text{Dist}\left(\mathbf{v}_d,\text{Span}(\mathbf{v}_1,\cdots,\mathbf{v}_d)\right)= \frac{\star(\mathbf{v}_1 \wedge \cdots, \wedge \mathbf{v}_d)}{\|\mathbf{v}_1 \wedge \cdots, \wedge \mathbf{v}_d\|_2}.
\end{equation}
    Here, $\mathbf{v}_1,\mathbf{v}_2$, let $\text{Dist}(\mathbf{v}_1, \mathbf{v}_2)$ denote the unsigned Euclidean distance $\|\mathbf{v}_1-\mathbf{v}_2\|_2$ between $\mathbf{v}_1$ and $\mathbf{v}_2$. The distance between a vector $\mathbf{v}$ and a set $S$ is $\min\{\text{Dist}(\mathbf{v},\mathbf{x}):\mathbf{x} \in S\}$. 
   % The wedge product $\mathbf{v}_1 \wedge \cdots, \wedge \mathbf{v}_d$ is a pseudoscalar whose signed magnitude is the signed volume of the $d$-dimensional polytope spanned by $\mathbf{v}_1, \cdots \mathbf{v}_d \in \mathbb{R}^d$, 
    In other words, $\mathbf{v}_1 \wedge \cdots \wedge \mathbf{v}_{d-1}$ is a vector whose magnitude is the unsigned volume of the $d-1$-dimensional parallelotope spanned by $\mathbf{v}_1, \cdots \mathbf{v}_{d-1}$, which represents the "base" of the $d$-dimensional parallelotope $\mathcal{P}(\mathbf{v}_1,\cdots,\mathbf{v}_d)$ and \eqref{eq:distance} is the signed 
"height" of $\mathcal{P}(\mathbf{v}_1,\cdots,\mathbf{v}_d)$, or the distance of $\mathbf{v}_d$ to the span of $\mathbf{v}_1, \cdots \mathbf{v}_d$.
. Note the Hodge star operator $\star$ converts pseudoscalars into a scalars.
\end{remark}

\section{Numerical results}\label{app:num_results}
\subsection{Simulated data}
In \Cref{fig:2dabsval},
 the training data is $(\mathbf{x}_1,y_1{=}2),(\mathbf{x}_2,y_2{=}0)$ for $3$ layers, $(\mathbf{x}_1,y_1{=}2),(\mathbf{x}_2,y_2{=}0),(\mathbf{x}_3,y_3{=}{-}1)$ for $4$ layers, and $(\mathbf{x}_1,y_1{=}1.75),(\mathbf{x}_2,y_2{=}0.25),(\mathbf{x}_3,y_3{=}0.75),(\mathbf{x}_4,y_4{=}{-}1.75)$ for $5$ layers, where $\mathbf{x}_1{=}(2,0),\mathbf{x}_2{=}(0,0),\mathbf{x}_3{=}(-1,0),\mathbf{x}_4{=}(-1.75,0)$.
 
\begin{figure}
    \centering
    \includegraphics[width=0.5\linewidth]{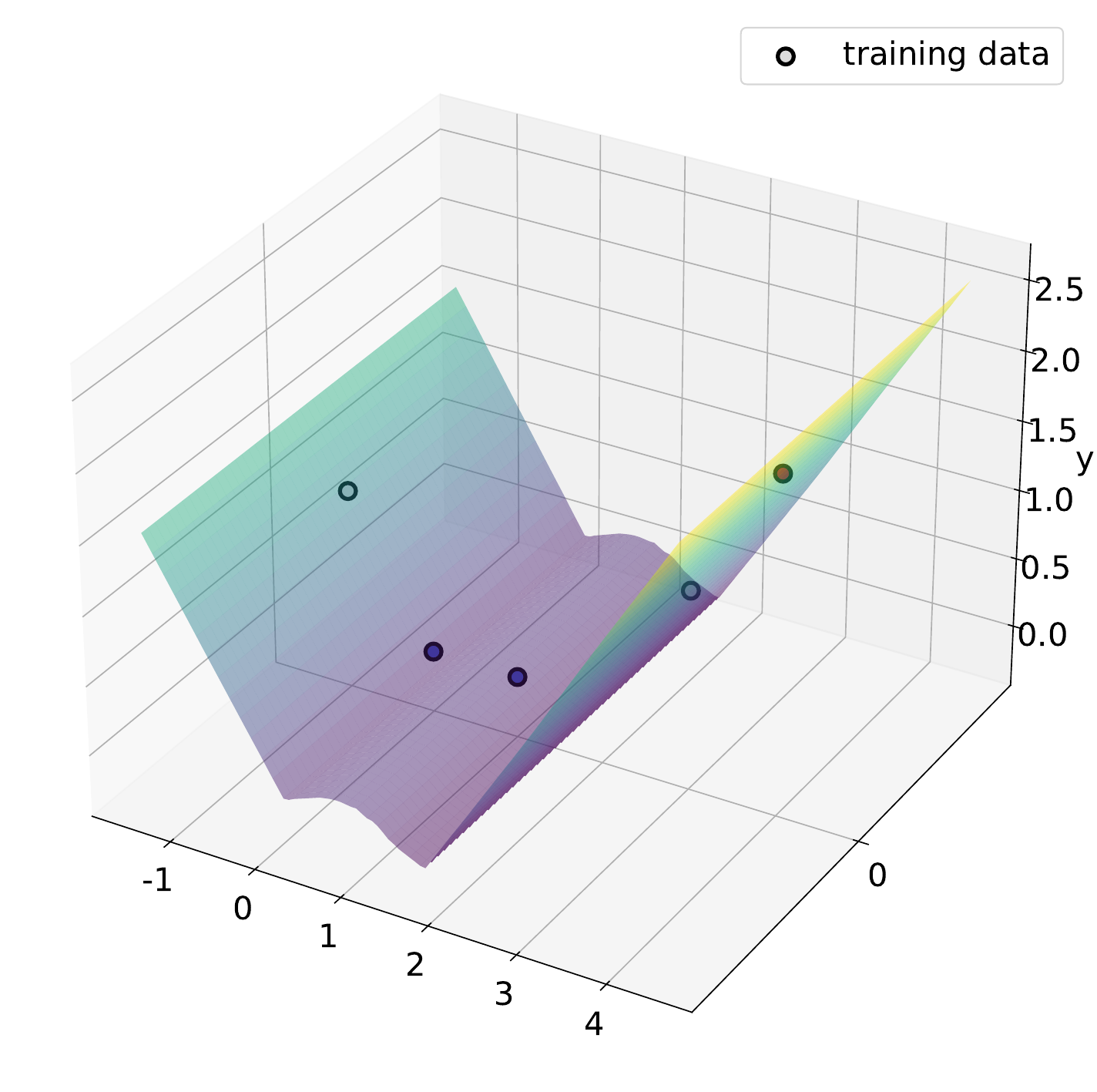}
    \caption{$3$-layer ReLU neural net. A \kink{} occurs at a reflection of training data, which is not in the training set.}
    \label{fig:relu}
\end{figure}

In \Cref{fig:relu}, the same experiment setup as used for \Cref{fig:2dabsval} is used for a $3$-layer network, except 
$\wl{1} {\in} \mathbb{R}^{d \times 2}, \bl{2} {\in} \mathbb{R}, \wl{2} {\in} \mathbb{R}^{2 {\times} m}, \bl{L{-}1} {\in} \mathbb{R}^{2 {\times} m}$ and we require all neuron weights in each layer have the same magnitude.  The data is $\mathbf{x}_1=(4,0), \mathbf{x}_2=(3,0),\mathbf{x}_3 = (1,0), \mathbf{x}_4 = (0,0), \mathbf{x}_5=(-1,0), y_1=2,y_2=1,y_3=0,y_4=0,y_5=1$. More general architectures are an area of exploration in future work. 

\subsection{Large Language Model data}
\begin{figure}%OpenAI IMDB. 13204
    \centering
    %%%%%seed1
   \begin{minipage}[t]{0.4\textwidth}
    \includegraphics[width=\linewidth]{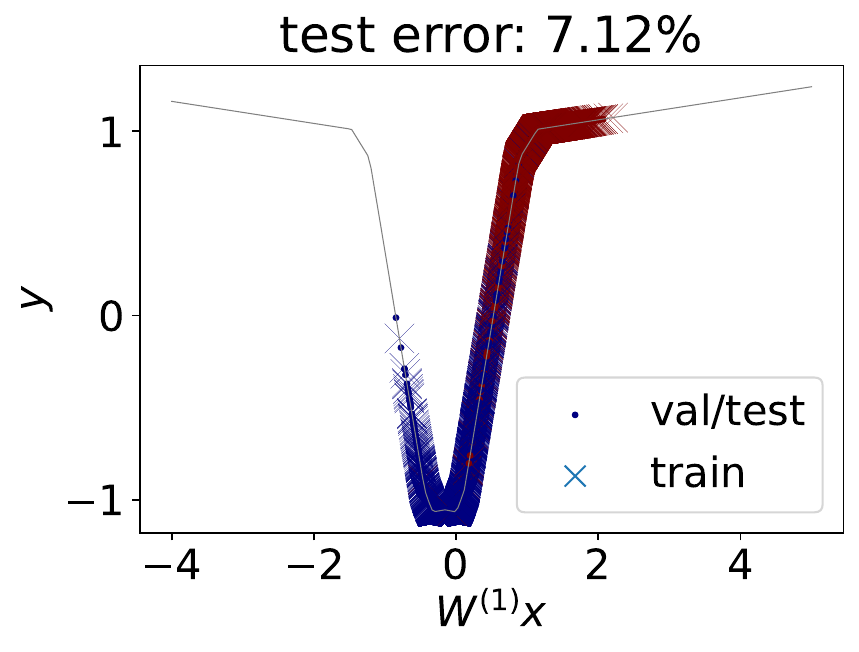}
    \end{minipage}
    \begin{minipage}[t]{0.4\textwidth}
    \includegraphics[width=\linewidth]{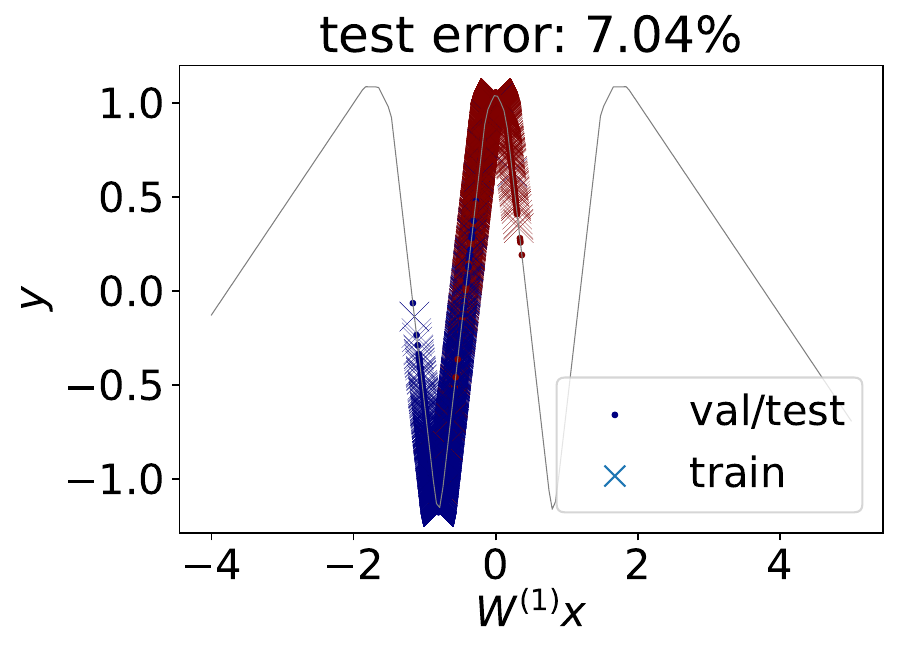}
    \end{minipage}
    %%%%%%%%seed3
   \begin{minipage}[t]{0.4\textwidth}
    \includegraphics[width=\linewidth]{results2/IMDB/OpenAI/L=3/seed3/W1plot.pdf}
    \end{minipage}
    \begin{minipage}[t]{0.4\textwidth}
    \includegraphics[width=\linewidth]{results2/IMDB/OpenAI/L=4/seed3/W1plot.pdf}
    \end{minipage}
    %%%%%%%%seed2
   \begin{minipage}[t]{0.4\textwidth}
    \includegraphics[width=\linewidth]{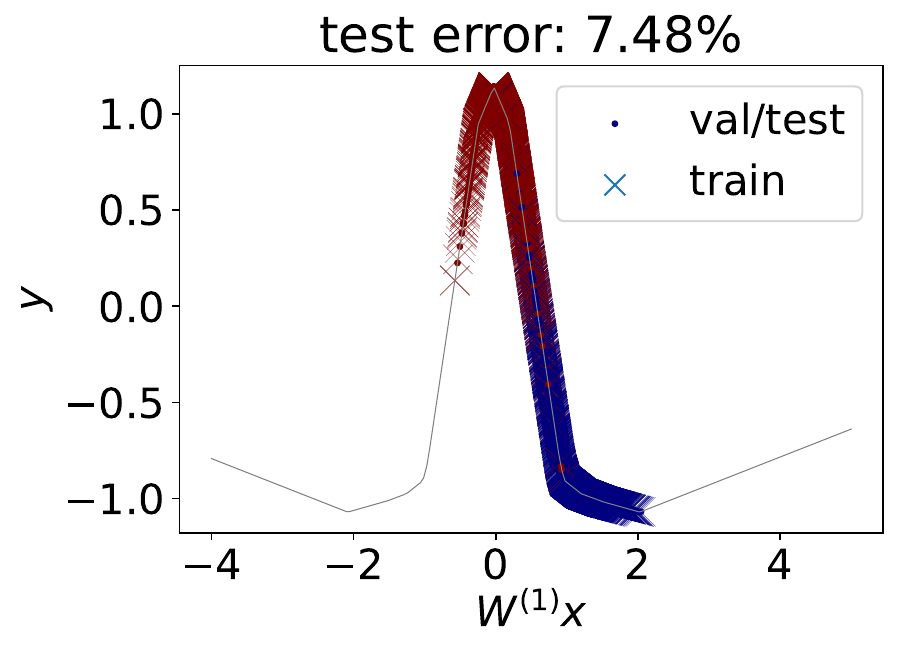}
    \end{minipage}
    \begin{minipage}[t]{0.4\textwidth}
    \includegraphics[width=\linewidth]{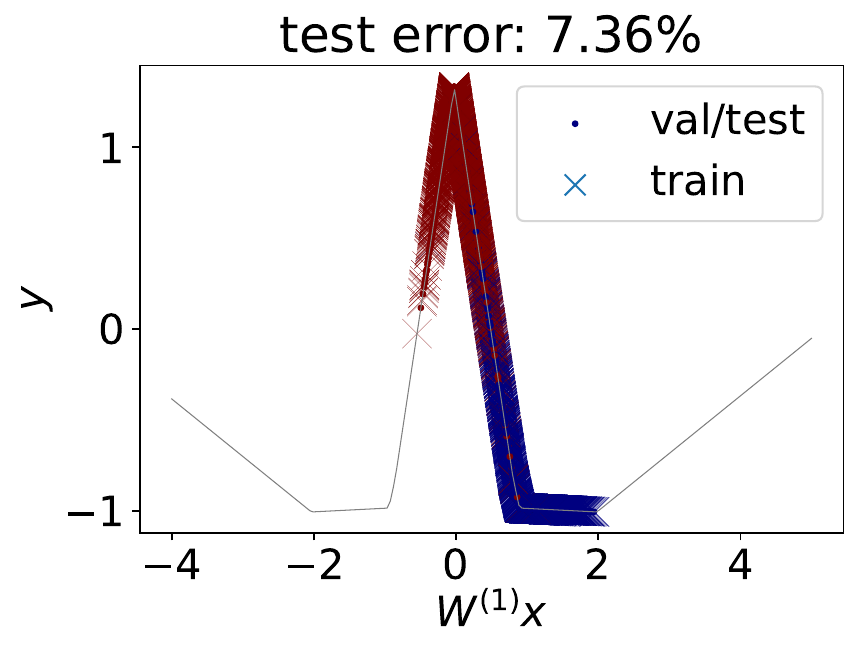}
    \end{minipage}
    %%%%%seed0
    \begin{minipage}[t]{0.4\textwidth}
    \includegraphics[width=\linewidth]{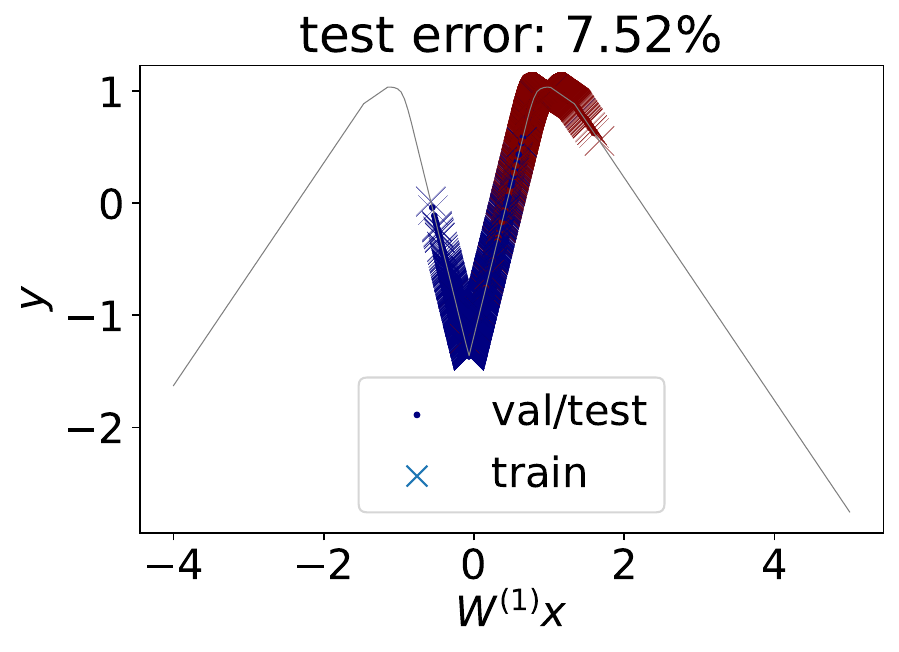}
    \end{minipage}
    \begin{minipage}[t]{0.4\textwidth}
    \includegraphics[width=\linewidth]{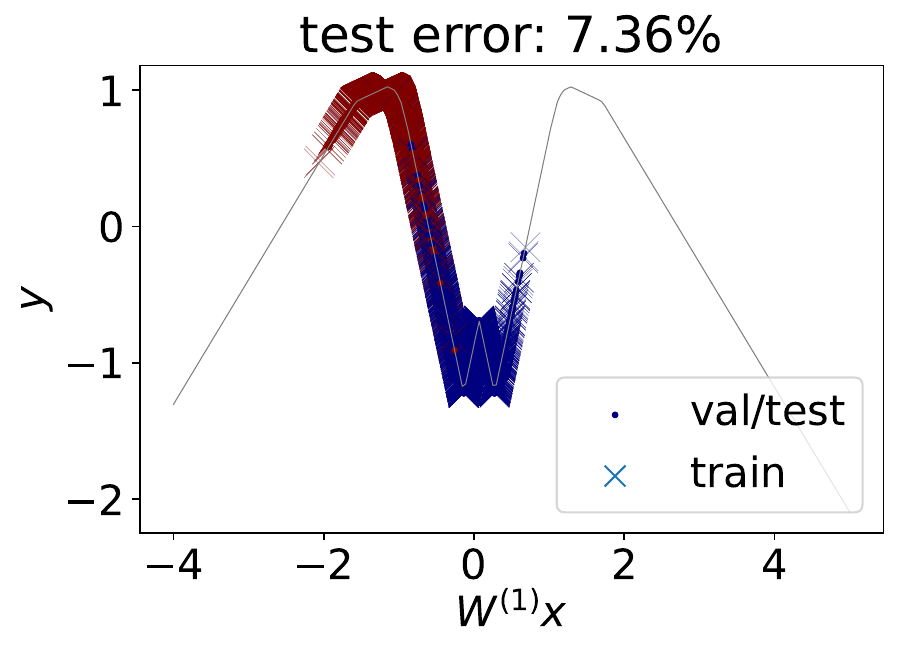}
    \end{minipage}
    %%%%%%%%seed4
   \begin{minipage}[t]{0.4\textwidth}
    \includegraphics[width=\linewidth]{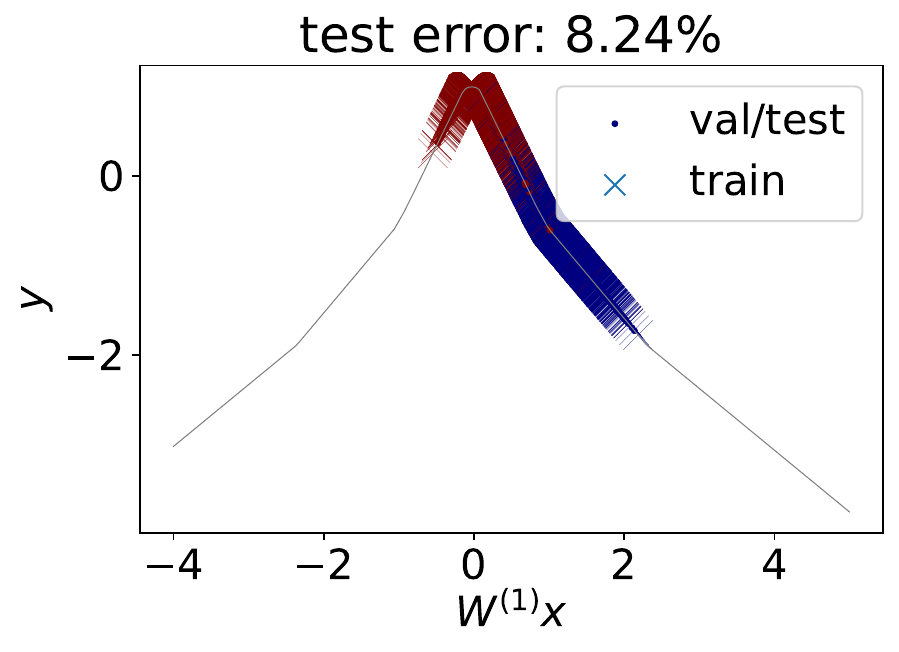}
    \end{minipage}
    \begin{minipage}[t]{0.4\textwidth}
    \includegraphics[width=\linewidth]{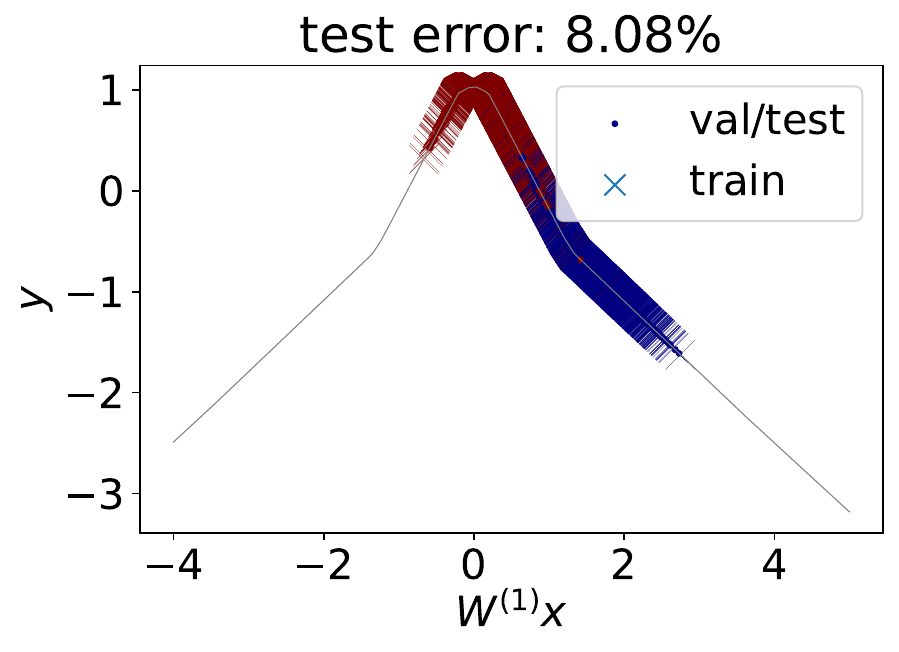}
    \end{minipage}

    \caption{$3$-layer (left column) and $4$-layer (right column) \reflectnn{}s trained on OpenAI embeddings of IMDB reviews. Each row is a different training initialization.}
    \label{fig:3lyrOpenAIIMDB}
\end{figure}

\begin{figure}%BertIMDB 42013
    \centering
    %%%%%%%%seed4
   \begin{minipage}[t]{0.4\textwidth}
    \includegraphics[width=\linewidth]{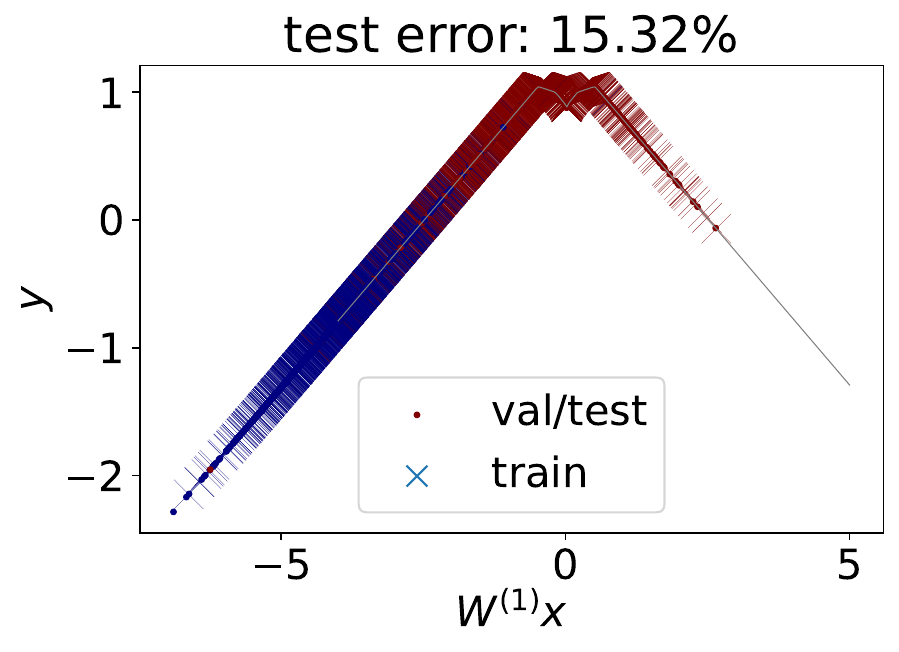}
    \end{minipage}
    \begin{minipage}[t]{0.4\textwidth}
    \includegraphics[width=\linewidth]{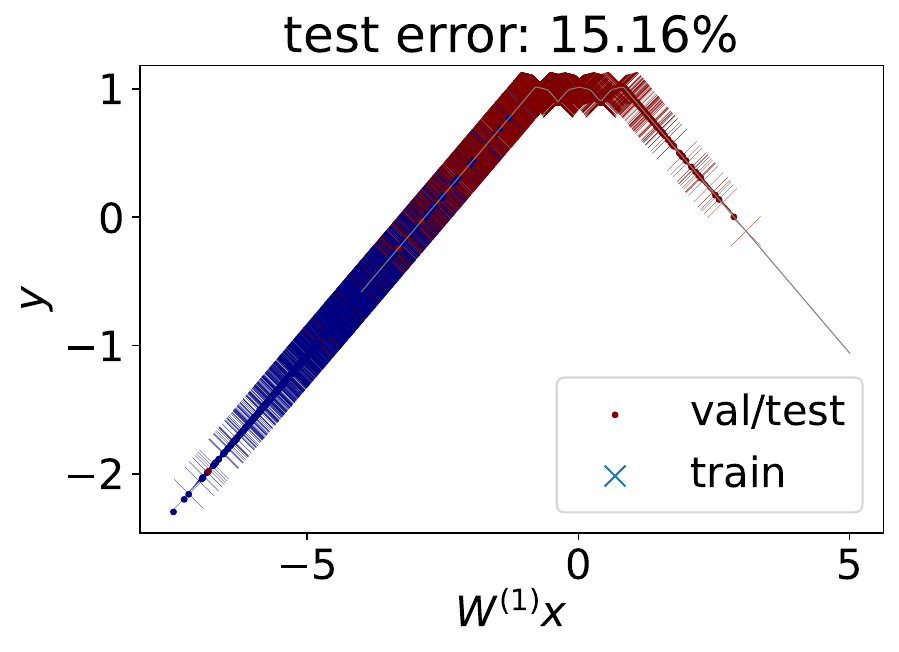}
    \end{minipage}
    %%%%%%%%seed2
   \begin{minipage}[t]{0.4\textwidth}
    \includegraphics[width=\linewidth]{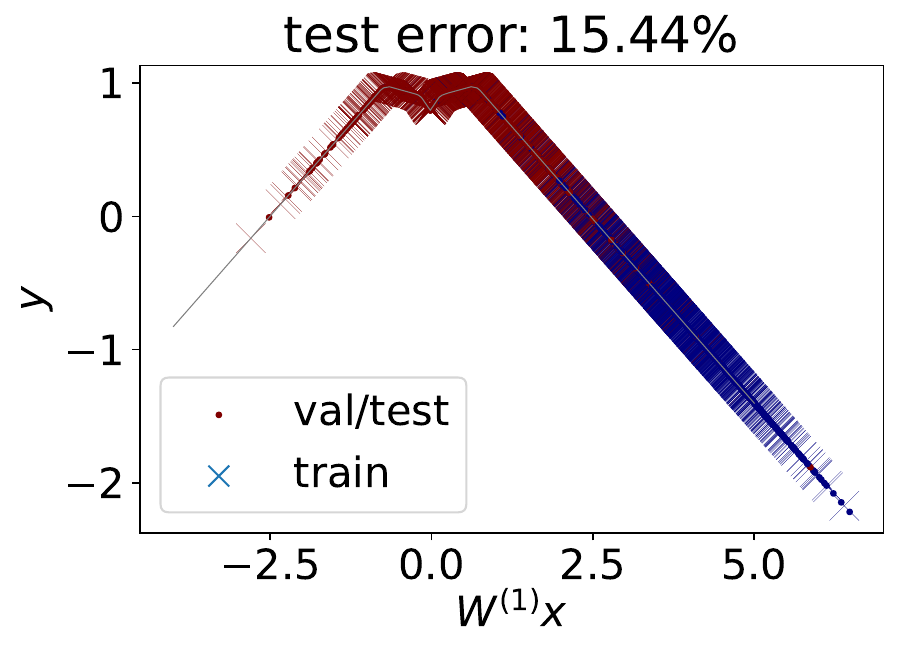}
    \end{minipage}
    \begin{minipage}[t]{0.4\textwidth}
    \includegraphics[width=\linewidth]{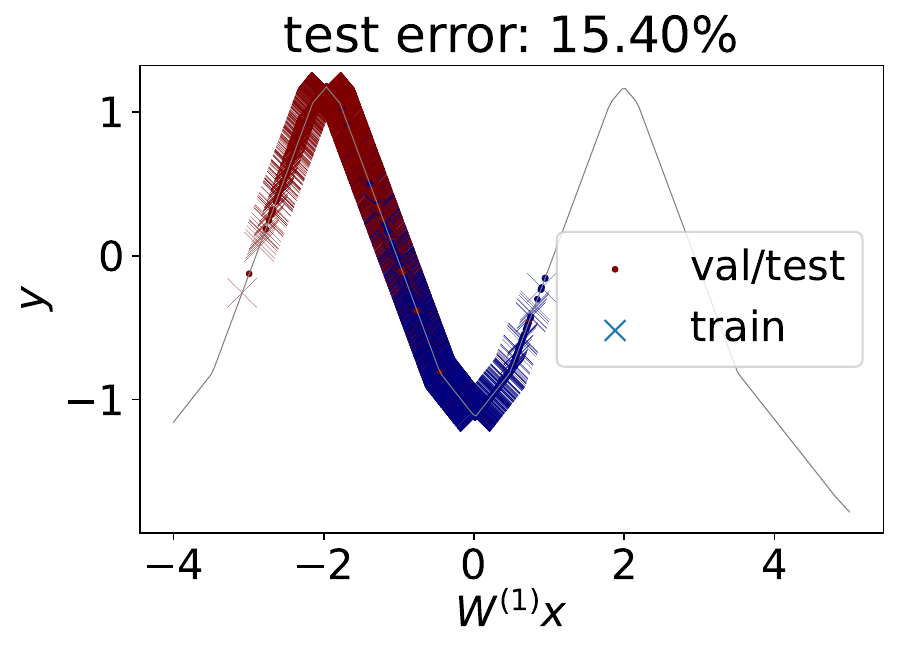}
    \end{minipage}
    %%%%%seed0
    \begin{minipage}[t]{0.4\textwidth}
    \includegraphics[width=\linewidth]{results2/IMDB/Bert/L=3/seed0/W1plot.pdf}
    \end{minipage}
    \begin{minipage}[t]{0.4\textwidth}
    \includegraphics[width=\linewidth]{results2/IMDB/Bert/L=4/seed0/W1plot.pdf}
    \end{minipage}
   %%%%%seed1
   \begin{minipage}[t]{0.4\textwidth}
    \includegraphics[width=\linewidth]{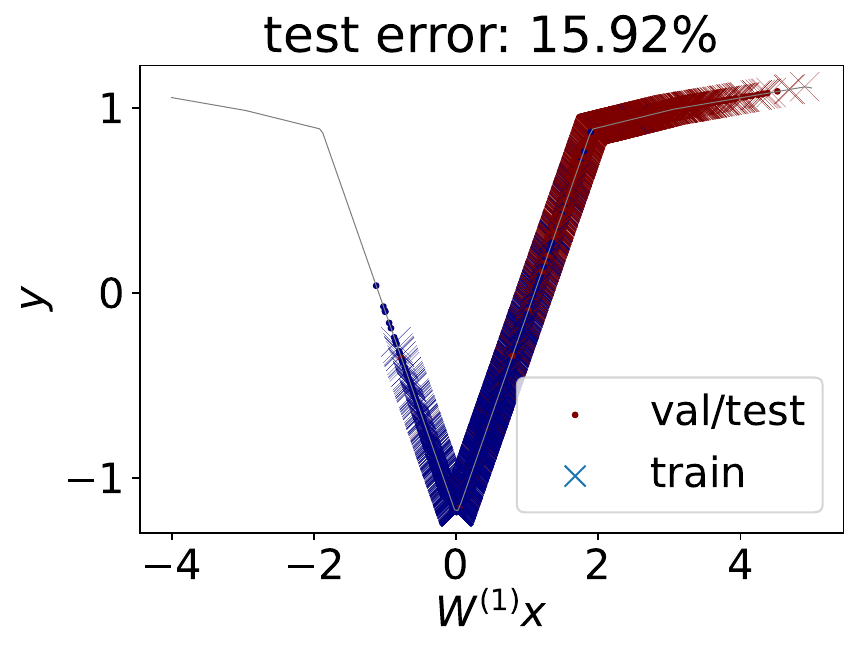}
    \end{minipage}
    \begin{minipage}[t]{0.4\textwidth}
    \includegraphics[width=\linewidth]{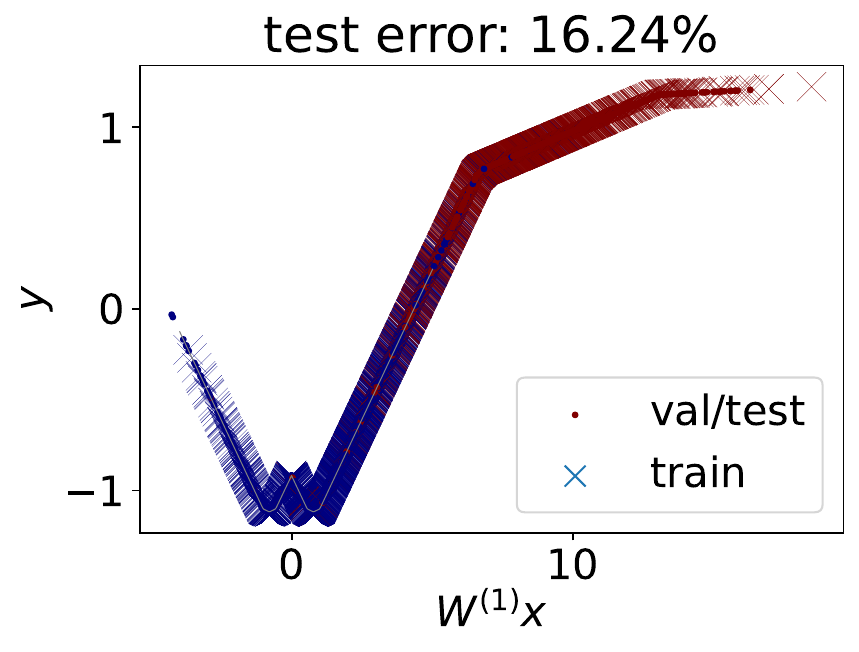}
    \end{minipage}
     %%%%%%%%seed3
   \begin{minipage}[t]{0.4\textwidth}
    \includegraphics[width=\linewidth]{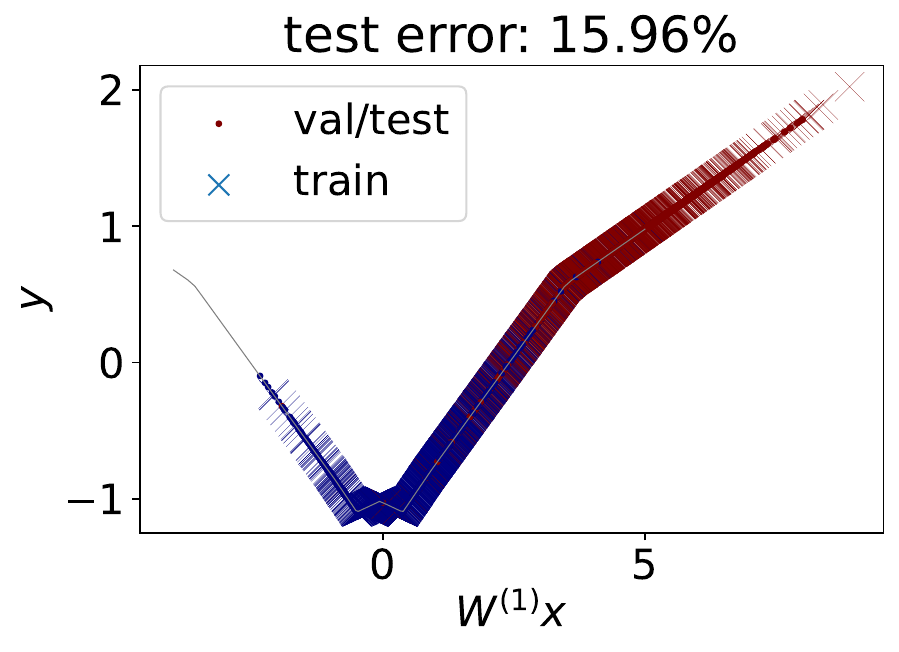}
    \end{minipage}
    \begin{minipage}[t]{0.4\textwidth}
    \includegraphics[width=\linewidth]{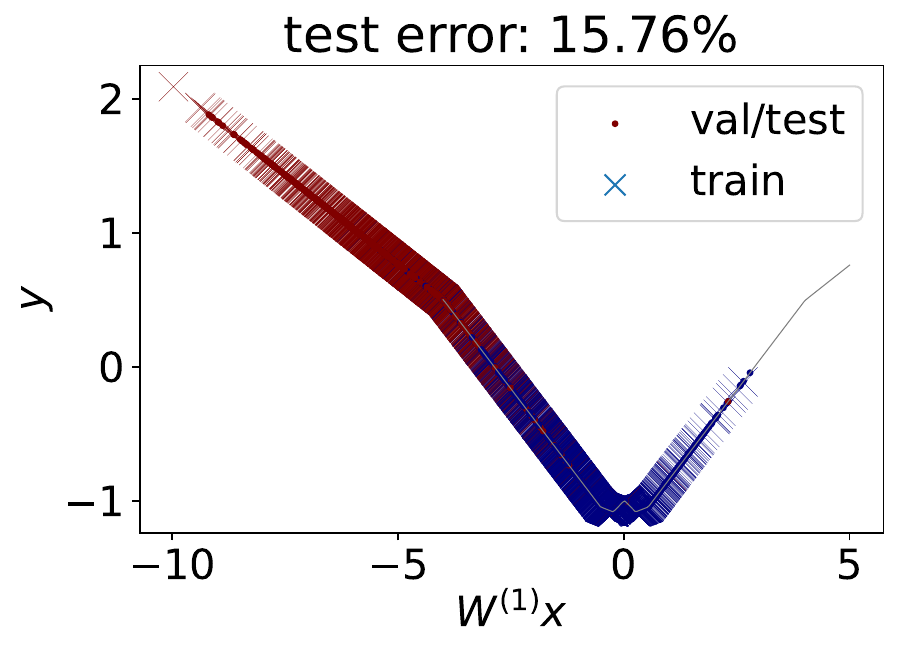}
    \end{minipage}
    \caption{$3$-layer (left column) and $4$-layer (right column) \reflectnn{}s trained on Bert embeddings of IMDB reviews. Each row is a different training initialization.}
    \label{fig:3lyrBertIMDB}
\end{figure}

\begin{figure}%OpenAI GlUE_COLA 32401
    \centering
    %%%%%%%%seed3
   \begin{minipage}[t]{0.4\textwidth}
    \includegraphics[width=\linewidth]{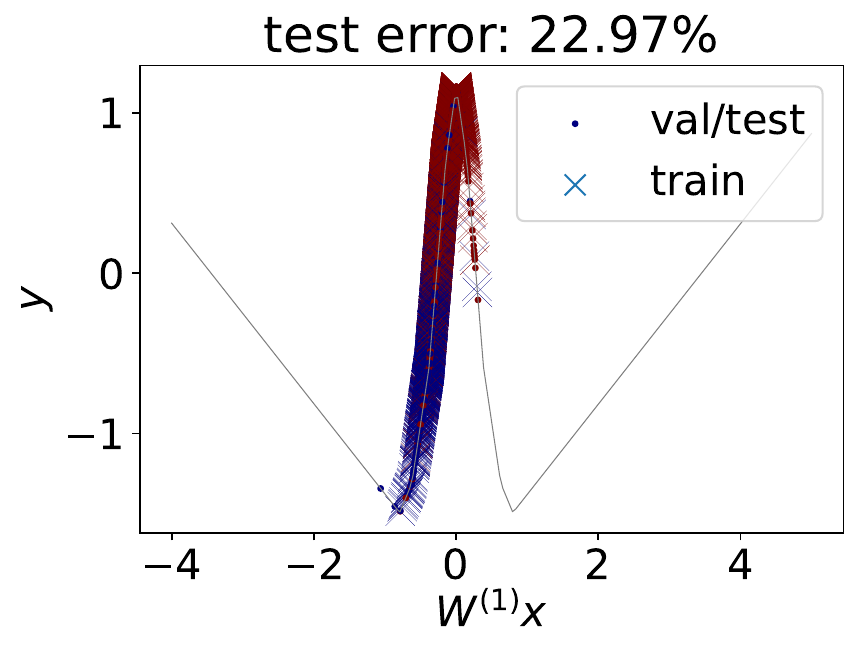}
    \end{minipage}
    \begin{minipage}[t]{0.4\textwidth}
    \includegraphics[width=\linewidth]{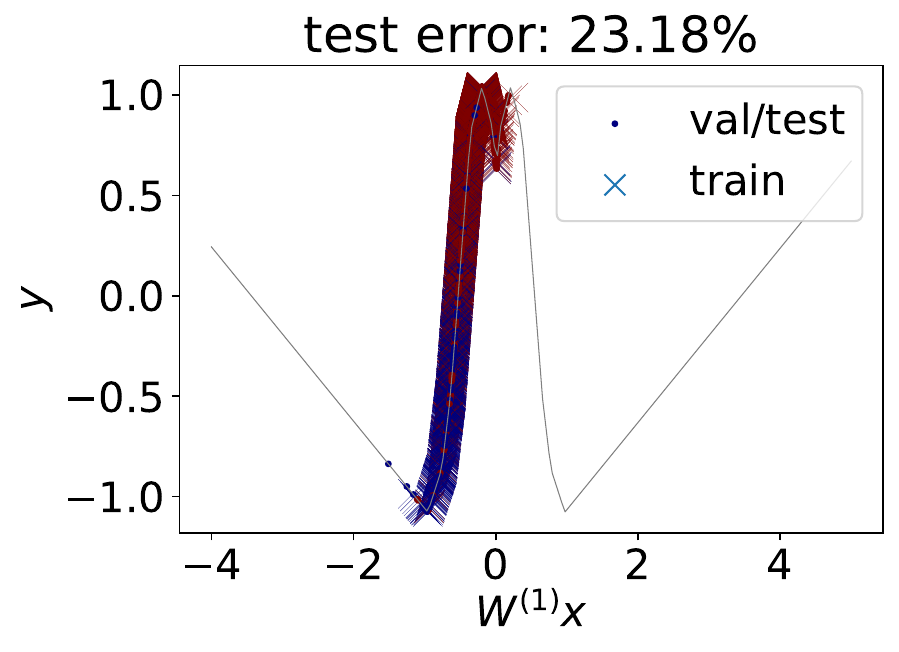}
    \end{minipage}
     %%%%%%%%seed2
   \begin{minipage}[t]{0.4\textwidth}
    \includegraphics[width=\linewidth]{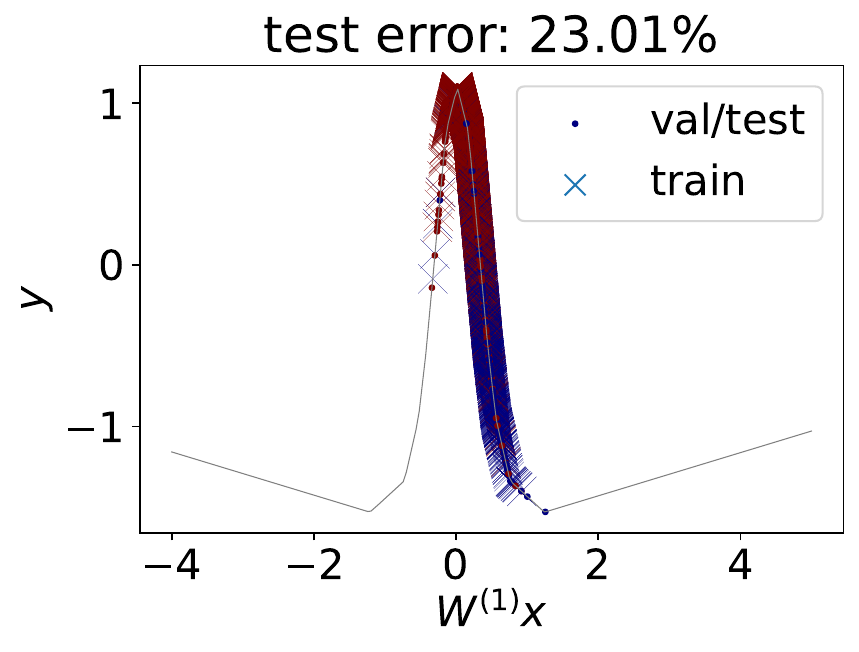}
    \end{minipage}
    \begin{minipage}[t]{0.4\textwidth}
    \includegraphics[width=\linewidth]{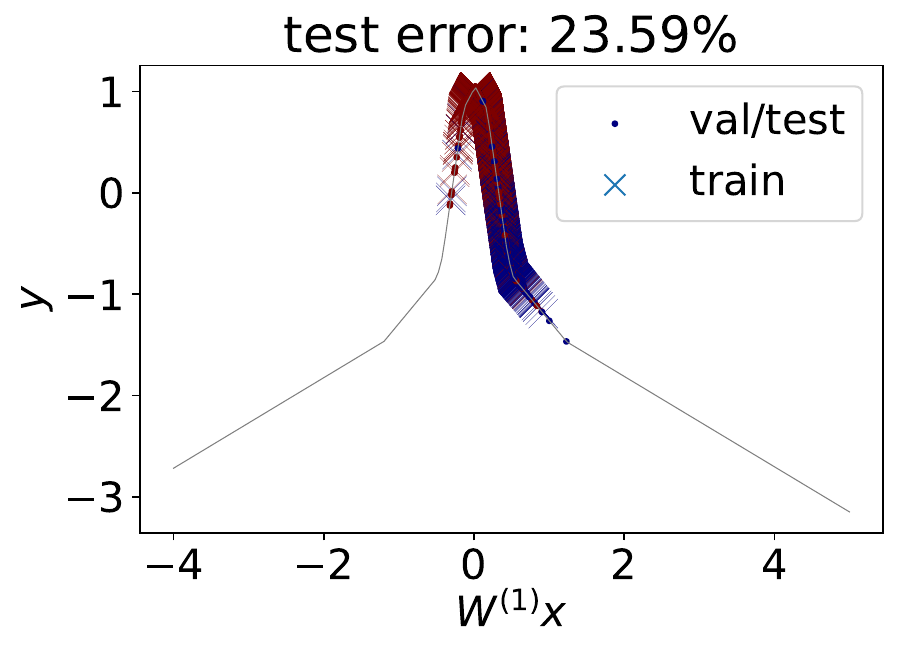}
    \end{minipage}
    %%%%%%%%seed4
   \begin{minipage}[t]{0.4\textwidth}
    \includegraphics[width=\linewidth]{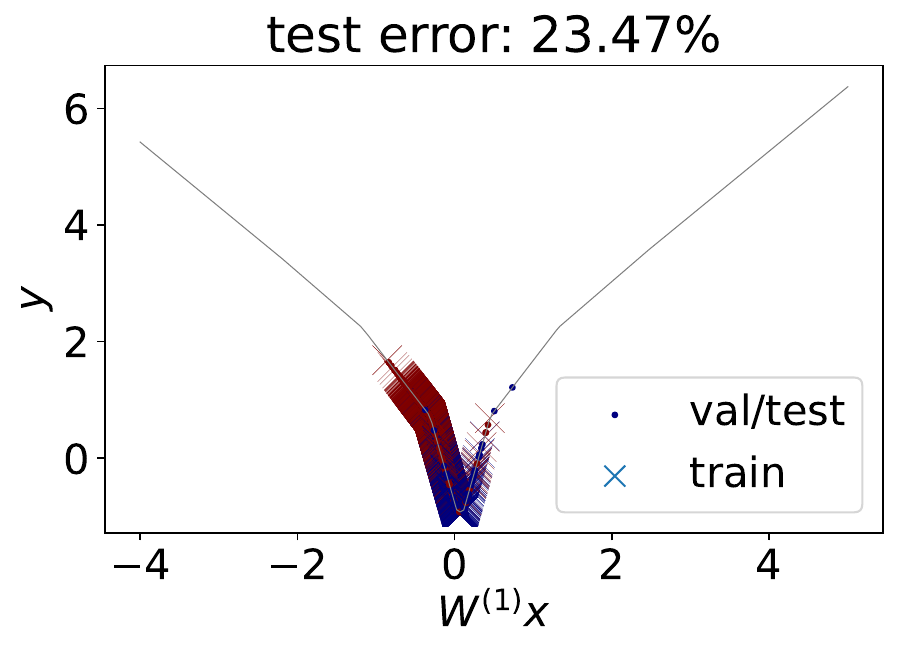}
    \end{minipage}
    \begin{minipage}[t]{0.4\textwidth}
    \includegraphics[width=\linewidth]{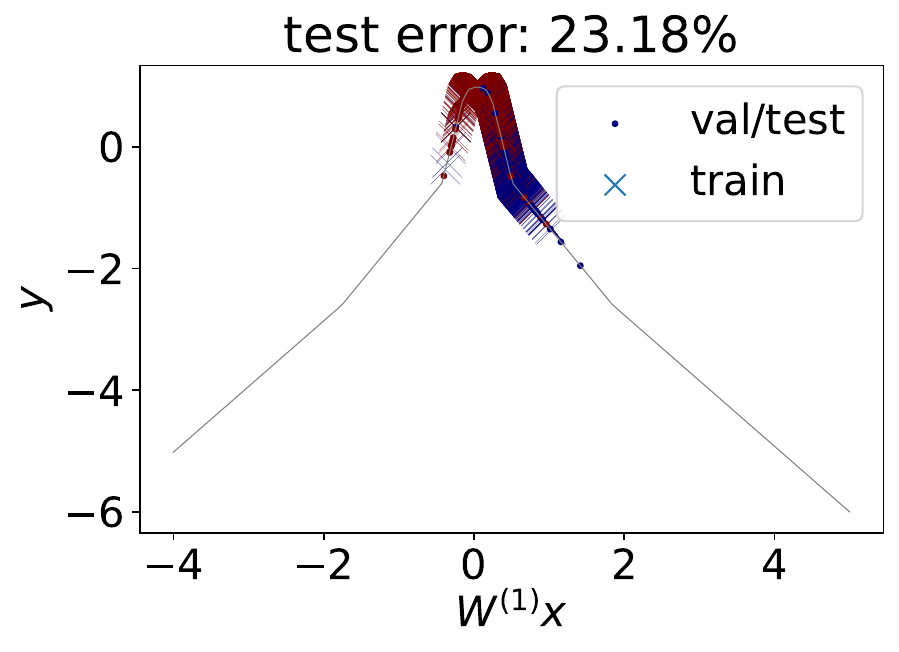}
    \end{minipage}
    %%%%%seed0
    \begin{minipage}[t]{0.4\textwidth}
    \includegraphics[width=\linewidth]{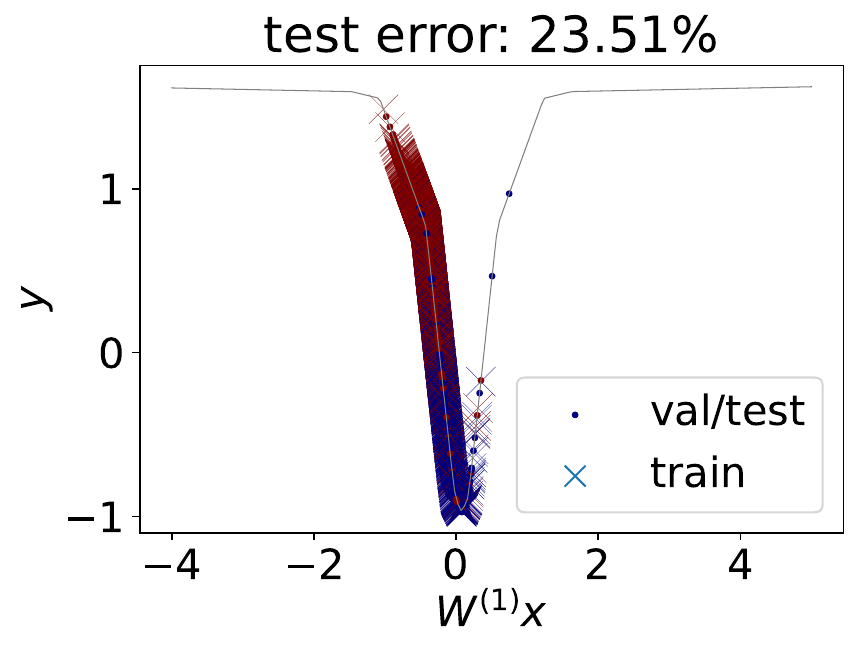}
    \end{minipage}
    \begin{minipage}[t]{0.4\textwidth}
    \includegraphics[width=\linewidth]{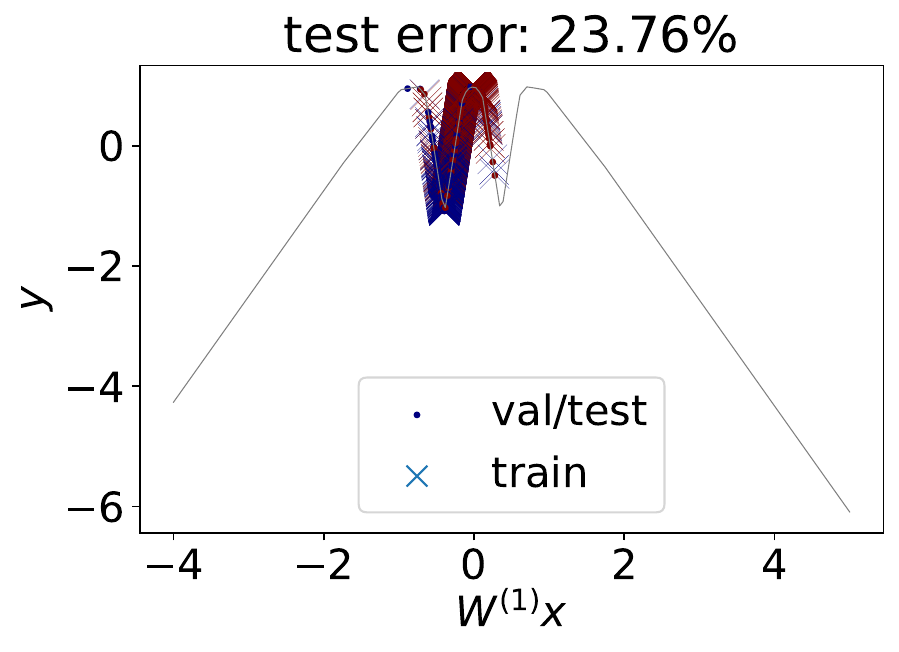}
    \end{minipage}
   %%%%%seed1
   \begin{minipage}[t]{0.4\textwidth}
    \includegraphics[width=\linewidth]{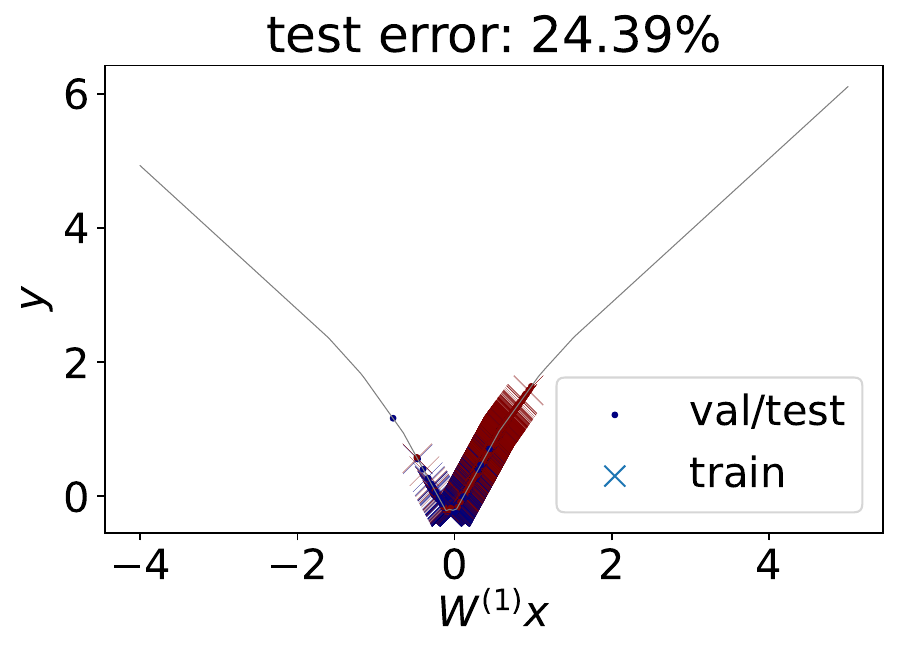}
    \end{minipage}
    \begin{minipage}[t]{0.4\textwidth}
    \includegraphics[width=\linewidth]{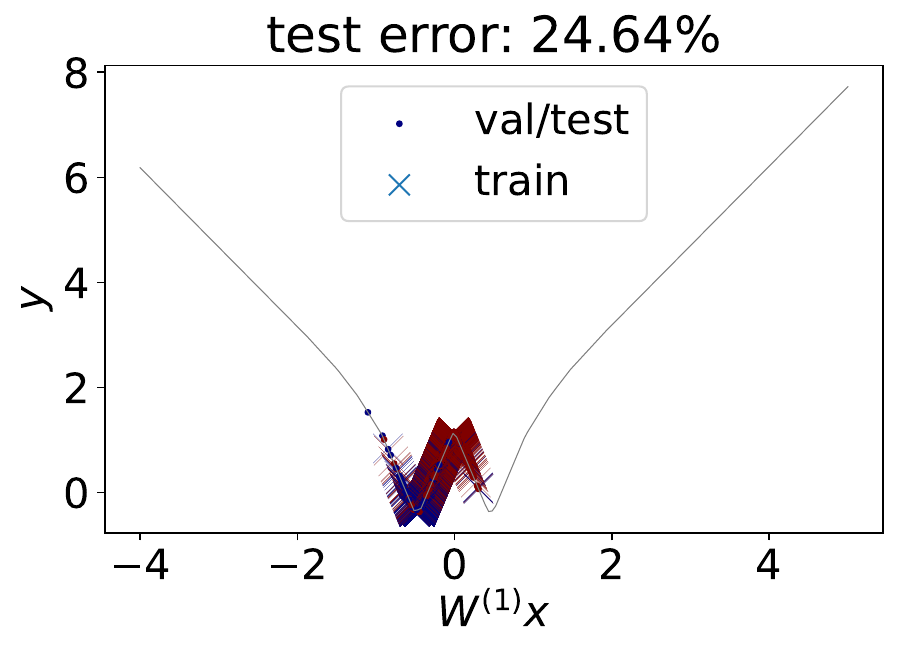}
    \end{minipage}

    \caption{$3$-layer (left column) and $4$-layer (right column) \reflectnn{}s trained on OpenAI embeddings of GLUE-CoLA text. Each row is a different training initialization.}
    \label{fig:3lyrOpenAIgluecola}
\end{figure}

\begin{figure}%Bert GlUE_COLA 14032
    \centering
    %%%%%seed1
   \begin{minipage}[t]{0.4\textwidth}
    \includegraphics[width=\linewidth]{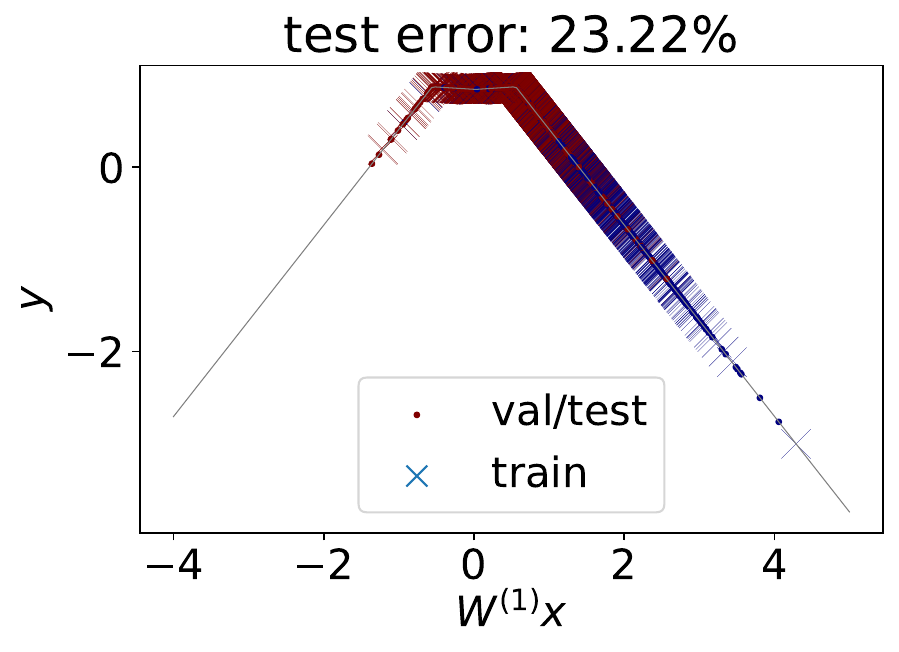}
    \end{minipage}
    \begin{minipage}[t]{0.4\textwidth}
    \includegraphics[width=\linewidth]{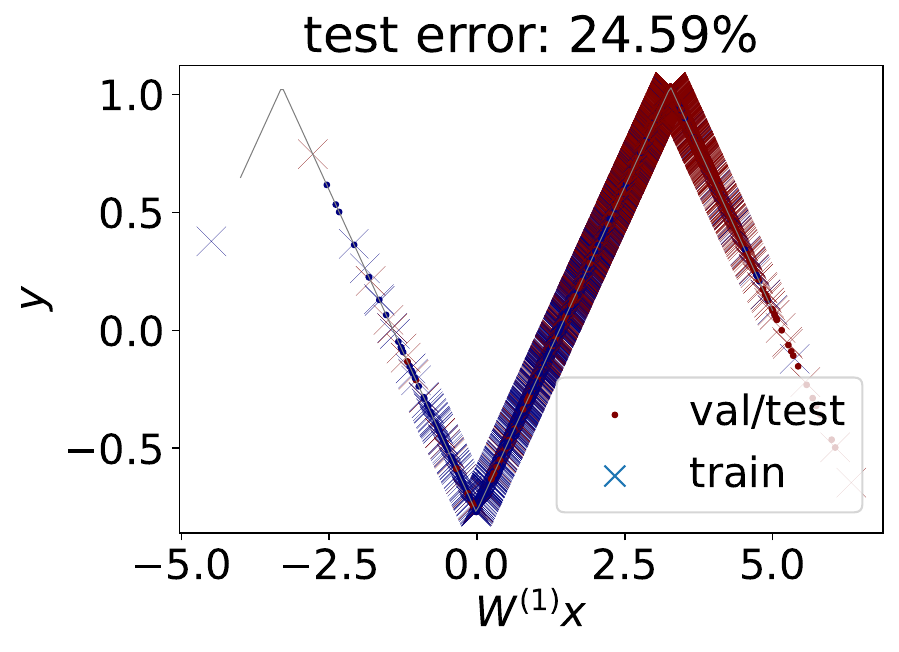}
    \end{minipage}
         %%%%%%%%seed4
   \begin{minipage}[t]{0.4\textwidth}
    \includegraphics[width=\linewidth]{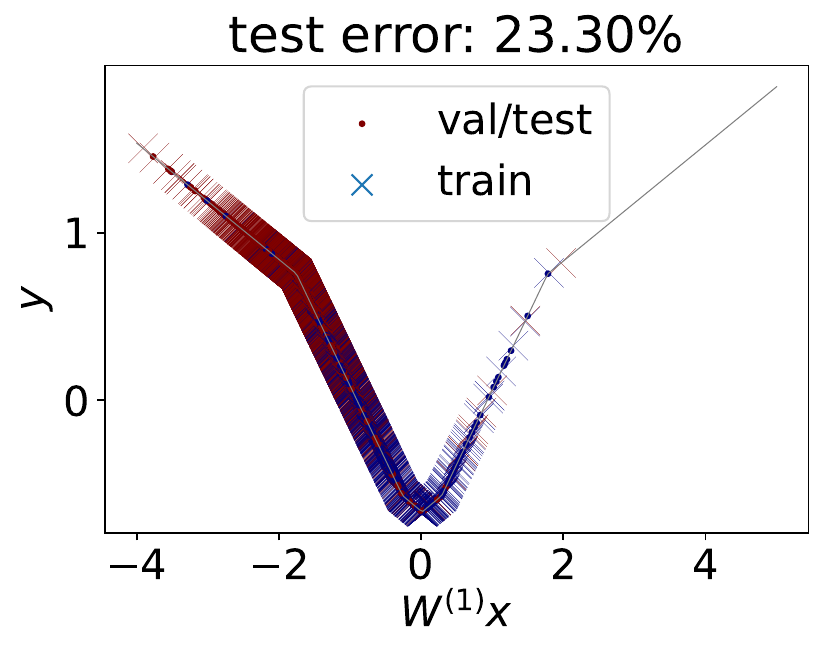}
    \end{minipage}
    \begin{minipage}[t]{0.4\textwidth}
    \includegraphics[width=\linewidth]{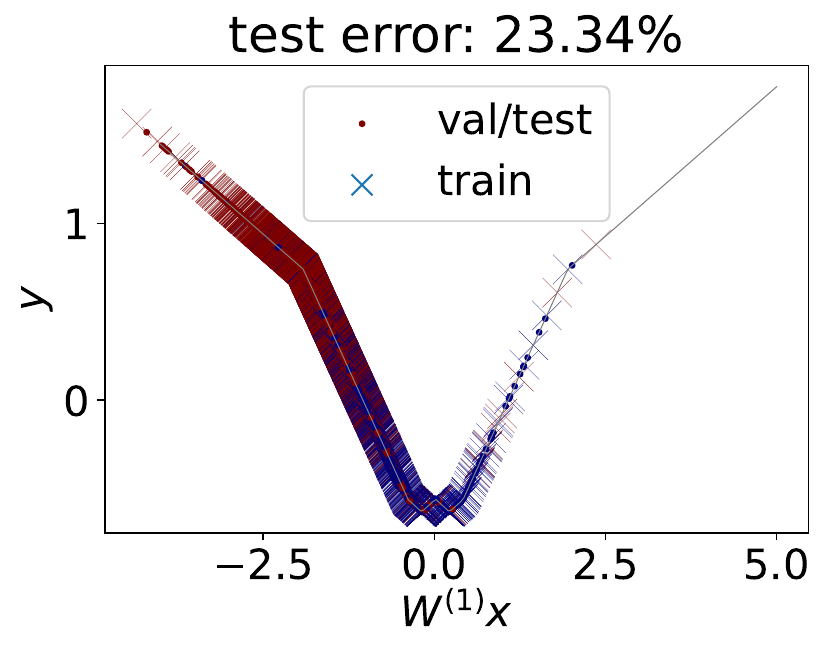}
    \end{minipage}
    %%%%%seed0
    \begin{minipage}[t]{0.4\textwidth}
    \includegraphics[width=\linewidth]{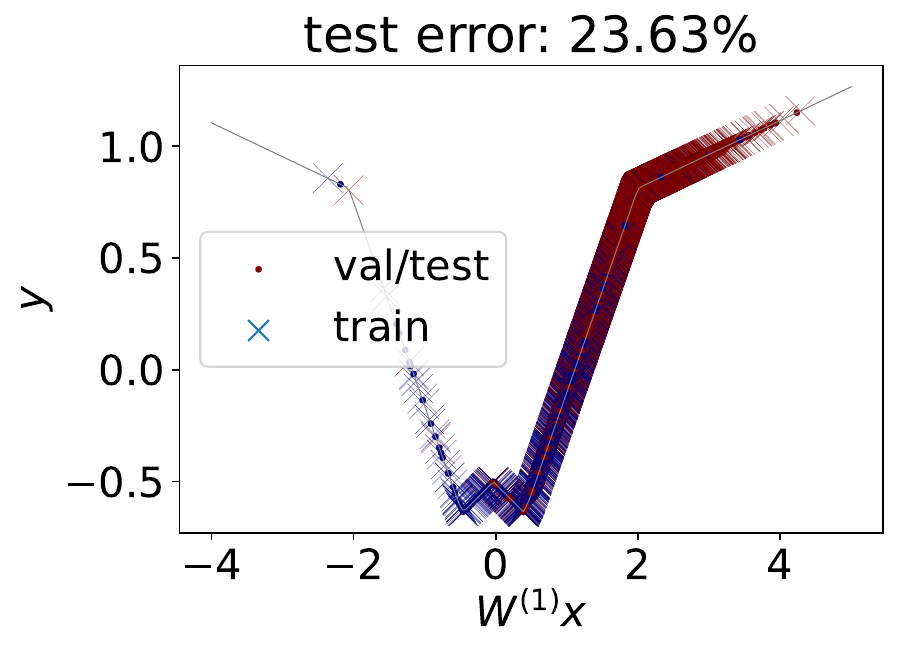}
    \end{minipage}
    \begin{minipage}[t]{0.4\textwidth}
    \includegraphics[width=\linewidth]{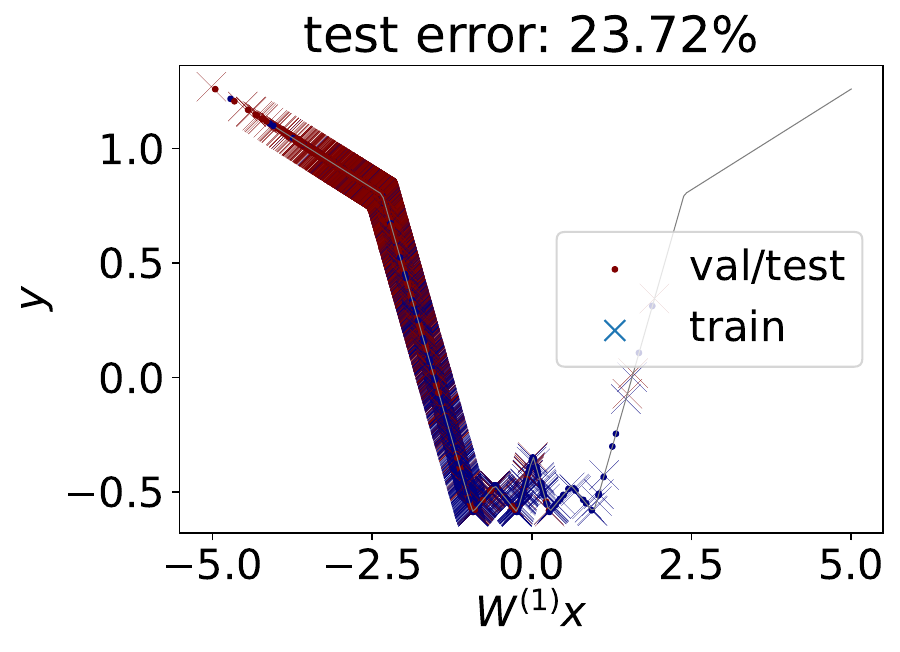}
    \end{minipage}
     %%%%%%%%seed3
   \begin{minipage}[t]{0.4\textwidth}
    \includegraphics[width=\linewidth]{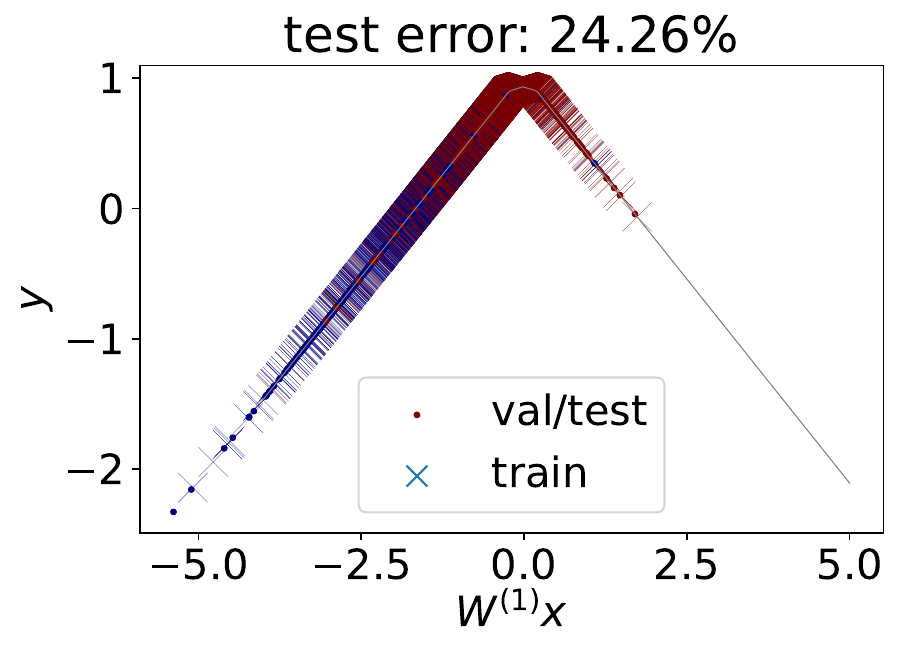}
    \end{minipage}
    \begin{minipage}[t]{0.4\textwidth}
    \includegraphics[width=\linewidth]{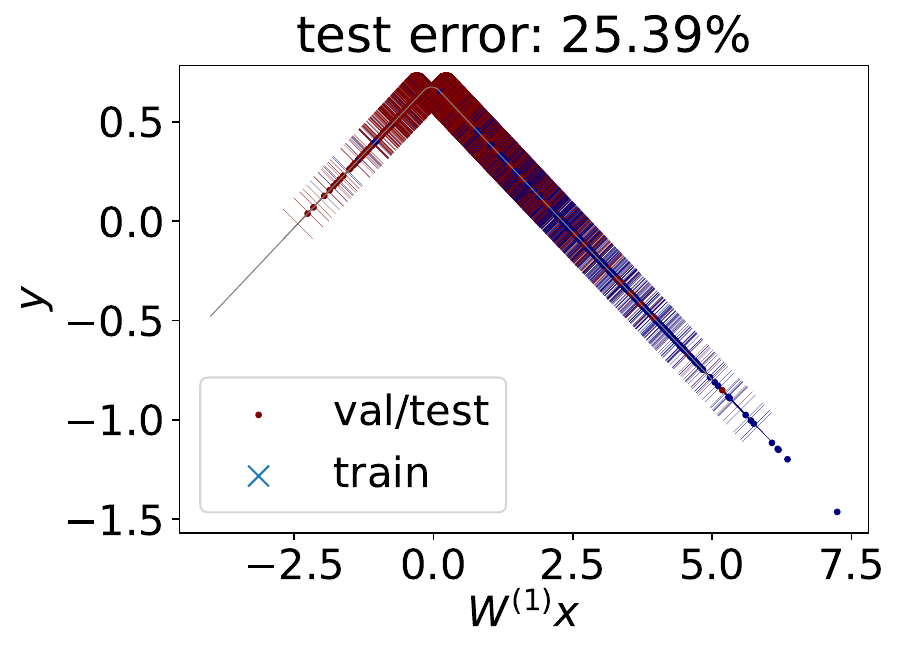}
    \end{minipage}
    %%%%%%%%seed2
   \begin{minipage}[t]{0.4\textwidth}
    \includegraphics[width=\linewidth]{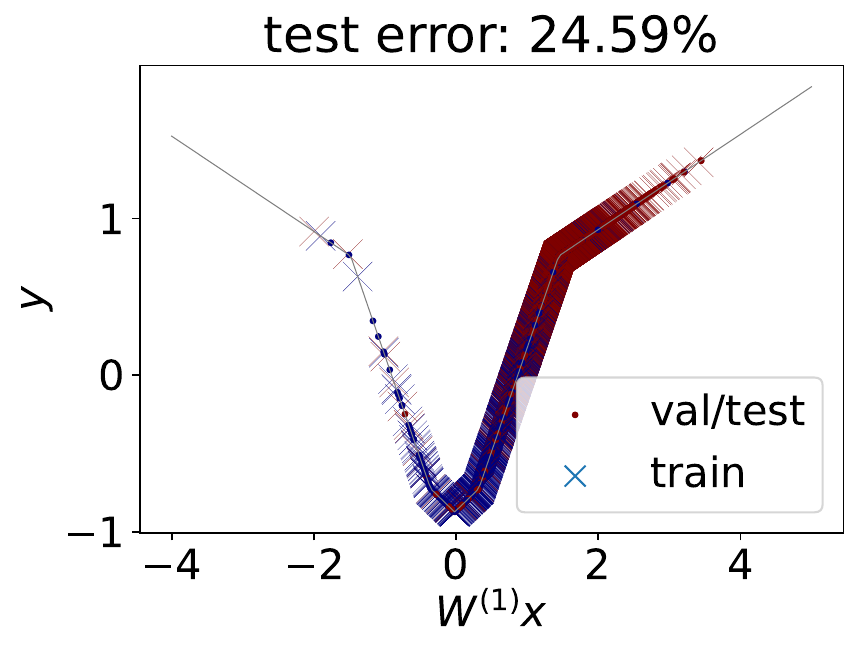}
    \end{minipage}
    \begin{minipage}[t]{0.4\textwidth}
    \includegraphics[width=\linewidth]{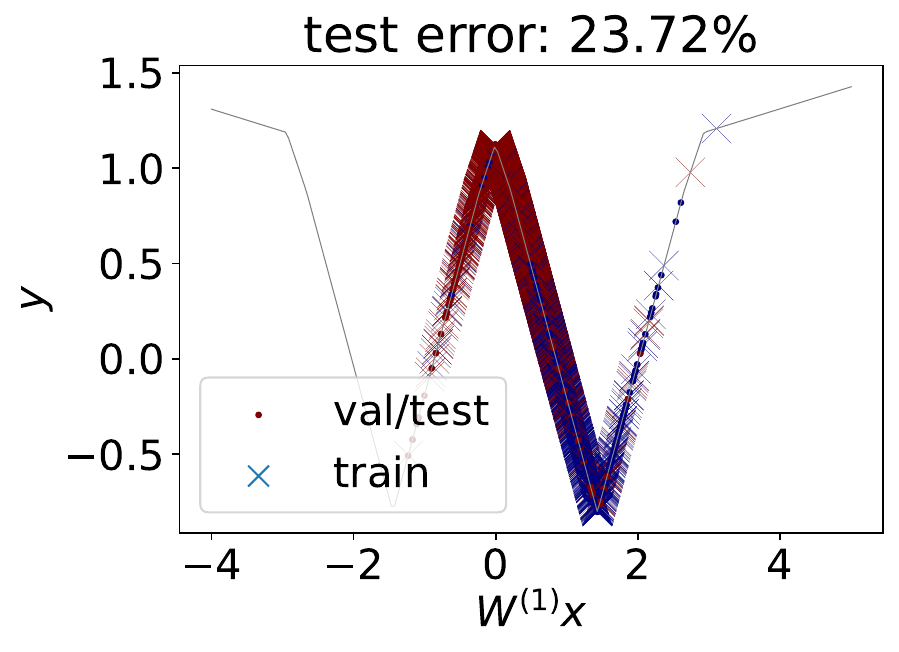}
    \end{minipage}

    \caption{$3$-layer (left column) and $4$-layer (right column) \reflectnn{}s trained on Bert embeddings of GLUE-CoLA text. Each row is a different training initialization.}
    \label{fig:3lyrBertgluecola}
\end{figure}

\begin{figure}%OpenAI GlUE_QQP 13204
    \centering
    %%%%%seed1
   \begin{minipage}[t]{0.4\textwidth}
    \includegraphics[width=\linewidth]{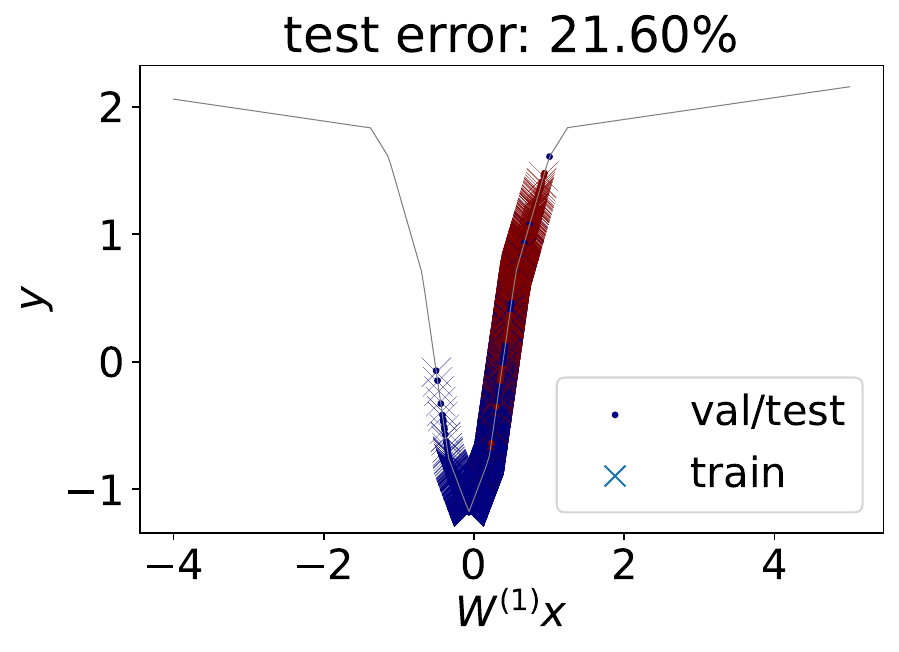}
    \end{minipage}
    \begin{minipage}[t]{0.4\textwidth}
    \includegraphics[width=\linewidth]{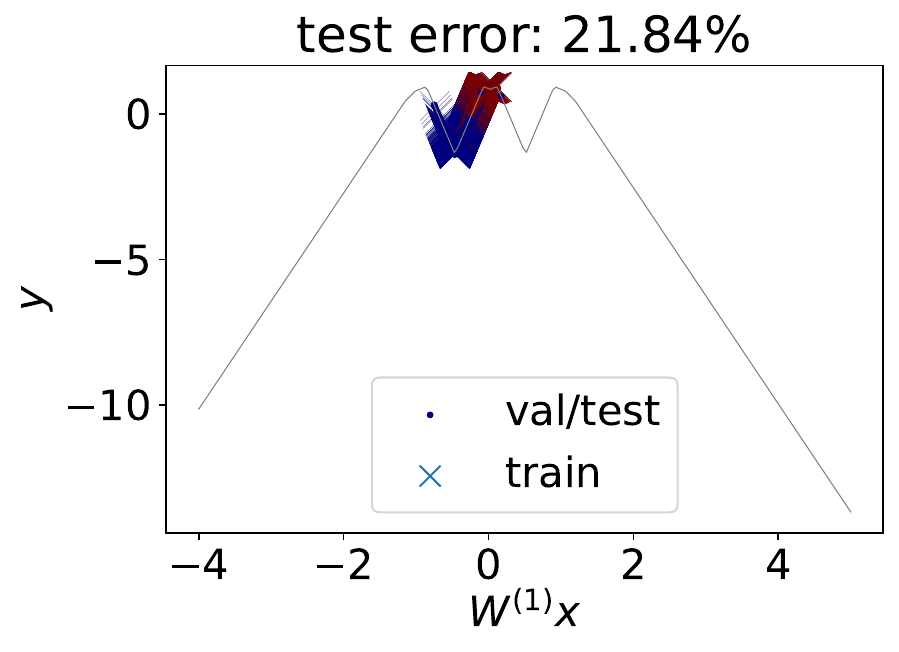}
    \end{minipage}
    %%%%%%%%seed3
   \begin{minipage}[t]{0.4\textwidth}
    \includegraphics[width=\linewidth]{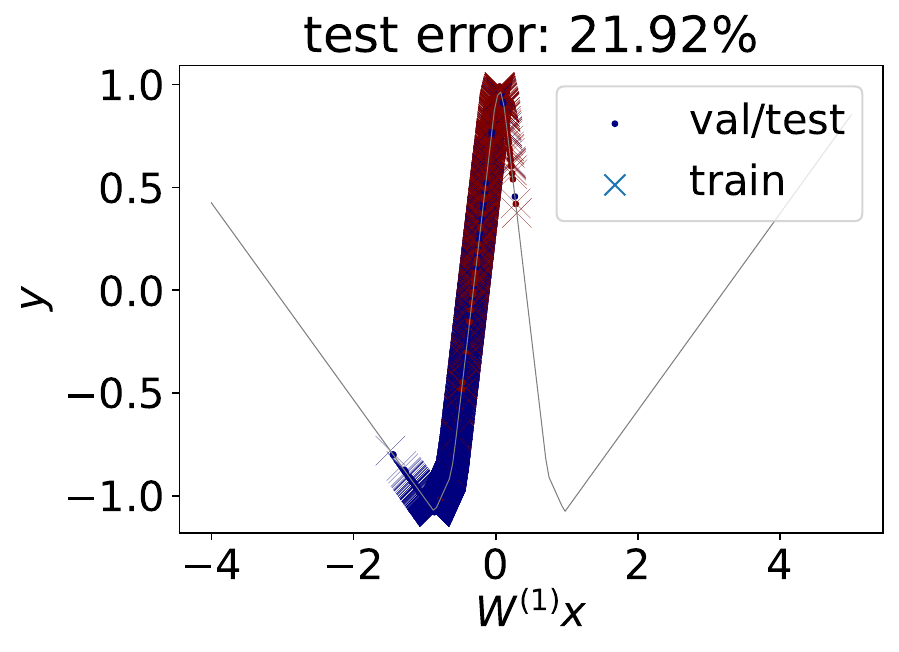}
    \end{minipage}
    \begin{minipage}[t]{0.4\textwidth}
    \includegraphics[width=\linewidth]{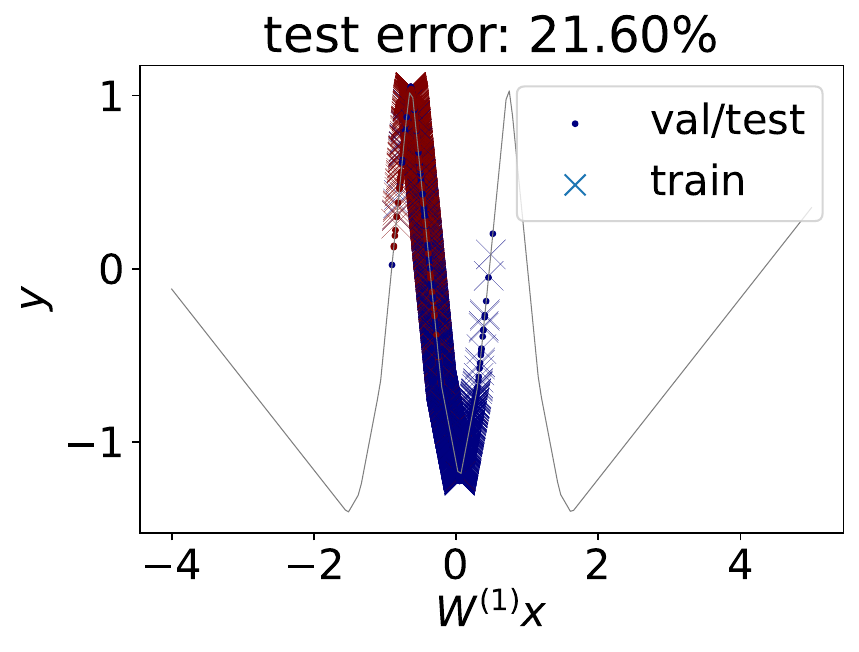}
    \end{minipage}
    %%%%%%%%seed2
   \begin{minipage}[t]{0.4\textwidth}
    \includegraphics[width=\linewidth]{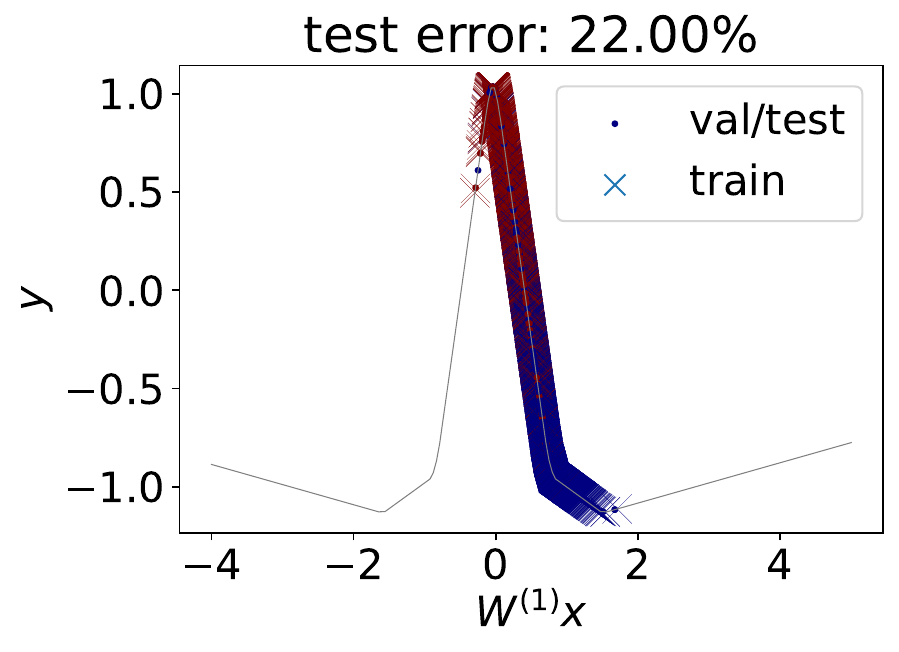}
    \end{minipage}
    \begin{minipage}[t]{0.4\textwidth}
    \includegraphics[width=\linewidth]{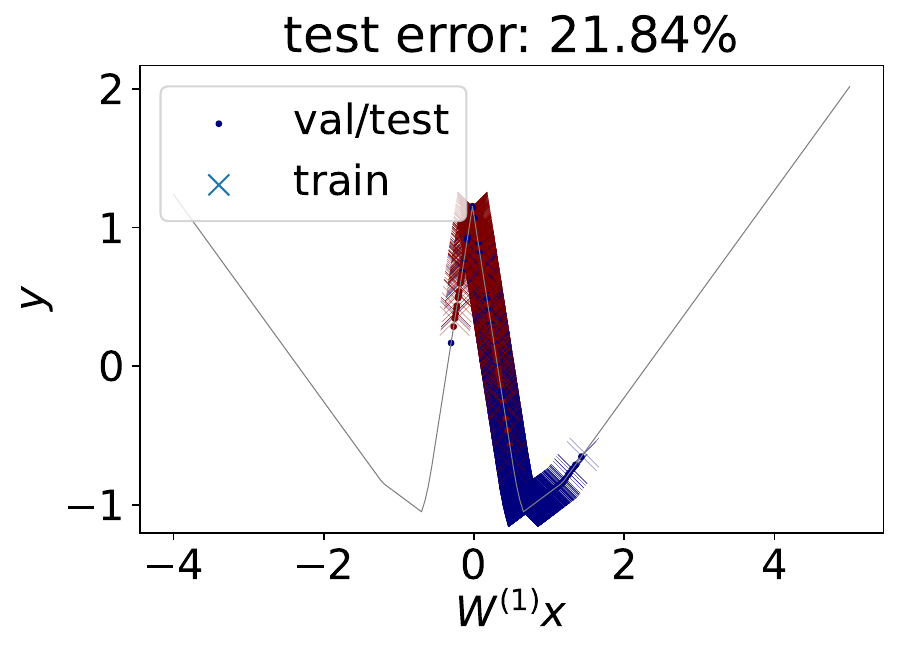}
    \end{minipage}
    %%%%%seed0
    \begin{minipage}[t]{0.4\textwidth}
    \includegraphics[width=\linewidth]{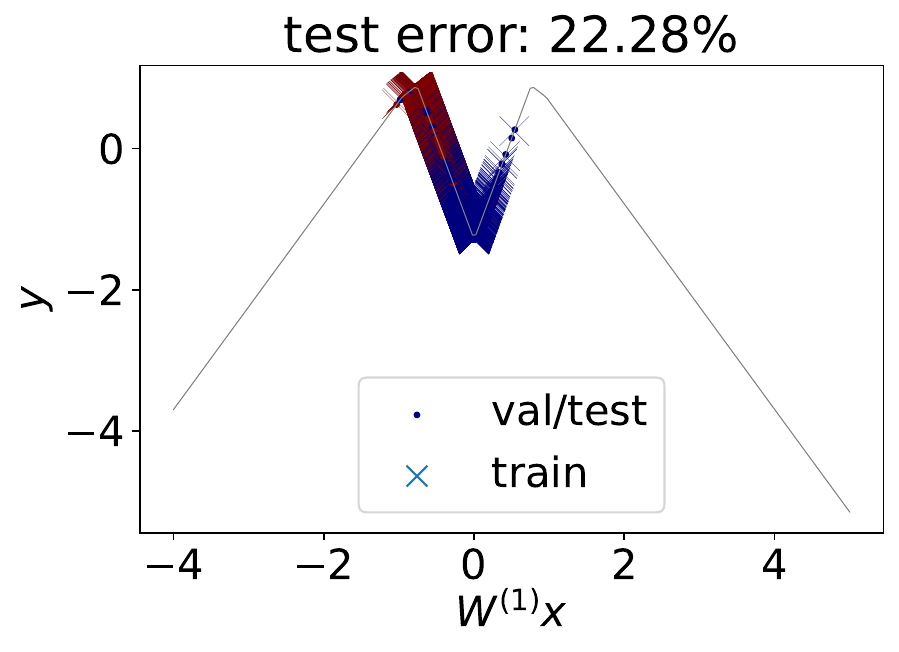}
    \end{minipage}
    \begin{minipage}[t]{0.4\textwidth}
    \includegraphics[width=\linewidth]{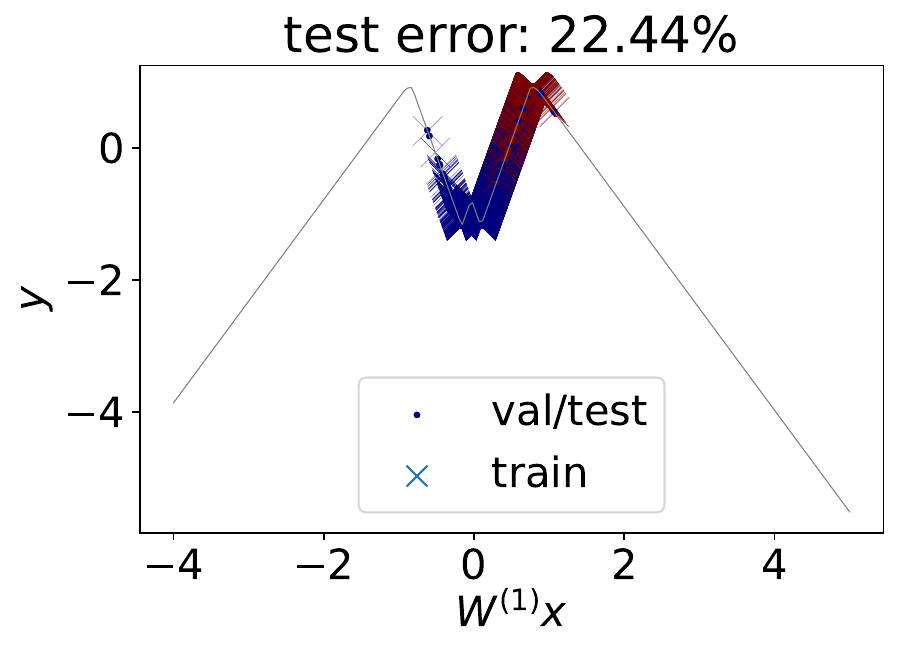}
    \end{minipage}
     %%%%%%%%seed4
   \begin{minipage}[t]{0.4\textwidth}
    \includegraphics[width=\linewidth]{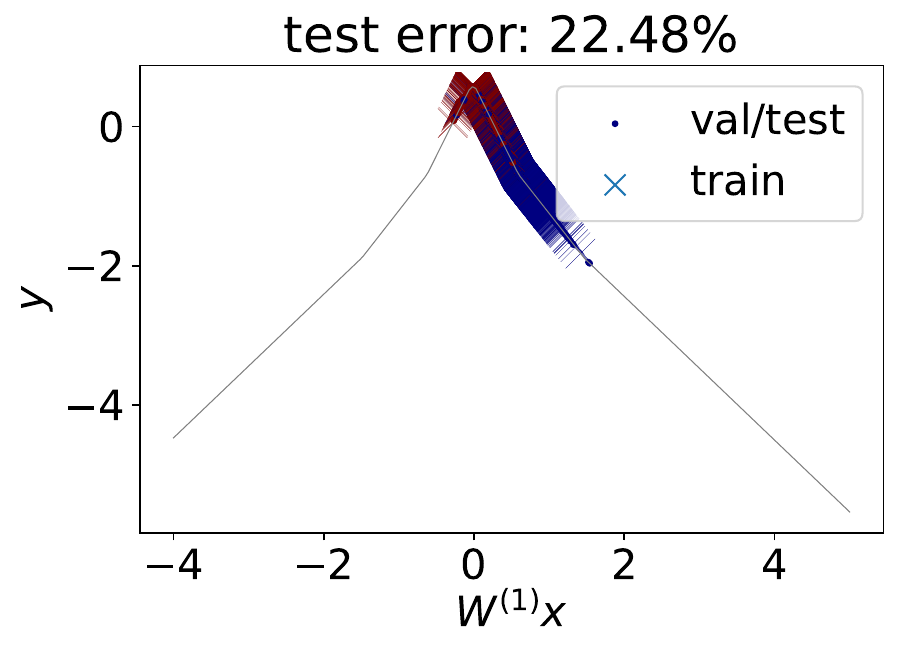}
    \end{minipage}
    \begin{minipage}[t]{0.4\textwidth}
    \includegraphics[width=\linewidth]{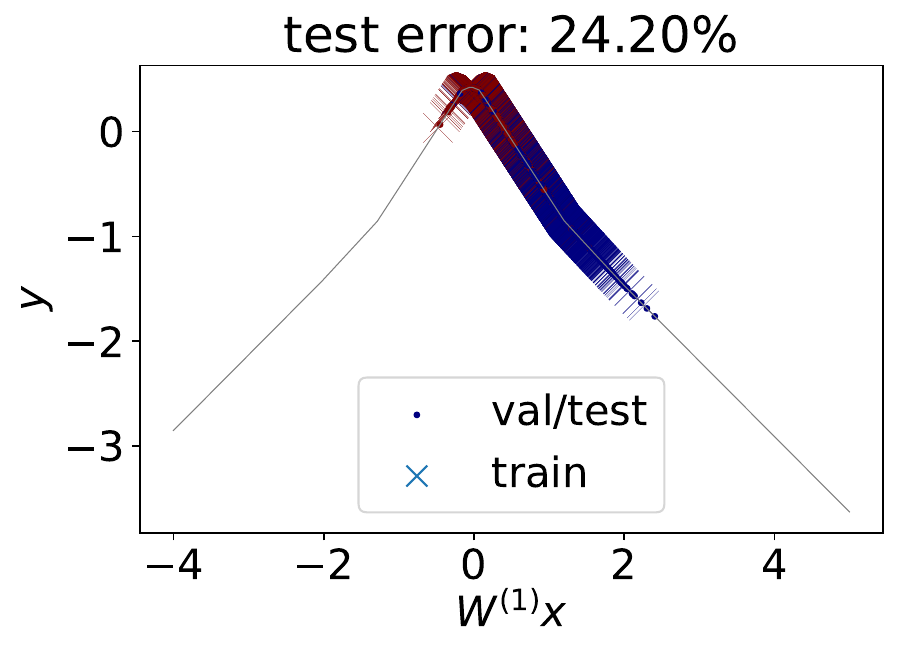}
    \end{minipage}
    \caption{$3$-layer (left column) and $4$-layer (right column) \reflectnn{}s trained on OpenAI embeddings of GLUE-QQP text. Each row is a different training initialization.}
    \label{fig:3lyrOpenAIglueqqp}
\end{figure}

\begin{figure}%Bert GlUE_QQP 31024
    \centering
     %%%%%%%%seed3
   \begin{minipage}[t]{0.4\textwidth}
    \includegraphics[width=\linewidth]{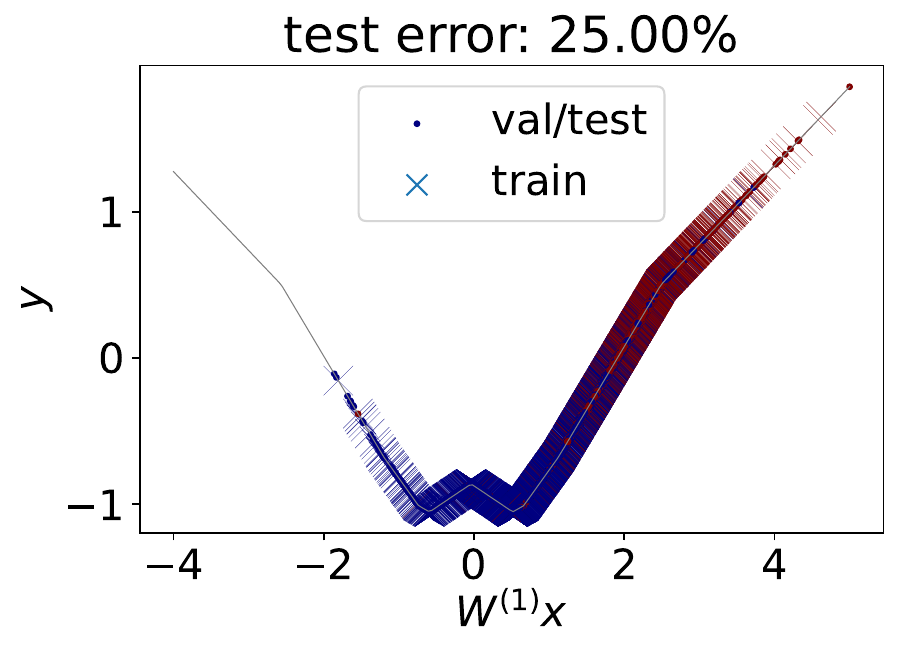}
    \end{minipage}
    \begin{minipage}[t]{0.4\textwidth}
    \includegraphics[width=\linewidth]{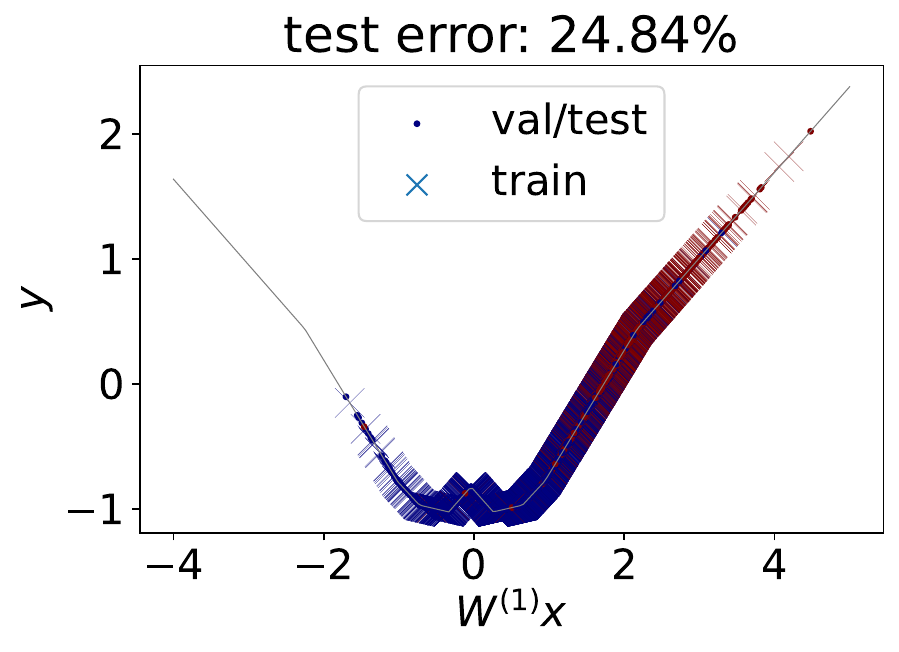}
    \end{minipage}
    %%%%%seed1
   \begin{minipage}[t]{0.4\textwidth}
    \includegraphics[width=\linewidth]{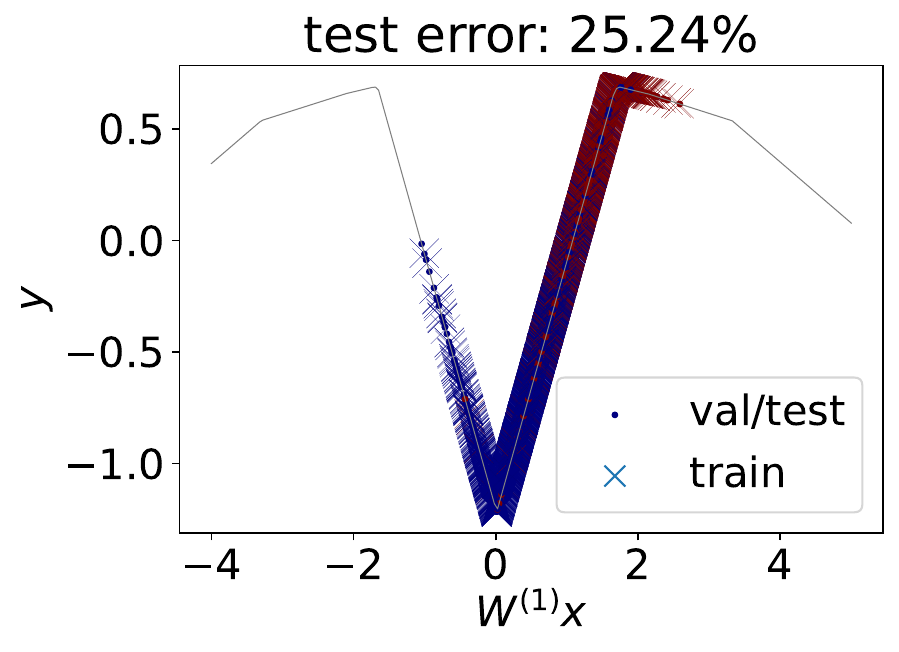}
    \end{minipage}
    \begin{minipage}[t]{0.4\textwidth}
    \includegraphics[width=\linewidth]{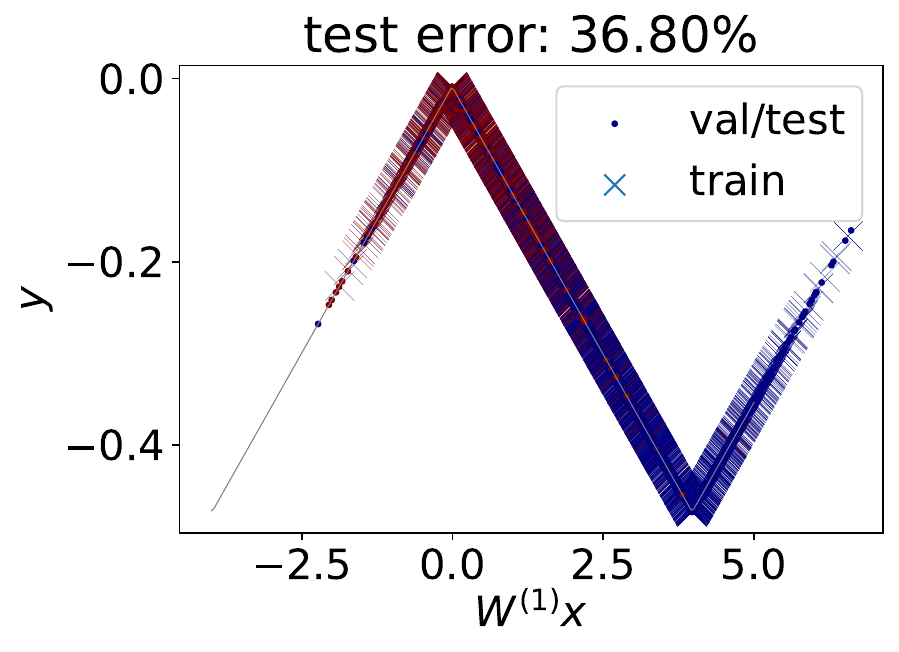}
    \end{minipage}
    %%%%%seed0
    \begin{minipage}[t]{0.4\textwidth}
    \includegraphics[width=\linewidth]{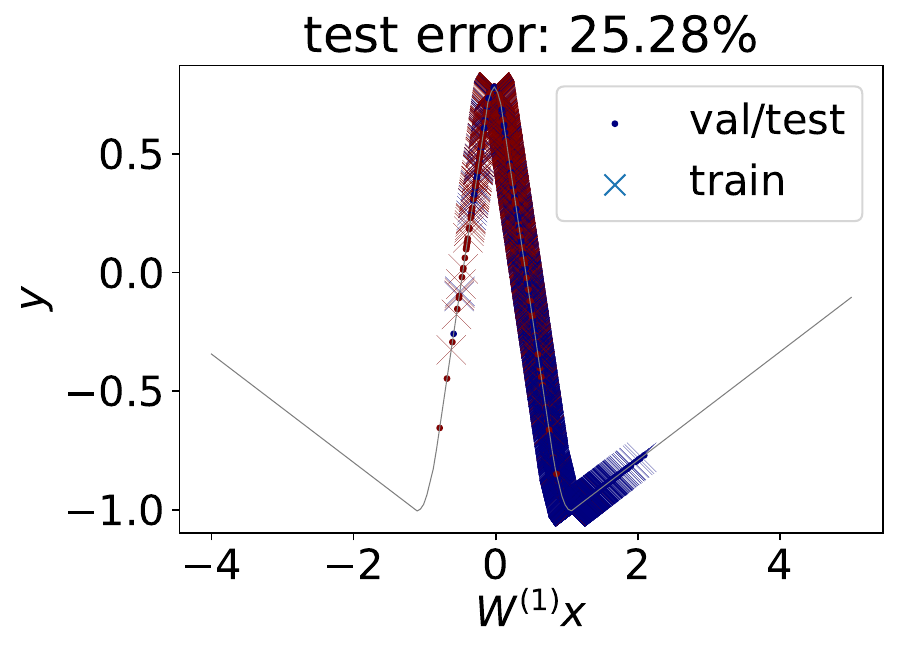}
    \end{minipage}
    \begin{minipage}[t]{0.4\textwidth}
    \includegraphics[width=\linewidth]{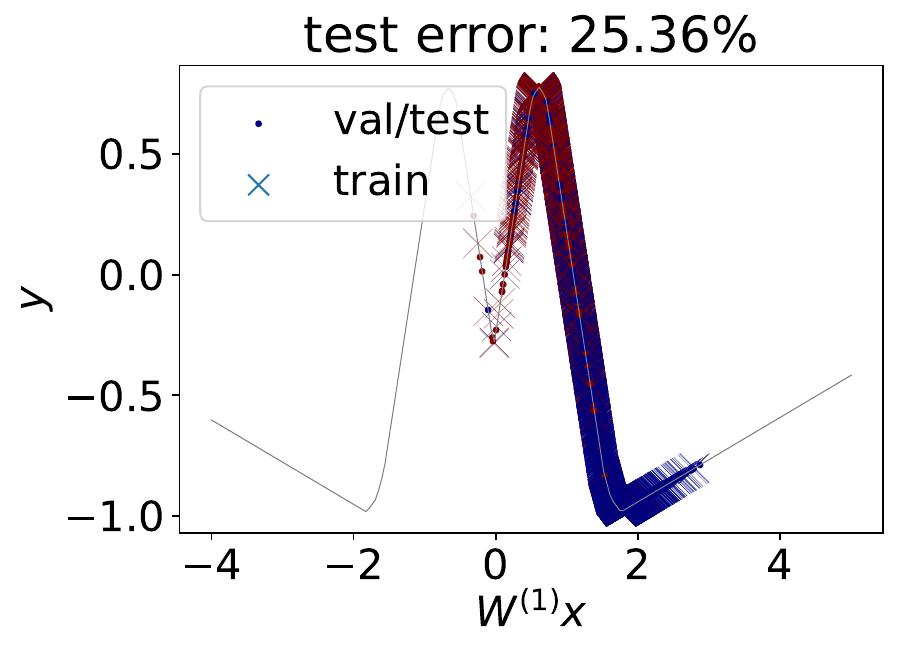}
    \end{minipage}
    %%%%%%%%seed2
   \begin{minipage}[t]{0.4\textwidth}
    \includegraphics[width=\linewidth]{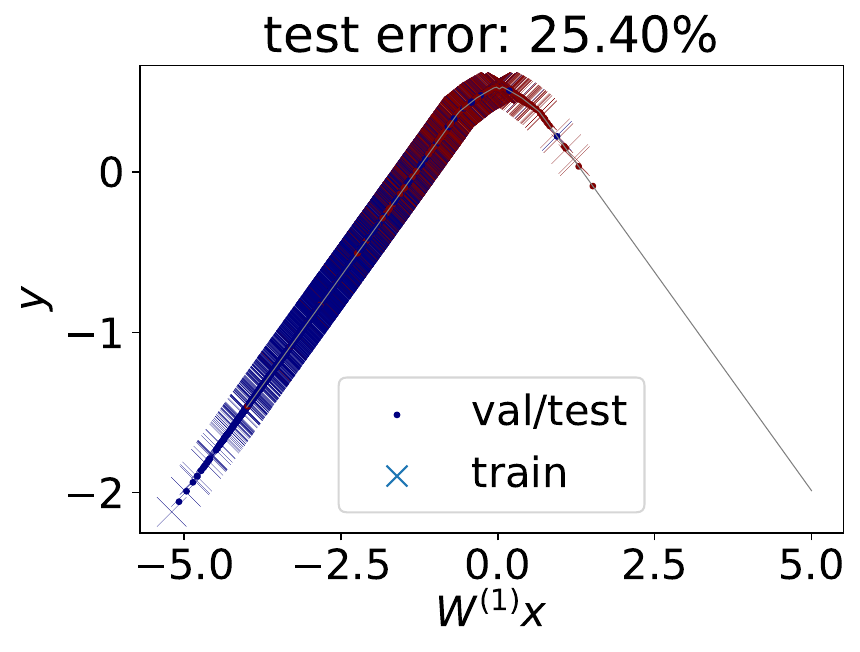}
    \end{minipage}
    \begin{minipage}[t]{0.4\textwidth}
    \includegraphics[width=\linewidth]{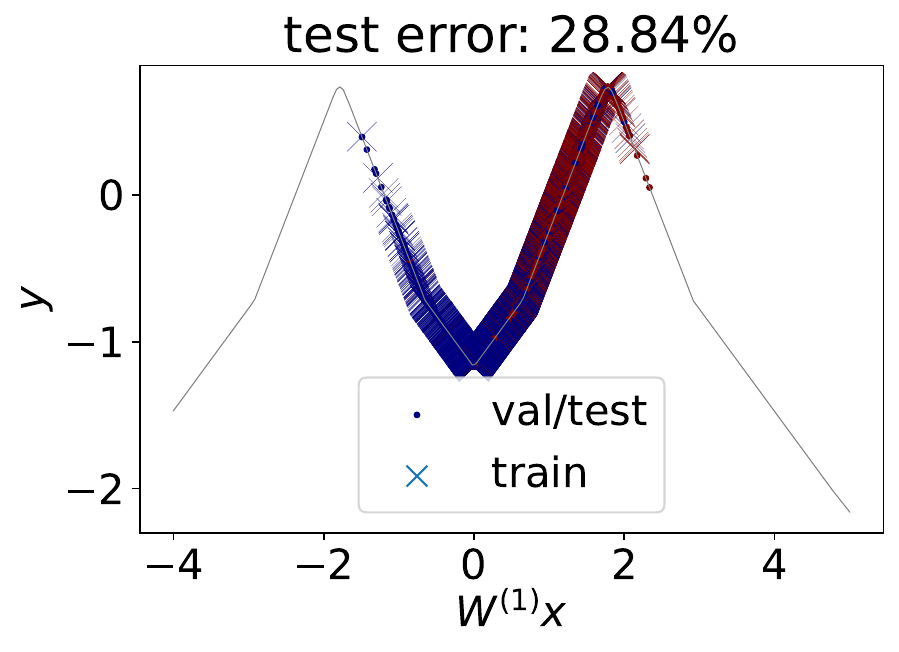}
    \end{minipage}
     %%%%%%%%seed4
   \begin{minipage}[t]{0.4\textwidth}
    \includegraphics[width=\linewidth]{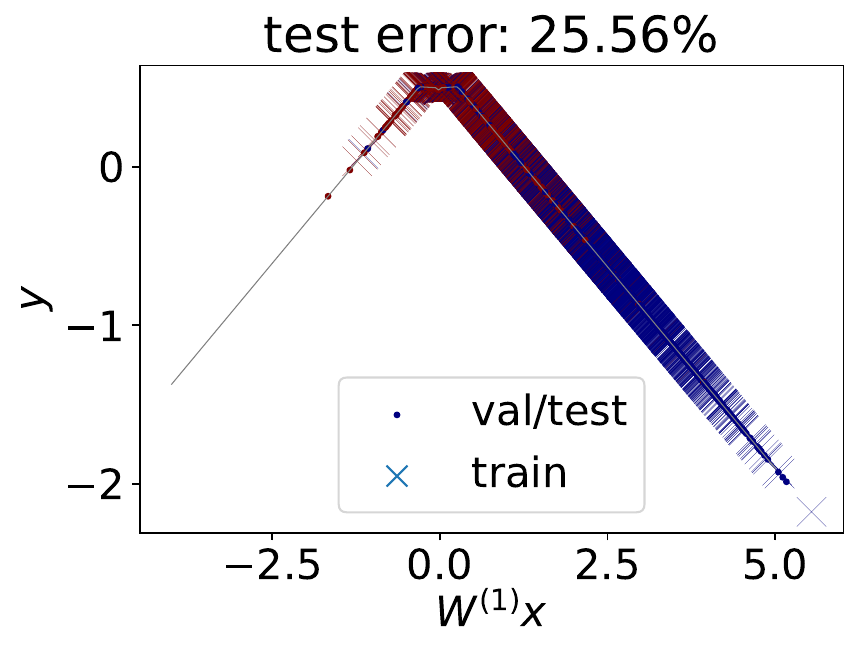}
    \end{minipage}
    \begin{minipage}[t]{0.4\textwidth}
    \includegraphics[width=\linewidth]{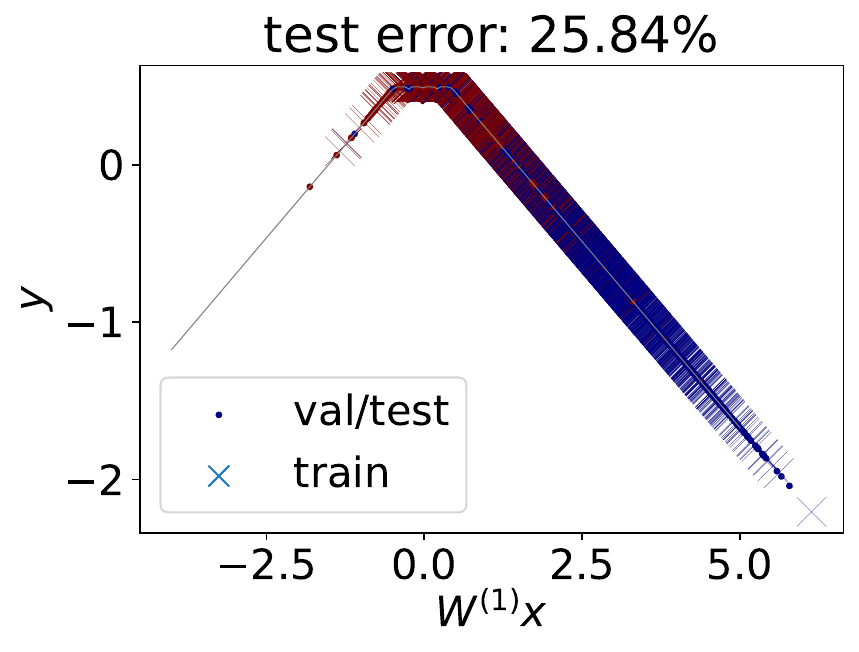}
    \end{minipage}
    \caption{$3$-layer (left column) and $4$-layer (right column) \reflectnn{}s trained on Bert embeddings of GLUE-QQP text. Each row is a different training initialization.}
    \label{fig:3lyrBertglueqqp}
\end{figure}

We perform the same experiments as in \Cref{section:LLM}, but with the GLUE data set, specifically GLUE-CoLA and GLUE-QQP \citep{glue, warstadt2018neural, socher2013recursive, dolan2005automatically, agirre2007semantic, williams2018broad, rajpurkar2016squad, dagan2006pascal, bar2006second, giampiccolo2007third, levesque2011winograd, bentivogli2009fifth}. GLUE (General Language Understanding Evaluation) is a benchmark for testing the performance of models learning language processing. GLUE-CoLA (Corpus of Linguistic Acceptability) contains text that is to be binary classified whether it is grammatically correct or not. GLUE-QQP (Quora Question Pairs) contains pairs of questions that are to be binary classified as having the same semantic meaning or not. \Cref{fig:3lyrOpenAIIMDB}, \Cref{fig:3lyrBertIMDB}, \Cref{fig:3lyrOpenAIgluecola}, \Cref{fig:3lyrBertgluecola}, \Cref{fig:3lyrOpenAIglueqqp}, \Cref{fig:3lyrBertglueqqp} plot the results. Vector embeddings from both GPT4 and Bert \citep{BERT} LLMs are used. 
As in \Cref{fig:imdb_openaiL=3absval}, points with $y_n{=}1$ are plotted in red, and points with $y_n{=}{-}1$ are plotted in blue. The overall network is plotted in gray. The training data is plotted with 'x' and the validation and test data are plotted with '.' markers.

A standard architecture \eqref{eq:standardnet} was trained using \adam{} with a learning rate of $5(10^{-3})$, weight decay of $10^{-4}$, and $\beta{=}10^{-7}$. There were $N{=}10^4$ data points, the network had $m{=}10$ neurons, and the network was trained over $100$ epochs.
 
For OpenAI's GLUE-ColA input embeddings, $175$ epochs and $m=10$ were used. For Bert GLUE-ColA, $125$ epochs and $m=5$ were used. For OpenAI and Bert GLUE-QQP, $150$ epochs and $m=10$ were used. For OpenAI and Bert IMDB, $100$ epochs and $m=10$ were used. These parameters were chosen based on validation set results. 

The networks are trained on the \nc{} problem, and their performance is comparable to results in \citep{Wang:2024}. What distinguishes this experiment is demonstrating that networks learn novel features consistent with \lasso{} models.

Networks trained on both OpenAI and Bert embeddings exhibit similarities to \lasso{} features \Cref{fig:2dabsval}.
However, the networks trained on OpenAI embeddings (\Cref{fig:3lyrOpenAIIMDB}) have higher accuracy and closer 
 matches to \lasso{} features than those trained using Bert. The higher-performing networks trained using GPT4, such as \Cref{fig:3lyrOpenAIIMDB}, have striking similarities to \lasso{} features \Cref{fig:2dabsval}. This is consistent with the network's equivalence to the \lasso{} model, as a network that is optimal in the \lasso{} problem is globally optimal. The figures show that when networks get deeper, they change by increasing their \kink{}s, with more \kinkpoint{}s appearing in the plots.

%\begin{figure}
%    \centering
%    \begin{minipage}[t]{0.5\textwidth}
%    \includegraphics[width=\linewidth]{results/glue-cola/OpenAI/L=3/W1plot.pdf}
%    \end{minipage}
%    \begin{minipage}[t]{0.5\textwidth}
%    \includegraphics[width=\linewidth]{results/glue-cola/OpenAI/L=4/W1plot.pdf}
%    \end{minipage}
%    \caption{Neural network classifier with absolute value activation trained on text embeddings of GLUE-COLA outputted by OpenAI GPT4. Top: 3 layers. Bottom: 4 layers. The predictions (correct or incorrect) are plotted as a function of the projected input $\mathbf{W}^{(1)}\mathbf{x}$, where $\mathbf{x}$ is an embedding vector.}
%    \label{fig:gluecola_openaiL=3absval}
%\end{figure}

%\begin{figure}
%    \centering
%    \begin{minipage}[t]{0.5\textwidth}
%    \includegraphics[width=\linewidth]{results/glue-qqp/OpenAI/L=3/W1plot.pdf}
%    \end{minipage}
%    \begin{minipage}[t]{0.5\textwidth}
%    \includegraphics[width=\linewidth]{results/glue-qqp/OpenAI/L=4/W1plot.pdf}
%    \end{minipage}
%    \caption{Neural network classifier with absolute value activation trained on text embeddings of GLUE-QQP outputted by OpenAI GPT4. Top: 3 layers. Bottom: 4 layers. The predictions (correct or incorrect) are plotted as a function of the projected input $\mathbf{W}^{(1)}\mathbf{x}$, where $\mathbf{x}$ is an embedding vector.}
%    \label{fig:glueqqp_openaiL=3absval}
%\end{figure}

\section{Neural net architecture}
\textbf{Note:}
The notation and some techniques for convexification  are similar to \citep{1DNN} and \citep{pilanci2023complexity}. For finding the \lasso{} features, we can assume that the data matrix is full column rank \citep{pilanci2023complexity}. All of the \lasso{} equivalences hold provided that $m^* \geq \|\mathbf{z}\|_0$.

Let $L \geq 2, m_0=d, m_{L-1}=1$ and $m_l \in \mathbb{N}$ for $l \in [L]-\{L-1\}$. 
A neural network \eqref{eq:nn_nonrecursive} can be recursively defined as
\begin{equation*}
    \nn{L}{\mathbf{x}}  = \displaystyle \sum_{i=1}^{m_{L}}  \Xul{i}{L} \sWul{i}{L} +  \extbias,  
\end{equation*}
 where $\Xul{i}{L}$ is defined recursively as
 \begin{equation}\label{eq:Xrecurse}
     \Xul{i}{l+1}= \sigma \left( \Xul{i}{l}\Wul{i}{l}+\bul{i}{l} \right),
 \end{equation}
 with initial condition
 \begin{equation*}
     \Xul{i}{1} = \mathbf{x}.
 \end{equation*}
 We can view $ \Xul{i}{l}$ as an input to to the $i^{\text{th}}$ \textit{unit} in layer $l$. 
 %For ease of notation, we will also denote $\Wul{i}{L}=\sWul{i}{L}$. %The parameter set is $\params = \left\{\Wul{i}{l}, \bul{i}{l} \extbias: i \in [m_L], l \in [L] \right\}$.
%For $i \in [m_{L}]$, the \textit{internal} weights and biases of layer $l \in [L-1]$ are $\Wul{i}{l} \in \mathbb{R}^{m_{l-1} \times m_{l}}$ and $\bul{i}{l} \in \mathbb{R}^{1 \times m_l}$. 
% The \textit{external} weights and biases are $w^{(i)}, \extbias \in \mathbb{R}$. 
Note that for $i \in [m_L], l \in [L]$, we have $ \Xul{i}{l}\in \mathbb{R}^{1 \times m_{l-1}}$. The set of regualarized parameters is $\regparams = \left\{\Wul{i}{l}, \alpha_i: i \in [m_L], l \in [L-1]  \right \} \subset \params$. 
%For ease of notation, we also let $\alpha_i=\Wul{i}{L}$. 
All functions extend to vector and matrix-valued inputs elementwise.

The training problem \eqref{eq:trainingprob_explicit} can be stated generally as 

\begin{equation*}
    \min_{\params \in \paramspace} \lossfunc\left( \nn{L}{\mathbf{X}}\right) + \frac{\beta}{L}\regfunc(\regparams),  
\end{equation*}

where $\lossfunc:\mathbb{R}^N \to \mathbb{R}$ is called the \textit{loss} function that is convex and parameterized by $\mathbf{y}$, and $\regfunc(\params)$ is a regularization term that penalizes large weight magnitudes. In \eqref{eq:trainingprob_explicit}, the loss function is $\lossfunc(\cdot) = \frac{1}{2}\left\|\cdot - \mathbf{y}\right\|_2^2$.
%We assume $\lossfunc$ is convex, for example, $l_2$ loss: $\lossfunc(\mathbf{z})=\frac{1}{2} \| \nn\left({\mathbf{X}}\right)  - \mathbf{y} \|^2_2 $.
We can omit the external bias $\extbias$ from the network $\nn{L}{\mathbf{X}}$ by absorbing it into the loss $\lossfunc$. We assume $\regfunc(\regparams)$ is of the form
\begin{equation}\label{eq:reg_as_sum}
   \regfunc(\regparams) =  \sum_{i=1}^{m_L}  \left(\left( \regfunc^{(i,L)}(\alpha_i) \right)^L + \sum_{l=1}^{L-1 } \left( \regfunc^{(i,l)}\left(\Wul{i}{l}\right) \right)^L \right)
\end{equation}
 where $\regfunc^{(i,l)}$ is a nonnegative function that is positively homogeneous, i.e., for any positive scalar $a$, $\regfunc^{(i,l)}\left(a \mathbf{W}\right) = a \regfunc^{(i,l)}\left(\mathbf{W}\right) $.
In \eqref{eq:trainingprob_explicit}, the regularization is $\regfunc^{(i,l)}\left( x \right) =  \left \|x\right\|_{1}$. %in this paper weights are all scalars or vectors anyway %= \left( \sum_{j \in [m_{l-1}]} \sum_{k \in [m_l]} \left| \Wul{i}{l}_{j,k} \right|^{p_l} \right)^{\frac{1}{p_l}}$. 

%todo: find suitable L

\begin{lemma}
    The training problem is equivalent to the \textit{rescaled} problem
    \begin{equation}\label{eq:noncvxtrain_genLlyrLLnorm_rescale}
    \min_{\params \in \paramspace: \regfunc^{(i,l)}\left( \Wul{i}{l}\right) = 1} \lossfunc\left(\nn{L}{\mathbf{X}}\right) + \beta  \sum_{i=1}^{m_L}\regfunc^{(i,L)}\left(\alpha_i\right). 
    \end{equation}
\end{lemma}

\begin{proof}
By the AM-GM inequality on \eqref{eq:reg_as_sum}, a lower bound on the training problem is

\begin{equation}\label{eq:rescaledlb}
    \min_{\params \in \paramspace} \lossfunc\left(\nn{L}{\mathbf{X}} \right) + \beta \sum_{i=1}^{m_L}  \regfunc^{(i,L)}(\alpha_i) \prod_{l=1}^{L-1} \regfunc_{(i,l)}\left(\Wul{i}{l}\right).
\end{equation}

Consider the minimization problem

\begin{equation}\label{eq:rescaledlboneconstr}
    \min_{\params \in \paramspace:\regfunc^{(i,l)}\left(\Wul{i}{l}\right)=1} \mathcal{L}_y \left(\nn{L}{\mathbf{X}} \right) + \beta \sum_{i=1}^{m_L}  \regfunc^{(i,L)}(\alpha_i) \prod_{l=1}^{L-1} \regfunc_{(i,l)}\left(\Wul{i}{l}\right).
    %\regfunc_{(i,L)}\left(\params^{(i,L)}\right).
\end{equation}

Problem \eqref{eq:rescaledlboneconstr} is an upper bound on \eqref{eq:rescaledlb}. Given optimal $\left\{\Wul{i}{l}, \alpha_i \right\}$ in \eqref{eq:rescaledlb}, the rescaled parameters ${\Wul{i}{l}}' {=}  \Wul{i}{l}/\regfunc^{(i,l)}\left(\Wul{i}{l}\right)$ and $\alpha_i' {=} \alpha_i \prod_{l=1}^{L-1} \regfunc^{(i,l)}\left(\Wul{i}{l}\right)$ (and rescaled bias parameters) achieve the same objective in \eqref{eq:rescaledlboneconstr}. 
Hence \eqref{eq:rescaledlboneconstr} and \eqref{eq:rescaledlb} are equivalent. Given an optimal parameter $q \in \left\{\Wul{i}{l},\alpha_i\right\}$ in \eqref{eq:rescaledlboneconstr}, the rescaled parameters $ q' {=} |\alpha_i|^{\frac{1}{L}}q$ (and rescaled bias parameters) achieve the same objective in the training problem, which is therefore equivalent to \eqref{eq:rescaledlboneconstr}. Simplifying \eqref{eq:rescaledlboneconstr} gives \eqref{eq:noncvxtrain_genLlyrLLnorm_rescale}.
\end{proof}

Let $\mathbf{X}^{(L)}=\mathbf{X}^{(1,L)}$. Assume $\regfunc^{(i,l)}\left(x\right)=\|x\|_1$. %could be component wise for paravector
\begin{lemma}
    A lower bound on the rescaled training problem %\eqref{eq:gen_rescale} 
    is 
    % need to say rescaled so that theta and function is clear
    \begin{equation}\label{deepgen_dual}
		\begin{aligned}
			\max_{\lambda \in \mathbb{R}^N} & -\mathcal{L}_{\mathbf{y}}^*( \lambda ) \quad 
			\text{ s.t. }                   &
			\max_{\params \in \paramspace}
			\left|\lambda^T  \mathbf{X}^{(L)}  \right| \leq \beta,
		\end{aligned}
	\end{equation}
  where $f^*(\mathbf{x}):=\max_{\mathbf{x}}  \left \{ \mathbf{z}^T \mathbf{x} - f(\mathbf{x}) \right \}$ is the convex conjugate of $f$.
\end{lemma}
\begin{proof}
  Find the dual of the rescaled training problem by rewriting it as
	\begin{equation}\label{deepgen_prep_to_dual}
		\begin{aligned}
			% \min_{\alpha \in \mathbb{R}^{m_L}, W^{(i,l)}: ||W^{(i,l)}_{1:m_{1-1}, 1:m_l}||_F=1, i \in [m_L], l \in [L-1]} 
			\min_{\params \in \paramspace}\,
			              & \mathcal{L}_{\mathbf{y}}( \mathbf{z} ) + \beta ||\boldsymbol{\alpha}||_1, \quad
			\text{ s.t. } & \mathbf{z} =  \sum_{i=1}^{m_L} \alpha_i \Xul{i}{L}  .
		\end{aligned}
	\end{equation}
	The Lagrangian of problem \eqref{deepgen_prep_to_dual} is
%
	%\begin{equation}\label{deepgen_lag}
		%\begin{aligned}
			% L\left(\lambda,\alpha, W^{(i,l)}: i \in [m_L], l \in [L-1]\right) 
   $\displaystyle 
			L\left(\lambda, \params \right)
			  = \mathcal{L}_{\mathbf{y}}( \mathbf{z} ) + \beta ||\boldsymbol{\alpha}||_1  - \lambda^T \mathbf{z} + \sum_{i=1}^{m_L} \lambda^T  \Xul{i}{L}   \alpha_i.  
    $
		%\end{aligned}
	%\end{equation}
%
	Minimize the Lagrangian over $\mathbf{z}$ and $\boldsymbol{\alpha}$ and use Fenchel duality \citep{boyd:2004}. The dual of \eqref{deepgen_prep_to_dual} is
	\begin{equation}\label{deepgen_dual_int}
		\begin{aligned}
			\max_{\lambda \in \mathbb{R}^N} & -\mathcal{L}_{\mathbf{y}}^*( \lambda ) \quad 
			\text{ s.t. }                   &
			% \max_{W^{(i,l)}: ||W^{(i,l)}_{1:m_{1-1}, 1:m_l}||_F=1, l \in [L-1]} 
			\max_{\params \in \paramspace}
			\left|\lambda^T   \Xul{i}{L}\right| \leq \beta, i \in [m_L].
		\end{aligned}
	\end{equation}
	Observe $ \Xul{i}{L}$ is of the same form for all $i \in [m_L]$. So the $m_L$ constraints in \eqref{deepgen_dual_int} collapse to just one constraint. Then we can write \eqref{deepgen_dual_int} as  \eqref{deepgen_dual}.  
\end{proof}

%Suppose the regularization is $L_1$ norm.

\section{\reflectnn}\label{app:maintheorems}
%can cite thorems in simod paper

Let $n^{(1)}, {\cdots}, n^{(L{-}1)} {\in} [N]$.
Define the \emph{\kink{} sets} $\mathcal{K}^{(L)} {=} \{\mathbf{x}_{n^{(L{-}1)}}\}$ if $L{=}2$
and $\mathcal{K}^{(L)}{=}\left\{\xl{L{-}2}_{n^{(L{-}2)}}, \frac{\xl{L{-}2}_{n^{(L{-}1)}}{+}\xl{L{-}2}_{n^{(L{-}2)}}}{2}\right\}$ otherwise.
%
%For $n^{(1)}, n^{(2)} \in [N]$, when $L=2$ let $\mathcal{K}_{n^{(1)}}=\{\mathbf{x}_{n^{(1)}}\}$ and when $L=3$, let 
%
%\begin{equation}\label{eq:Kn1n2}
%   \mathcal{K}_{n^{(1)},n^{(2)}}{=}\left\{\mathbf{x}_{n^{(1)}}, \frac{\mathbf{x}_{n^{(2)}}{+}\mathbf{x}_{n^{(1)}}}{2}\right\}. 
%\end{equation}
%
For $\xprime {\in} \mathcal{K}^{(3)}$, let $\mathbf{M}_1 {=} \mathbf{X}{-}\mathbf{1}\xprime$ and

\begin{equation}\label{eq:signpatternfunctions}
    \begin{aligned}
        &f_{\xprime}: \mathbb{R}^d {\to} \mathbb{R}^N, &f_{\xprime}(\mathbf{W})&{=}\mathbf{M}_1\mathbf{W}  \\
        &&&{=} \left(\mathbf{X}-\mathbf{1}\xprime\right)\mathbf{W}\\
        %&f_{\mathbf{d}^{(1)},\mathbf{a},n^{(2)}}:\mathbb{R}^d {\to} \mathbb{R}^N, &f_{\mathbf{d}^{(1)},\mathbf{a},n^{(2)}}(\mathbf{W})&{=}   \mathbf{d}^{(1)} {\cdot} f_\mathbf{a}(\mathbf{W}) {-} d^{(1)}_{n^{(2)}} f_\mathbf{a}(\mathbf{W})_{n^{(2)}}  \mathbf{1} \\
        %&&&{=}\left(\text{Diag}(\mathbf{d}^{(1)})-d^{(1)}_{n^{(2)}} \mathbf{E}_{n^{(2)}} \right)f_\mathbf{a}\left(\mathbf{W}\right)\\
        &S_{\xprime}:\mathbb{R}^d {\to} \{-1,1\}^N, &S_{\xprime}(\mathbf{W})&{=}\text{sign}\left(f_{\xprime}(\mathbf{W})\right).
        %&S_{\mathbf{d}^{(1)},\mathbf{a}, n^{(2)}}{:}\mathbb{R}^d {\to} \{-1,1\}^{N\times N}, &S_{\mathbf{d}^{(1)},\mathbf{a},n^{(2)}}(\mathbf{W})&{=}\text{sign}\left( f_{\mathbf{d}^{(1)},\mathbf{a},n^{(2)}}(\mathbf{W}) \right)\\
    \end{aligned}
\end{equation}

For $\mathbf{d}^{(1)} \in S_{\xprime}\left(\mathbb{R}^d\right)$, let $\mathbf{M}_2 {=} \left(\text{Diag}(\mathbf{d}^{(1)}){-}d^{(1)}_{n^{(2)}} \mathbf{E}_{n^{(2)}} \right)\mathbf{M}_1$ and

\begin{equation}\label{eq:signpatternfunctions2}
    \begin{aligned}
        %&f_\mathbf{a}: \mathbb{R}^d {\to} \mathbb{R}^N, &f_\mathbf{a}(\mathbf{W})&{=}\left(\mathbf{X}-\mathbf{1}\mathbf{a}\right)\mathbf{W}  \\
        &f_{\mathbf{d}^{(1)},\xprime,n^{(2)}}:\mathbb{R}^d {\to} \mathbb{R}^N, &f_{\mathbf{d}^{(1)},\xprime,n^{(2)}}(\mathbf{W})&{=} \mathbf{M}_2 \mathbf{W} \\
        &&&{=}  \left| \left(\xl{1}_n {-} \xprime\right)\mathbf{W}\right|-\left| \left(\xl{1}_{n^{(2)}}   {-}\xprime \right) \mathbf{W}\right| \\
        %&&&{=}
       % \left|f_\mathbf{a}(\mathbf{W}) \right| - \left|f_\mathbf{a}(\mathbf{W})_{n^{(2)}} \mathbf{1} \right|\\
        %&{=}
        %\mathbf{d}^{(1)} {\cdot} f_\mathbf{a}(\mathbf{W}) {-} d^{(1)}_{n^{(2)}} f_\mathbf{a}(\mathbf{W})_{n^{(2)}}  \mathbf{1} \\
        %&&&{=}\left(\text{Diag}(\mathbf{d}^{(1)})-d^{(1)}_{n^{(2)}} \mathbf{E}_{n^{(2)}} \right)f_\mathbf{a}\left(\mathbf{W}\right)\\
        %&S_\mathbf{a}:\mathbb{R}^d {\to} \{-1,1\}^N, &S_\mathbf{a}(\mathbf{W})&{=}\text{sign}\left(f_\mathbf{a}(\mathbf{W})\right)\\
        &S_{\mathbf{d}^{(1)},\xprime, n^{(2)}}{:}\mathbb{R}^d {\to} \{-1,1\}^{N\times N}, &S_{\mathbf{d}^{(1)},\xprime,n^{(2)}}(\mathbf{W})&{=}\text{sign}\left( f_{\mathbf{d}^{(1)},\xprime,n^{(2)}}(\mathbf{W}) \right)
    \end{aligned}
\end{equation}

where $\cdot$ denotes elementwise multiplication and $\mathbf{E}_{n^{(2)}} $ denotes a $N {\times} N$ matrix whose $(n^{(2)})^{\text{th}}$ column is $\mathbf{1}$ and all other elements are $0$. 
%
%Recall the first-layer sign pattern functions \eqref{eq:signpatternfunctions}. Similarly define the second-layer sign pattern functions $S_{\mathbf{d},\mathbf{a}, n}{:}\mathbb{R}^d {\to} \{-1,1\}^{N\times N}, S_{\mathbf{d},\mathbf{a},n}(\mathbf{W}){=}\text{sign}\left(   \mathbf{d} {\cdot} f_\mathbf{a}(\mathbf{W}) {-} d_n f_\mathbf{a}(\mathbf{W})_n  \mathbf{1} \right){=}\text{sign}\left(\left(\text{Diag}(\mathbf{d})-d_n \mathbf{E}_{n^{(2)}} \right)f_\mathbf{a}\left(\mathbf{W}\right)\right)$, 
%where $\cdot$ denotes elementwise multiplication and $\mathbf{E}_{n^{(2)}} $ denotes a $N {\times} N$ matrix whose $(n^{(2)})^{\text{th}}$ column is $\mathbf{1}$ and all other elements are $0$. 
These functions \eqref{eq:signpatternfunctions}, \eqref{eq:signpatternfunctions2} represent the sign patterns of a neuron in the first and second layers.
%
%Given $n^{(1)} {\in} [N]$, let $ \Tilde{\mathcal{X}}^{(2)}{=} \Delta \mathcal{X}^{(2)}_\perp$ (\Cref{section:lasso_equivalence}).
%
Given $n^{(1)}, n^{(2)} {\in} [N], \xprime {\in} \mathcal{K}^{(3)}$, a fixed sign pattern $\mathbf{d}^{(1)} \in S_{\xprime}\left(\mathbb{R}^d\right)$, let $\wl{1} {\in} S^{-1}_{\xprime}\left(\mathbf{d}^{(1)}\right)$ and
% $\mathbf{d}=\text{sign}\left(\left(\mathbf{X}-\mathbf{1}\xprime\right)\mathbf{W}\right)$,
%$\mathbf{d} \in S_\mathbf{a}\left(\mathbb{R}^d\right)$ 

\begin{equation}\label{eq:augdataset}
   \begin{aligned}
      % % \Tilde{\mathcal{X}}^{(2)}&{=} \left\{\frac{\times_{i=1}^{d{-}1} \Delta {\mathbf{x}}_i }{ \left\|\times_{i{=}1}^{d-1} \Delta{\mathbf{x}}_i \right\|_1}: \Delta{\mathbf{x}}_i \in \left\{ \mathbf{x}_n {-}\mathbf{x}_n^{(1)}: n^{(1)} {\in} [N] \right\} \cup \mathcal{E}\right\}  \\
       %\Tilde{\mathcal{X}}^{(3)}&{=}  \left\{\frac{\times_{i=1}^{d{-}1} \Delta {\mathbf{x}}_i }{ \left\|\times_{i{=}1}^{d-1} \Delta{\mathbf{x}}_i \right\|_1} : \Delta {\mathbf{x}}_i 
 %\in \left\{ \mathbf{x}_1{-}\xprime, {\cdots}, \mathbf{x}_N{-}\xprime, 
   %    \shiftx(\mathbf{x}_1), \cdots, \shiftx(\mathbf{x}_N)
       %&d_1 \mathbf{x}_1 {-} d_{n^{(2)}} \mathbf{x}_{n^{(2)}} {-} \left(d_1 {-} d_{n^{(2)}}\right)\mathbf{a}, {\cdots}, d_N \mathbf{x}_N {-} d_{n^{(2)}} \mathbf{x}_{n^{(2)}} {-} \left(d_N {-} d_{n^{(2)}}\right)\mathbf{a}, \\
     %  \right\} \cup \mathcal{E}\right\}\\
        \mathcal{X}^{(3)}&{=}   \left\{ \mathbf{x}_n{-}\xprime\right\}_n \cup \left\{\mathbf{x}_n{-}\mathbf{x}_{n^{(2)}}\right\}_n \cup
       \left\{\mathbf{x}_n {-} \reflect{\mathbf{x}_{n^{(2)}}}{\mathbf{a}} \right\}_n \cup  \{e_1, \cdots, e_d\}\\
       %&d_1 \mathbf{x}_1 {-} d_{n^{(2)}} \mathbf{x}_{n^{(2)}} {-} \left(d_1 {-} d_{n^{(2)}}\right)\mathbf{a}, {\cdots}, d_N \mathbf{x}_N {-} d_{n^{(2)}} \mathbf{x}_{n^{(2)}} {-} \left(d_N {-} d_{n^{(2)}}\right)\mathbf{a}, \\
        \Delta \mathcal{X} &= \left\{\pm \frac{\times_{i=1}^{d{-}1} \Delta \mathbf{x}_i}{\left\|\times_{i=1}^{d{-}1} \Delta \mathbf{x}_i\right\|_1}: \Delta \mathbf{x}_i \in \mathcal{X}^{(3)}\right\}.
   \end{aligned} 
\end{equation}

%where $\shiftx{}$ is defined in \eqref{eq:x_d}. %$\Tilde{\mathcal{X}}^{(L)}$ is the \textit{hyperplane-specific}  orthogonal difference data set (\Cref{section:lasso_equivalence}).

%Then \eqref{eq:XL_nexpand} can be rewritten as
%
%\begin{equation}\label{eq:XL_nexpandfa}
%    \xl{L}_n = \left|  S_\mathbf{a}\left(\mathbf{W}^{(1)}\right)_n S_{\mathbf{a}}(\mathbf{W})_{n^{(2)}} f_\mathbf{a}\left(\mathbf{W}^{(1)}\right)_n  -f_\mathbf{a}\left(\mathbf{W}^{(1)}\right)_{n^{(2)}}   \right|.
%\end{equation}
%
%can we pull out S_a and then divide based on other value - that also sort of determines S_a though?
%      
Let $\delta B {=} \left\{\mathbf{W} \in \mathbb{R}^d: \|\mathbf{W}\|_1 {=} 1 \right\}$ be the boundary of the $l_1$ ball of radius $1$. 

%\begin{theorem}\label{thrm:3layer}
%    The training problems for $2$ and $3$-layer \reflectnn{}s and $d$-dimensional data is equivalent to a \lasso{} problem where the dictionary columns (indexed by $j$) correspond to all %subsets $\left\{\tilde{\mathbf{x}}_1,{\cdots},\tilde{\mathbf{x}}_{d-1}\right\}$ of all possible augmented data sets, with
%    possible feature functions,
%    \begin{equation}\label{eq:Aij}
%        \begin{aligned}
%            \mathbf{A}_{i,j} &= 
%            f(\mathbf{x}_i)
            %%\frac{\text{Vol}\left(\tilde{\mathbf{x}}_d, \Tilde{\mathbf{x}}_1, \cdots, \tilde{\mathbf{x}}_{d-1} \right)}{\left\|\times_{i=1}^{d-1} \Tilde{\mathbf{x}}_i \right\|_1}
 %       \end{aligned}
 %   \end{equation}

%where $f(\mathbf{x}_i)$ is the corresponding feature function, provided that $m \geq \|\mathbf{z}^*\|_0$, where $\mathbf{z}^*$ is optimal in the \lasso{} problem.
  %  where the row index $i$ determines %don't reuse i

%\textcolor{red}{todo: give complexity}
%\begin{equation}\label{eq:x_d}
%    \begin{aligned}
%        \tilde{\mathbf{x}}_d &= 
%        \begin{cases}
%            \mathbf{x}_i - \mathbf{x}_{n^{(2)}} &\mbox{ if } \mathbf{x}_i-\mathbf{a}, \mathbf{x}_{n^{(2)}}-\mathbf{a} \mbox{ are on the same side of } \times_{i=1}^{d-1} \Tilde{\mathbf{x}}_i\\
%            \mathbf{x}_i - \reflect{\mathbf{x}_{n^{(2)}}}{\mathbf{a}} &\mbox{ else. }
%        \end{cases}
%    \end{aligned}
%\end{equation}

%\end{theorem}

%%%%%
\begin{proof}[Proof of \Cref{thrm:3layersimpler}]\label{proof_3lyrsimpler}

In a \reflectnn{} \eqref{eq:deepnarrowabsvalnetwork}, $\Wul{i}{2}, {\cdots}, \Wul{i}{L-1}, \bl{l}, {\cdots}, \bl{L-1}$ are scalar-valued and $\sigma(x){=}|x|$. For fixed $\wl{1},{\cdots},\wl{L{-}1}$, $\bl{l}{=}{\bl{l}}^*$ is \emph{optimal} if ${\bl{l}}^* {\in} \arg\max_{\bl{l}}\max_{\bl{l+1}} {\cdots} \max_{\bl{L-1}}\left| \sum_{n{=}1}^N \lambda_n \xl{L}_n \right|$, and analogously for $\wl{l}$. We will analyze the constraint of the dual \eqref{deepgen_dual}:

\begin{equation}\label{eq:dual_constraint}
    \max_{\params \in \paramspace}
			\left|\lambda^T  \mathbf{X}^{(L)}  \right| \leq \beta.
\end{equation}

Satisfying \eqref{eq:dual_constraint} requires $\mathbf{1}^T \lambda {=} 0$, otherwise $\max_{\params \in \paramspace} \left|\lambda^T  \mathbf{X}^{(L)} \right|{=}\infty$. So assume $\mathbf{1}^T \lambda {=} 0$. We will use the following property: if $f:\mathbb{R}\to\mathbb{R}$ is a bounded piecewise linear function, $\arg\max_x |f(x)|$ contains a \kink{} (or a \emph{breakpoint} in $\mathbb{R}$) of $f$. 
%Suppose $m_{L-2}=1$. %Then $\wl{L-1}, \bl{L-2}$ are scalars. check todoa
The objective in the dual constraint \eqref{eq:dual_constraint} is $\left|\lambda^T \mathbf{X}^{(L)}\right| {=} \left|\sum_{n=1}^N \lambda_n \xl{L}_n\right|$ where 
\begin{equation}\label{eq:XL_bL-1}
    \xl{L}_n{=} \left| \mathbf{X}^{(L{-}1)}_n \wl{L{-}1} {+}\bl{L{-}1}\right|.
\end{equation}
 The {\kink}s of $\xl{L}_n$ as a function of $\bl{L-1}$ occur at  $\bl{L-1}{=} {-}\mathbf{X}^{(L{-}1)}_n \wl{L-1}$.
%, and by \Cref{rk:finiteconstraint} and the assumption involving $\lambda^T \mathbf{1}=0$, these {\kink}s contain an optimal $b^{(L-1)}$.
Therefore for some $n^{(L{-}1)} {\in} [N]$, $ {\bl{L{-}1}}^* {=}{-}\mathbf{X}^{(L{-}1)}_{n^{(L{-}1)}} \wl{L{-}1} $ is optimal. Now suppose $L{>}2$.
%a $b^{(L-1)}$ that zeros out the argument of an activation that depends on it, i.e. $b^{(L-1)} = -\mathbf{X}^{(L-2)}_{n^{(L-1)}} w^{(L-1)}$ for some $n^{(L-1)} \in [N]$. 
Plugging ${\bl{L{-}1}}^*$ into $\xl{L}_n$ \eqref{eq:XL_bL-1} gives 
\begin{equation}\label{eq:lambda^TX_L-1}
    \begin{aligned}
         \mathbf{X}^{(L)}_n   &{=}  \left| \left( \mathbf{X}^{(L{-}1)}_n{-}\mathbf{X}^{(L{-}1)}_{n^{(L{-}1)}}  \right)\wl{L{-}1} \right|   \\
                                        &{=}  \left|  \left| \mathbf{X}^{(L{-}2)}_n \wl{L{-}2}{+} \bl{L{-}2}\right|-\left| {\mathbf{X}^{(L{-}2)}_{n^{(L{-}1)}}} \wl{L{-}2}  {+} \bl{L{-}2}\right|  \right|.
    \end{aligned}
\end{equation}

%todo w2 can be 1

By \eqref{eq:lambda^TX_L-1}, for fixed $n^{(L{-}1)}$, the set of \kink{}s of $\sum_n \lambda_n \xl{L}_n$ as a function of $\bl{L{-}2}$ is $\left\{-\mathbf{x}'_{n^{(L{-}2)}}\wl{L{-}2}: \mathbf{x}'_{n^{(L{-}2)}}{\in} \mathcal{K}^{(L)}, n^{(L{-}2)} {\in} [N]\right\}$. 
%= \sum_{n=1}^N \lambda_n \sigma \left( c^{(L-1)}_{n,n^{(L-1)}} \right)
%
%
%Now assume $\sigma(x)$ is leaky ReLU. Since $m_{L-2}=1$, $\bl{L-2}$ is a scalar, so let $x=b^{(L-2)}, \alpha = \mathbf{X}^{(L-2)}_n w^{(L-2)}, \beta = \mathbf{X}^{(L-1)}_n w^{(L-2)}, s = w^{(L-1)}$ in \Cref{theorem:kinks_general}. 
%Assume $\sigma(x)=|x|$. 
So there exists $n^{(L{-}2)}{\in}[N]$ and $\mathbf{x}'_{n^{(L{-}2)}}{\in}\mathcal{K}^{(L)}$ %\eqref{eq:Kn1n2} 
such that 

\begin{equation}\label{eq:optbL-2}
   {\bl{L{-}2}}^*{=}{-}\mathbf{x}'_{n^{(L-2)}}\wl{L{-}2}
\end{equation}

 is optimal. %proof for discrete finite set of features for any depth. 

 %todo say 2 layer

%remember to do lamdna^T1 =0
%Let $\mathbf{d}\left(\wl{1},{\bl{1}}\right) =   \text{sign} \left( \mathbf{X}\wl{1}+ \bl{1}   \right) \in \{-1,0,1\}$. 

%\textbf{Now we partition}. 
%For $\mathbf{j},a \in \tilde{\theta}$, let $\mathcal{D}_{\mathbf{j},\mathbf{a}} = \left \{ \text{Diag} \left( \mathbf{1} \left\{ \left(\mathbf{X}- \mathbf{1a} \right) \mathbf{w} \right\} \right): \mathbf{w} \in \mathbb{R}^d \right\} $. Enumerate $\left \{D_{\mathbf{j},\mathbf{a},1}, \cdots D_{\mathbf{j},\mathbf{a}, P} \right \} = \mathcal{D}_{\mathbf{j},\mathbf{a}}$. For $p \in [P]$, let $\mathcal{C}_{\mathbf{j},\mathbf{a},p} = \left \{\mathbf{w} \in \mathbb{R}^d: \left(2D_{\mathbf{j},\mathbf{a},p} -I\right) \left( \mathbf{X}-\mathbf{1}\mathbf{a} \right) \mathbf{w} \geq 0  \right \}$. 

%There are a finite number of such $\mathbf{d}\left(\wl{1},{\bl{1}}\right)$. %, at most $3^N$ for fixed $\wl{1},\bl{1}$.

Now suppose $L{=}3$. Let $\xprime \in \mathcal{K}_{n^{(1)},n^{(2)}}$.
%Then $\mathcal{K}_{n^{(L{-}1)},n^{(L{-}2)}}{=}\mathcal{K}_{n^{(1)},n^{(2)}}$ \eqref{eq:Kn1n2} and ${\bl{1}}^*{=}{-}\mathbf{a}\wl{1}$ is optimal. 
By \eqref{eq:optbL-2}, $\bl{1}{=}-\xprime \wl{1}$ and plugging this into \eqref{eq:lambda^TX_L-1} gives  
\begin{equation}\label{eq:XL_nexpand}
\begin{aligned}
        \xl{3}_n &= \left|  \left| \left(\xl{1}_n {-} \xprime\right)\wl{1}\right|-\left| \left(\xl{1}_{n^{(2)}}   {-}\xprime \right) \wl{1}\right|  \right| %= \left| \shiftx\left(\xl{1}_n\right)\wl{1}\right|.
      %  &= \begin{cases}
      %      \left | \right(\xl{1}_n -\xl{1}_{n^{(2)}} \left)\wl{1}\right| &\mbox{ if } \xl{1}_n -\mathbf{a}, \xl{1}_{n^{(2)}}-\mathbf{a} \text{ are on the same side of } \wl{1}\\
       %     \left| \left( \xl{1}_n - \reflect{\xl{1}_{n^{(2)}}}{\mathbf{a}}\right) \wl{1} \right| &\mbox{ else }.
        %\end{cases}
\end{aligned}
\end{equation}
By \eqref{eq:XL_nexpand}, the dual constraint \eqref{eq:dual_constraint} is equivalent to the following: %for all $\wl{2} {\in} \{-1,1\}, n^{(1)},n^{(2)}{\in}[N], \mathbf{a}{\in}\mathcal{K}_{n^{(1)},n^{(2)}}, \mathbf{d}^{(1)} {\in} \mathcal{D}^{(1)}\left(\mathbf{a}\right), \mathbf{d}^{(2)} {\in} \mathcal{D}^{(2)}\left(\mathbf{d}^{(1)},\mathbf{a},n^{(2)}\right),\wl{1} {\in} \mathcal{P}^{(1)}\left(\mathbf{d}^{(1)},\mathbf{a}\right)\cap \mathcal{P}^{(2)}\left(\mathbf{d}^{(2)},\mathbf{d}^{(1)},
%\mathbf{a},n^{(2)}\right)$, %\bl{1}{=}{-}\mathbf{a}\wl{1}, 
%\\ d^{(2)}{=}\text{sign}\left( \left(  \left( \mathbf{X}_n \wl{1}{+} \bl{1}\right)d^{(1)}_{n}{-}  \left( {\mathbf{X}_{n^{(2)}}} \wl{1}  {+} \bl{1}\right)d^{(1)}_{n^{(2)}} \right)\right)$,

\begin{equation}\label{eq:L=234lyr_max_constr}
\scalemath{0.8}{
  \begin{aligned} 
   % \text{ for all }
    %k \in \{0,1\}, \mathbf{s} \in \{-1,1\}^{L-1}, \mathbf{j} \in [N]^{L-1}, 
    %\left(\wl{1}, \cdots, \wl{L-1}, \bl{1}, \cdots, \bl{L-1} \right) \in \paramspace^*,
    %\mathbf{a} %=\xl{L}(\mathbf{X}) 
    %\in \featureset,
    %\quad &  
     & \max_{\wl{2} {\in} \{-1,1\}}
     \max_{n^{(1)},n^{(2)} {\in} [N] } 
    \max_{{\xprime} {\in} \mathcal{K}^{(3)}} \max_{\mathbf{d}^{(1)} {\in} 
    %\mathcal{D}^{(1)}(\mathbf{a})
    S_{\xprime}\left(\mathbb{R}^d\right)
    }
    \max_{\mathbf{d}^{(2)} {\in} 
    S_{\mathbf{d}^{(1)},{\xprime},n^{(2)}}\left(\mathbb{R}^d\right)
    %\mathcal{D}^{(2)}(\mathbf{d}^{(1)},\mathbf{a},n^{(2)})
    }
    \max_{\wl{1} {\in} %\mathcal{P}^{(1)}\left(\mathbf{d}^{(1)},\mathbf{a}\right)
    S_{\mathbf{a}}^{-1}(\mathbf{d}^{(1)})
    {\cap} 
    S_{\mathbf{d}^{(1)},{\xprime},n^{(2)}}^{-1}(\mathbf{d}^{(2)})  %\mathcal{P}^{(2)}\left(\mathbf{d}^{(2)},\mathbf{d}^{(1)},\mathbf{a},n^{(2)}\right)
{\cap} \delta B }  
\left | \sum_{n{=}1}^N \lambda_n \xl{L}_n \right| {\leq} \beta, \\
    %&\text{where }
    %\mathbf{X}^{(L)}_n =
   %   \left(  \left( \left(\mathbf{X}_n {-} \mathbf{a} \right)\wl{1} \right)d^{(1)}_{n} {-}  \left(\left( {\mathbf{X}_{n^{(2)}}} {-} \mathbf{a}\right)\wl{1}  \right)d^{(1)}_{n^{(2)}} \right)  \mathbf{d}^{(2)}_{n}\\
    &\mathbf{1}^T \lambda = 0.
 %   \mbox{ for } l \in [L-1]  \\
  \end{aligned}
  }
  \end{equation}

For each fixed $\mathbf{d}^{(1)} {\in} S_{\xprime}(\mathbb{R}^d)$ and $ \mathbf{d}^{(2)} {\in} 
S_{\mathbf{d}^{(1)},\xprime,n^{(2)}}\left(\mathbb{R}^d\right)$, the objective term $\xl{L}_n{=}\left(  \left( \left(\mathbf{X}_n {-} \xprime\right)\wl{1} \right)d^{(1)}_{n} {-}  \left(\left( {\mathbf{X}_{n^{(2)}}} {-} \xprime\right)\wl{1}  \right)d^{(1)}_{n^{(2)}} \right)  \mathbf{d}^{(2)}_{n}$ is linear in $\wl{1}$.
%Now suppose $r^{(1)}\left(\wl{1}\right)=\left\|\wl{1}\right\|_1$. 
Let $B {=} \left\{\mathbf{W} \in \mathbb{R}^d: \|\mathbf{W}\|_1 {\leq} 1 \right\}$ be the filled $l_1$ ball of radius $1$.
If we replace $\delta B$ by $B$ in \eqref{eq:L=234lyr_max_constr},
%If we redefine the sets $\mathcal{P}^{(1)}(\mathbf{d},\mathbf{a}){=}\left(S_{\mathbf{a}}^{-1}(\mathbf{d}) \cap \overline{B}\right), \mathcal{P}^{(2)}(\mathbf{d},\mathbf{d}^{(1)},\mathbf{a},n){=}S_{\mathbf{d}^{(1)},\mathbf{a},n}^{-1}(\mathbf{d}) \cap \overline{B}$, 
%If the constraint $\|\wl{1}\|_1{=}1$ is relaxed to $\|\wl{1}\|_1{\leq}1$ in the definition of $\mathcal{P}^{(1)}(\mathbf{d},\mathbf{a})$, 
the constraint \eqref{eq:L=234lyr_max_constr} remains equivalent since the $\wl{1}$ that maximizes the left hand side of \eqref{eq:L=234lyr_max_constr} will have the largest feasible magnitude.
%Then $\mathcal{P}^{(1)}(\mathbf{d},\mathbf{a}), \mathcal{P}^{(2)}(\mathbf{d},\mathbf{d}^{(1)},\mathbf{a},n)
%$ are polytopes. %$\mathcal{P}^{(1)}\left(\mathbf{d}^{(1)},\mathbf{a}\right){=}\left\{\wl{1}\in\mathbb{R}^{d}:\text{sign}\left(\mathbf{X}\wl{1}{+}\mathbf{1}\bl{1}\right){=}\mathbf{d}^{(1)},\right\|\wl{1}\left\|_1\leq 1\right\} $, % and 
%$\mathcal{P}^{(2)}\left(\mathbf{d},\bl{1},\wl{1}\right)= \\ \left\{\wl{2} \in \mathbb{R}:  \text{sign}\left( \left(  \mathbf{d}\left(\wl{1},{\bl{1}}\right)_{n}\left( \mathbf{X}_n \wl{1}{+} \bl{1}\right)- \mathbf{d}\left(\wl{1},{\bl{1}}\right)_{n^{(2)}}\left( {\mathbf{X}_{n^{(2)}}} \wl{1}  {+} \bl{1}\right) \right) \wl{2}\right) = \mathbf{d},  \left|\wl{2}\right|\leq1\right\} = \{-1,1\}$, 
%which is a polytope. 
Then for each fixed $n^{(1)},n^{(2)},\xprime,\bl{1},\mathbf{d}^{(1)}$, the left hand side of the dual constraint \eqref{eq:dual_constraint} is the maximization of a linear function 
%$\pm \lambda^T \xl{L}$ 
of $\wl{1}$ over the polytope 
\begin{equation}\label{eq:polytope}
   \mathcal{P}{=}   S_{\mathbf{a}}^{-1}\left(\mathbf{d}^{(1)}\right){\cap} 
    S_{\mathbf{d}^{(1)},\xprime,n^{(2)}}^{-1}\left(\mathbf{d}^{(2)}\right){\cap}B 
\end{equation}
which is equivalent to the same maximization over the finite set of extremal points of $\mathcal{P}$. Let $\mathbf{E}_{n^{(2)}} $ be a $N {\times} N$ matrix whose $(n^{(2)})^{\text{th}}$ column is $\mathbf{1}$ and all other elements are $0$. The first-layer sign pattern function is the sign of a linear function of $\mathbf{W}:\text{sign}\left(\mathbf{X}-\mathbf{1}\xprime\right) $. The second-layer sign pattern function can also be written as the sign of a linear function of $\mathbf{W}$, as
$S_{\mathbf{d},\mathbf{a},n}(\mathbf{W}){=}\text{sign}\left(\left(\text{Diag}(\mathbf{d})-d_n \mathbf{E}_{n} \right)\left(\mathbf{X}-\mathbf{1}\xprime\right)\mathbf{W}\right)$, where $\cdot$ denotes elementwise multiplication. Therefore we can apply Lemma 19 in \citep{pilanci2023complexity} as follows.
Let $\mathbf{I}$ denote an identity matrix, and let  
\begin{equation*}
\mathbf{M} = 
    \begin{pmatrix}
    \mathbf{M}_1\\
        \mathbf{M}_2 \\
         \mathbf{I}
    \end{pmatrix}.
\end{equation*}
The size of $\mathbf{M}$ is $(2N{+}d){\times} d$. For $S \subset [N]$, let $\mathbf{M}_S$ denote a submatrix of $\mathbf{M}$ consisting of the rows indexed by $S$. The matrix $\mathbf{M}$ depends on $\xprime,\mathbf{d}^{(1)}$ and $n^{(2)}$. Let
\begin{equation}
\begin{aligned}
    \mathcal{E}_{\mathbf{d}^{(1)},\xprime,n^{(2)}} &= \left \{\wl{1} \in \mathbb{R}^d: \exists S \subset [N], |S| = d-1, \text{ s.t. } \mathbf{M}_s \wl{1} =\mathbf{0}, \text{ rank}(\mathbf{M}_s)=d-1 \right \}.
  % &= \left\{\frac{\times_{i=1}^{d{-}1} {\mathbf{M}_s}_i}{\left\|\times_{i=1}^{d{-}1} {\mathbf{M}_s}_i\right\|_1}: \Delta {\mathbf{x}}_i \in \mathbf{M}_s,  \text{ rank}(\mathbf{M}_s)=d-1\right\}\\
\end{aligned}
\end{equation}

%does x need to be full rank
Let ${\mathbf{M}_s}_i$ denote the $i^{\text{th}}$ row of the submatrix $\mathbf{M}_s$. By the orthogonality property of generalized cross products (\Cref{section:geoalgebra}),
\begin{equation}\label{eq:Worthog}
\begin{aligned}
    \mathcal{E}_{\mathbf{d}^{(1)},\xprime,n^{(2)}} \cap \delta B
   = \left\{\pm \frac{\times_{i=1}^{d{-}1} {\mathbf{M}_s}_i}{\left\|\times_{i=1}^{d{-}1} {\mathbf{M}_s}_i\right\|_1}:  \text{ rank}(\mathbf{M}_s)=d-1\right\}
   \subset  \Delta \mathcal{X} 
\end{aligned}
\end{equation}
where $\Delta \mathcal{X}$ is defined in \eqref{eq:augdataset}.
By Lemma 19 in \citep{pilanci2023complexity}, 
%for  $\mathbf{a} \in \mathcal{K}_{n^{(1)},n^{(2)}}$ such that $\bl{1}=-\mathbf{a}\wl{1}$ and $\mathbf{d}^{(1)}\in \mathcal{D}^{(1)}\left(\bl{1}\right)$, 
the set of extremal points of $\mathcal{P}$ is
\begin{equation}\label{eq:extremal}
\mathcal{E}_{\mathbf{d}^{(2)},\mathbf{d}^{(1)},\mathbf{a},n^{(2)}} {=}  S_{\xprime}^{-1}\left(\mathbf{d}^{(1)}\right){\cap} 
S_{\mathbf{d}^{(1)},\xprime,n^{(2)}}^{-1}\left(\mathbf{d}^{(2)}\right){\cap  }\mathcal{E}_{\mathbf{d}^{(1)},\xprime,n^{(2)}} {\cap} \delta B.
\end{equation}

Therefore for fixed $n^{(1)},n^{(2)},\mathbf{a}$, plugging \eqref{eq:Worthog} into \eqref{eq:extremal} gives
\begin{equation}\label{eq:DeltamathcalX}
    \displaystyle \bigcup_{\mathbf{d}^{(1)} {\in} S_\mathbf{a}\left(\mathbb{R}^d\right), \mathbf{d}^{(2)}{\in}S_{\mathbf{d}^{(1)},\xprime,n^{(2)}}\left(\mathbb{R}^d\right)}\mathcal{E}_{\mathbf{d}^{(2)},\mathbf{d}^{(1)},\mathbf{a},n^{(2)}} {\subset}  \Delta \mathcal{X}. 
\end{equation}

Plugging in \eqref{eq:DeltamathcalX} as a superset of extremal points of $\mathcal{P}$ into \eqref{eq:L=234lyr_max_constr} shows that the dual constraint \eqref{eq:dual_constraint} is equivalent to  

\begin{equation}\label{eq:dualexpanded}
\begin{aligned}
 \max_{\wl{2} {\in} \{-1,1\}} 
    \max_{n^{(1)},n^{(2)} {\in} [N] } 
    \max_{\xprime {\in} \mathcal{K}_{n^{(1)},n^{(2)}}} 
    %
    %\max_{\mathbf{d}^{(1)} {\in} S_\mathbf{a}\left(\mathbb{R}^d\right)} 
    %
   % \max_{\wl{1} {\in} \mathcal{E}_{\mathbf{d}^{(1)},\mathbf{a},n^{(2)}}{\cap}S^{-1}_{\mathbf{a}}\left(\mathbf{d}^{(1)}\right){\cap} B }
    \max_{\wl{1} {\in} \Delta \mathcal{X}}
    \left | \sum_{n=1}^N \lambda_n \xl{L}_n \right| &\leq \beta,
    \mathbf{1}^T \lambda &= 0.
\end{aligned}
\end{equation}
%
%where 
%$\xl{L}$ can be computed as \eqref{eq:lambda^TX_L-12}, $\bl{1}=-\mathbf{a}\wl{1}$ and
%
%\begin{equation*}
%\scalemath{0.8}{
%%\mathcal{E}_{\mathbf{j},\mathbf{a}} = \left \{\wl{1} \in \mathbb{R}^d: \exists S \subset [N], |S| = d-1, \text{ s.t. } \mathbf{M}_s \wl{1} =\mathbf{0}, \text{ rank}(\mathbf{M}_s)=d-1 \right \}.
%\mathcal{E}_{\mathbf{d}^{(1)},\mathbf{a},n^{(2)}}{\cap}S^{-1}_{\mathbf{a}}\left(\mathbf{d}^{(1)}\right){=}\left \{\wl{1} \in \mathbb{R}^d{:} \text{sign}\left(\left(\mathbf{X}-\mathbf{1}\mathbf{a}\right)\wl{1} \right){=} \mathbf{d}^{(1)}, {\exists} S {\subset} [N], |S| {=} d{-}1, \text{ s.t. } \mathbf{M}_s \wl{1} {=}\mathbf{0}, \text{ rank}(\mathbf{M}_s){=}d{-}1 \right \}.
%}
%\end{equation*}

Using \eqref{eq:Xrecurse}, we can write \eqref{eq:optbias} as $\bl{l} = -\xl{l}_{n^{(l)}}$ for $l>1$, where $\wl{l'}=1$ for $l'>1$.

The dual of \eqref{eq:dualexpanded} is the \lasso{} problem \eqref{eq:lasso} where $\mathbf{A}_{i,j}$ is determined by \eqref{eq:XL_nexpand}, or
\begin{equation}\label{eq:Amati}
    \Amat_i=\xl{3}\left(\mathbf{X}\right).
\end{equation}
In \eqref{eq:Amati}, $\xl{3}\left(\mathbf{X}\right)$ is the output of the network $\xl{3}$ with data matrix $\mathbf{X}$ as input, where $\wl{2}=1$ and $\bl{l}$ determined by \eqref{eq:optbias} (with $\mathbf{x}_{j_1}'=\mathbf{a}$,
i.e. which have $i^{th}$ element

\begin{equation}\label{eq:Aij_xl}
    \begin{aligned}
        \mathbf{A}_{i,j} %&= \xl{L}\left(\mathbf{X}\right)_i\\
        &=\left|\left|\left(\mathbf{x}_i {-}\xprime \right)\wl{1} \right| {-} \left| \left(\mathbf{x}_{n^{(2)}} {-}\xprime \right)\wl{1}\right| \right|
    \end{aligned}
\end{equation}

%for all $\wl{2} {\in} \{-1,1\}, n^{(1)},n^{(2)} {\in} [N], \mathbf{a} {\in} \mathcal{K}_{n^{(1)},n^{(2)}},  \mathbf{d}^{(1)} {\in} S_\mathbf{a}\left(\mathbb{R}^d\right), \wl{1} {\in} \mathcal{E}_{\mathbf{d}^{(1)},\mathbf{a},n^{(2)}}{\cap}S^{-1}_{\mathbf{a}}\left(\mathbf{d}^{(1)}\right){\cap} \delta B $, corresponding to columns of $\mathbf{A}$.

%$ \bl{1}{=}{-}\mathbf{a}\wl{1}$ and $ \bl{2}{=} {-}\xl{2}_{n^{(2)}}\wl{2}$ which is computed using $\bl{1}$. 

The \lasso{} problem is the bidual of the rescaled problem, and therefore a lower bound on it, which is equivalent to the original training problem.
In fact, \Cref{remark:reconstructionsame} shows that 
%given an optimal solution to the \lasso{} problem, observe that following the reconstruction given in \Cref{def:reconstruct} gives a neural network that achieves the same objective in the training problem. %following the reconstruction \Cref{def:reconstruct} setting $\boldsymbol{\alpha} {=} \mathbf{z}^*, \gamma = \gamma^*, \Wul{i}{l}=1$ for $l>1$, and for each dictionary column $\mathbf{A}_i$, consider the corresponding $\mathbf{a}, \wl{1}, n^{(2)}$ and let $\bul{i}{l}=\bl{l}$ as shown in \eqref{eq:optbias}. 
the \lasso{} problem is equivalent to the original training problem, and the reconstruction holds.
%
%Note that $S_\mathbf{a}\left(\mathbb{R}^d\right)$ is the set of hyperplane arrangements, which is finite. %can be sampled in practice
%And $\mathcal{E}_{\mathbf{d}^{(1)},\mathbf{a},n^{(2)}} \cap \delta B\mathcal{E}_{\mathbf{d}^{(1)},\mathbf{a},n^{(2)}} \cap \delta B$ is also finite. %is this obvious? l1 not l2 ball
%
%For $\mathbf{v}_1,{\cdots},\mathbf{v}_d{\in}\mathbb{R}^{d}$, let Vol$(\mathbf{v}_1,{\cdots},\mathbf{v}_d)$ be the unsigned volume spanned by $\mathbf{v}_1,{\cdots},\mathbf{v}_d$.
%
Lastly, rename $j_0 {=} n^{(2)}, j_{-1}{=}n^{(1)}$. The volume formulation \eqref{eq:3layerdicsimpler} holds from \Cref{rk:feature_as_wedge} and \eqref{eq:volume}. The orthogonality property of $\wl{1}$ holds by the orthogonality property of cross products.
%Plugging in \eqref{eq:Ea} into $\mathbf{A}_{i,j}$ \eqref{eq:Aij_xl}, the $i^{\text{th}}$ element of a dictionary column $\mathbf{A}_j$, with corresponding $ \mathbf{a} {\in} \left\{\mathbf{x}_{n^{(1)}}, \frac{\mathbf{x}_{n^{(2)}}{+}\mathbf{x}_{n^{(1)}}}{2}\right\}, \wl{1}$ gives $\mathbf{A}_{i,j}=f(\mathbf{x}_i)$ as given in \eqref{eq:3layerfeature}.

%\begin{equation}\label{eq:Aijexpand}
%    \begin{aligned}
%        \mathbf{A}_{i,j} &= \left|\left|\left(\mathbf{x}_i {-}\mathbf{a} \right)\frac{\times_{i{=}1}^{d{-}1} \Tilde{\mathbf{x}}_i }{\|\times_{i{=}1}^{d{-}1} \Tilde{\mathbf{x}}_i \|_1}  \right| {-} \left| \left(\mathbf{x}_{n^{(2)}} {-}\mathbf{a} \right)\frac{\times_{i{=}1}^{d{-}1} \Tilde{\mathbf{x}}_i }{\left\|\times_{i{=}1}^{d{-}1} \Tilde{\mathbf{x}}_i \right\|_1} \right| \right|\\
%        &= \left|\frac{\text{Vol}\left(\mathbf{x}_i {-}\mathbf{a} , \Tilde{\mathbf{x}}_1, \cdots, \tilde{\mathbf{x}}_{d{-}1} \right){-}\text{Vol}\left(\mathbf{x}_{n^{(2)}} {-}\mathbf{a} , \Tilde{\mathbf{x}}_1, \cdots, \tilde{\mathbf{x}}_{d{-}1} \right)}{\left\|\times_{i{=}1}^{d{-}1} \Tilde{\mathbf{x}}_i \right\|_1}
%         \right|%\\
%       % &= \frac{\text{Vol}\left(\tilde{\mathbf{x}}_d, \Tilde{\mathbf{x}}_1, \cdots, \tilde{\mathbf{x}}_{d{-}1} \right)}{\left\|\times_{i{=}1}^{d{-}1} \Tilde{\mathbf{x}}_i \right\|_1}
%         %\\
%    \end{aligned}
%\end{equation}

%Simplifying \eqref{eq:Aijexpand} gives \eqref{eq:Aij} and \eqref{eq:x_d}.
%
\end{proof}

\begin{proof}[Proof of \Cref{lemma:Asize}]\label{proof_of_lemmaAsize}
    Since $j_{{-}1}, j_0, j_2, j_4, {\cdots}, j_{2(d{-}1)} \in [N]$, there are $N^{d+1}$ choices for determining $\mathbf{x}_{j_{{-}1}}, \mathbf{x}_{j_0}, \mathbf{x}_{j_2}, \mathbf{x}_{j_4}, {\cdots}, \mathbf{x}_{j_{2(d{-}1)}}$. For each $j_{{-}1},j_0$ there are $\left\|\left\{\frac{\mathbf{x}_{j_{-1}}+\mathbf{x}_{j_0}}{2},\mathbf{x}_{j_{-1}}\right\}\right\|=2$ options for $\mathbf{x}'_{j_1}$. For each $j_{{-}1},j_0, \mathbf{x}_{j_{2k}}$, there are $\left|\left\{\mathbf{x}'_{j_1}, \reflect{\mathbf{x}_{j_0}}{\mathbf{x}'_{j_1}}\right\} {\cup} \{\mathbf{x}_{j_{2k}}{-}e_l\}_{l\in[d]}\right|{=}2{+}d$ options for $\mathbf{x}'_{j_{2k+1}}$. So the total number of options for $j$ is at most $2(N^{d+1})(2{+}d)^{d-1}=O((Nd)^d)$.
    The complexity for training a $2$-layer network must be exponential in $d$ unless $\mathbf{P}=\mathbf{NP}$ \cite{pilanci2020neural}.
\end{proof}

%\begin{proof}[Proof of \Cref{thrm:2layersimple}, \Cref{thrm:3layersimple}]
%Since 
%$\Delta \mathcal{X}^{(L)}_\perp \supset \left\{\frac{\times_{i{=}1}^{d{-}1} \Delta \mathbf{x}_i }{\|\times_{i{=}1}^{d{-}1} \Delta \mathbf{x}_i \|_1}: \Delta \mathbf{x}_i {\in} \tilde{\mathcal{X}}^{(L)}\right \}$ and $\mathcal{X}' \supset  \{u(\mathbf{x}_n):n \in [N] \}$, the  maximization constraint \eqref{eq:dualexpanded} remains equivalent in the proof of \Cref{thrm:3layer} if we instead take $\wl{1} \in \Delta \mathcal{X}^{(L)}_\perp$ and $\bl{1} \in \mathcal{X}' $.
%\end{proof}

%\Tilde{\mathbf{x}}_{1}, {\cdots}, \Tilde{\mathbf{x}}_{d-1} \in

\begin{remark}\label{rk:explicit_features}
The $2$-layer feature function is
\begin{equation}\label{eq:2layerfeature}
    \begin{aligned}
         f(\mathbf{x}) &=   \left|\left(\mathbf{x} -\mathbf{x}_{n^{(1)}}\right)\overbrace{\frac{\times_{i=1}^{d-1} \Delta{\mathbf{x}}_i }{\left \| \times_{i=1}^{d-1} \Delta{\mathbf{x}}_i \right \|_1}}^{\wl{1} \perp \Delta{\mathbf{x}}_1,\cdots,\Delta{\mathbf{x}}_{d-1}}\right| .
    \end{aligned}
\end{equation}

Similarly the $3$-layer feature function is 

\begin{equation}\label{eq:3layerfeature}
    \begin{aligned}
         f(\mathbf{x}) &=  \left| \left|\left(\mathbf{x} -\xprime\right)\overbrace{\frac{\times_{i=1}^{d-1} \Delta{\mathbf{x}}_i }{\left \| \times_{i=1}^{d-1} \Delta{\mathbf{x}}_i \right \|_1}}^{\wl{1} \perp \Delta{\mathbf{x}}_1,\cdots,\Delta{\mathbf{x}}_{d-1}}\right|  - \overbrace{\left|\left(\mathbf{x}_{n^{(2)}} -\xprime\right)\frac{\times_{i=1}^{d-1} \Delta{\mathbf{x}}_i }{\left \| \times_{i=1}^{d-1} \Delta{\mathbf{x}}_i \right \|_1}\right|}^{-\sbl{2}} \right |.
    \end{aligned}
\end{equation}

   % The reconstruction formula (\Cref{def:reconstruct}) simply makes the network a linear combination of the features \eqref{eq:feature} corresponding to each column of the dictionary matrix, scaled by the solution $\mathbf{z}^*_i$.

\end{remark}

\begin{definition}\label{def:reconstruct}
    Given an optimal solution to the \lasso{} problem whose features are \eqref{eq:feature}, the $L$-layer \textit{reconstruction} is as follows.
    Set $\boldsymbol{\alpha} {=} \mathbf{z}^*, \gamma{=} \gamma^*, \Wul{i}{l}{=}1$ for $l{>}1$, and for each dictionary column $\mathbf{A}_i$, consider the corresponding $ \wl{1}, \bl{l}$ and let $\Wul{i}{1}{=}\wl{1} $ and $ \bul{i}{l}{=}\bl{l}$. For data feature biases, use \eqref{eq:optbias} for $\bl{l}$. For $L{=}3$, $\wl{1}{\in} \Delta \mathcal{X}^{(3)}$ \eqref{eq:augdataset}. This gives a \emph{rescaled network}.
    Then, let $\gamma_i {=}  \left|\alpha_i\right|^{\frac{1}{L}}$.
%\begin{definition}\label[definition]{def:unscaling}
    Finally change variables as $q' {=} \mathrm{sign}(q)\gamma_i$ for $q \in \left \{\alpha_i, \Wul{i}{l} \right\}$ and ${\bul{i}{l}}'{=} \bul{i}{l}  \left(\gamma_i\right)^l $ for the reconstructed network. 
    For example, a reconstructed $3$-layer neural network from a \lasso{} solution $\mathbf{z}^*, \xi^*$ is %equivalent to 
%todo explain that even when a is the midpoint, just get usual not special reflection

\begin{equation}
\scalemath{0.85}{
    \begin{aligned}
         \nn{3}{\mathbf{x}} %&{=} \sum_j z^*_j f^{(j)}(\mathbf{x}) + \xi^*\\
         &{=} \sum_{j} (z^*_j)^{\frac{1}{3}} \left|(z^*_j)^{\frac{1}{3}}  \left|(z^*_j)^{\frac{1}{3}} \left(\mathbf{x} {-}{\xprime}^{(j)}\right)\frac{\times_{i=1}^{d{-}1} \Delta{\mathbf{x}}^{(j)}_i }{\left \| \times_{i=1}^{d{-}1} \Delta{\mathbf{x}}^{(j)}_i \right \|_1}\right|  {-} (z^*_j)^{\frac{1}{3}} \left|(z^*_j)^{\frac{1}{3}} \left(\mathbf{x}^{(j)}_{n^{(2)}} {-}{\xprime}^{(j)}\right)\frac{\times_{i=1}^{d{-}1} \Delta{\mathbf{x}}^{(j)}_i }{\left \| \times_{i=1}^{d-1} \Delta{\mathbf{x}}^{(j)}_i \right \|_1}\right| \right| {+} \xi^*.
    \end{aligned}
    }
\end{equation}

where the index $(j)$ indexes over all possible $\wl{1}, \xprime, \mathbf{x}_{n^{(2)}}$ corresponding to the $j^{\mathrm{th}}$ feature vector.
\end{definition}

\begin{remark}\label{remark:reconstructionsame}
    %Given an optimal solution to the \lasso{} problem, the reconstruction given in \Cref{def:reconstruct} gives a neural network that achieves the same objective in the training problem. 
   % The function $\sum_i z^*_i \featurefunc_j(\mathbf{x})+\xi^*$ is the network where $\Wul{i}{l}=\wl{}$  
   The rescaled network in \Cref{def:reconstruct} achieves the same value in the rescaled problem as the \lasso{} optimal value. \Cref{def:reconstruct} then "un-scales" the weights in the rescaled network to give a reconstructed network that achieves the same objective in the original training problem as the optimal \lasso{} value. The rescaled network, reconstructed network and the function $\sum_i z^*_i {\featurefunc}_j(\mathbf{x})+\xi^*$ are all equivalent as functions, but have different interpretations of neuron weights.
\end{remark}

\begin{proof}[Proof of \Cref{theorem:geometricInterp}]\label{proof:thrm_geom_interp}
    %The first equality follows from the definition of the feature function. The second equality follows from $\|\mathbf{x}\times_{i=1}^{d-1} \Tilde{\mathbf{x}}_i \|=\text{Vol}(\mathbf{x},\Tilde{\mathbf{x}}_1,\cdots,\tilde{\mathbf{x}}_{d-1}) $. For the third equality, 
The wedge product formulation follows from \Cref{theorem:geometricInterp}. 
To show the distance formulation, observe that the features are given by \eqref{eq:Aij_xl} as 
\begin{equation}
    \begin{aligned}
        \featurefunc(\mathbf{x}) &=  \left| \left|\left(\mathbf{x} -\xprime\right)\wl{1}\right|  - \left|\left(\mathbf{x}_{n^{(2)}} -\xprime\right)\wl{1}\right| \right |
    \end{aligned}
\end{equation}
with $\wl{1}=\frac{\mathbf{w}}{\|\mathbf{w}\|_1} \in \Delta \mathcal{X}^{(3)}$ \eqref{eq:augdataset}, where
$\mathbf{w}{=}\times_{i=1}^{d{-}1} \Delta{\mathbf{x}}^{(j)}_i$. Let $l_1$ and $l_2$ be the hyperplanes defined by $\left(\mathbf{x}{-}\reflect{\mathbf{\mathbf{x}_{n^{(2)}}}}{\xprime}\right)\mathbf{w}{=}0$ and $\left(\mathbf{x}{-}\reflect{\mathbf{\mathbf{x}_{n^{(2)}}}}{\xprime}\right)\mathbf{w}{=}0$, respectively. Let $l=l_1 \cup l_2$. By \Cref{rk:distance_volume},
\begin{equation}
    \begin{aligned}
         \featurefunc(\mathbf{x}) %&=  \left| \left|\left(\mathbf{x} -\xprime\right)\wl{1}\right|  - \left|\left(\mathbf{x}_{n^{(2)}} -\xprime\right)\wl{1}\right| \right | \\
            &= \min \left\{\left|\left(\mathbf{x}-\mathbf{x}_{n^{(2)}}\right)\wl{1}\right|, \left|\left(\mathbf{x}-\reflect{\mathbf{\mathbf{x}_{n^{(2)}}}}{\xprime}\right)\wl{1}\right| \right\}\\
            &= \lratiopenalty(\mathbf{w}) \min \left\{\left|(\mathbf{x}-\mathbf{x}_{n^{(2)}})\frac{\mathbf{w} }{\left \|\mathbf{w}\right \|_2}\right|, \left|\left(\mathbf{x}-\reflect{\mathbf{\mathbf{x}_{n^{(2)}}}}{\xprime}\right)\frac{\mathbf{w} }{\left \|\mathbf{w}\right \|_2} \right| \right\}\\
           % &= \lratiopenalty(\mathbf{w})  \min \left\{\mathrm{Dist}\left(\mathbf{x},\mathcal{H}_{(j_1,j_{2:2d-1})}\cup\reflect{\mathcal{H}_{(j_1,j_{2:2d-1})}}{\mathcal{H}_{(j_0,j_{2:2d-1})}}\right) \right\}
            &= \lratiopenalty(\mathbf{w}) \min \{ d(\mathbf{x},l_1), d(\mathbf{x},l_2)\}\\
            &= \lratiopenalty(\mathbf{w})
            d(\mathbf{x},l).
    \end{aligned}
\end{equation} 

Now rename $j_{0}=n^{(2)}, j_{-1}=j_1$. Observe that if $\xprime{=}\frac{\mathbf{x}_{j_{-1}}+\mathbf{x}_{j_{0}}}{2}$ then $\reflect{\mathbf{x}_{j_0}}{\xprime}=\mathbf{x}_{\mathbf{x}_{j_{-1}}}$. A case analysis on $\xprime \in \left\{\frac{\mathbf{x}_{j_{-1}}+\mathbf{x}_{j_{0}}}{2},\mathbf{x}_{j_{-1}}\right\}$ gives \eqref{eq:f_as_distance_wedge}.
\end{proof}

%\begin{proof}[Proof of \Cref{lemma:orthog}]
%Follows from the reconstruction formulation (\Cref{def:reconstruct}, %\Cref{lemma:reconsruction_optimal} and \eqref{eq:Ea}).    
%\end{proof}

%might not need these defs at all
%Note we can define bias parameter $\bl{l}$ as a \emph{data feature bias} if $\bl{l}{=}{-}\xl{l}(\mathbf{x}_0 )\wl{l}$ for some column vector $\mathbf{x}_0 {\in} \{x_1,{\cdots},x_N\}^{m_l}$.
%\end{definition}
%%Using the data feature bias, we define a deep library.
%%\begin{definition}[Deep Library]\label{def:FeatDicStdNN}
% Then, the deep library is a set of all network outputs $\xl{L}(\mathbf{X})$ %defined in \eqref{eq:parallelnn_rec} 
% with data or midpoint feature biases and all elements of $\wl{1}$ being $\pm 1$ for $l>1$.
% %and $\wl{1} \in  \mathcal{E}_{\mathbf{d}^{(1)},\mathbf{a},n^{(2)}}$ \eqref{eq:Ea}.

%\begin{proof}[Proof of \Cref{lemma:reconsruction_optimal}]
%    Follows from \Cref{remark:reconstructionsame} and \Cref{thrm:3layer}.
%\end{proof}

\begin{proof}[Proof of \Cref{lemma:deep}]\label{proof_lemmadeep}
    Recall the proof of \Cref{thrm:3layersimpler} until \eqref{eq:optbL-2}. Continuing this argument of finding a finite set of possible \kink{}s of $\lambda^T \xl{L}$ as a function of $\bl{L{-}l}$, plugging them in to $\xl{L}$ and further expanding $\lambda^T \xl{L}$ as a function of $\bl{L{-}l{-}1}$ until $l {=}L{-}2$ shows that for each fixed $\wl{1}$, there is a finite set of optimal $\bl{1}$, which determines a finite set of optimal $\bl{2}$, and so on determining a finite set of all other possible $\bl{l}$. We also see that for all $l{>}1$, $\wl{l}{=}1$ and $\wl{l}{=}{-}1$ are both optimal. Furthermore, the optimal values of $\bl{l}$ include 
    \begin{equation}\label{eq:general_biasfeature}
        \bl{l}{=}{-}\xl{l}_{n^{(L{-}l)}}\wl{l}
    \end{equation}
    for $n^{(L{-}l)} {\in} [N]$. We can express $\xl{L}_n$ as a function of $\wl{1}$, generalizing \eqref{eq:XL_nexpand}. 
    Next, partition $\mathbb{R}^d$ into a finite set of polytopes $\mathcal{P}_1,\cdots,\mathcal{P}_p$ corresponding to all possible sign patterns of activation arguments (generalizing \eqref{eq:polytope}). The objective linearizes for $\wl{1}$ within each polytope, so mazimizing $\left|\lambda^T \xl{L} \right|$ is equivalent to maximizing over the finite set of extreme points of the polytope, generalizing \eqref{eq:Worthog}. So the dual constraint \eqref{eq:dual_constraint} consists of a finite set of linear constraints. As in the proof of \Cref{thrm:3layersimpler}, the bidual of the training problem is a \lasso{} problem whose dictionary columns are $\xl{L}(\mathbf{X})$ (generalizing \eqref{eq:Amati}) over all possible optimal $\wl{1}, \bl{l}$.
    A similar reconstruction as \Cref{def:reconstruct} of an optimal neural network from a \lasso{} solution holds and shows that the \lasso{} problem and the training problem are equivalent. And \eqref{eq:general_biasfeature} shows that the dictionary contains a \sublibrary{}. Note that the $4$-layer features plotted in the middle and right plots of \Cref{fig:absvalfeat} depict $\wl{1}$ as orthogonal to the difference of training samples. By a similar argument as the proof of \Cref{thrm:3layersimpler}, this property holds for deeper layers as well.
\end{proof}

%\begin{proof}[Proof of \Cref{absvalsublib}]
 %   Follows from proofs of \Cref{thrm:3layer} and \Cref{lemma:deep} \eqref{eq:general_biasfeature} and the reconstruction formulas \Cref{def:reconstruct}, \Cref{lemma:reconsruction_optimal}. 
%\end{proof}

\section{Deep features with higher order reflections}\label{app:deep}

We define $\mathbf{x}$ to be an order-$l$ reflection of a \emph{set} $\mathcal{S}^{(l)}{=}\left\{\mathbf{x}_0, {\cdots}, \mathbf{x}_k\right\}$ if $\mathbf{x}$ can be written recursively in the form $R_l{=}R_l(\mathcal{S}^{(l)}){\in} \left\{
\reflect{R_{l{-}1}\left(\mathcal{S}^{(l{-}1)}\right)}{\mathbf{x}'},
\reflect{\mathbf{x}'}{R_{l{-}1}\left(\mathcal{S}^{(l)}\right)} \right\}$
where $\mathcal{S}^{(l{-}1)} {\subset} \mathcal{S}^{(l)}$ with possibly repeating elements and $\left|\mathcal{S}^{(l)}\right|{=}l{+}1$.  
%the arguments $\mathbf{x}_0, \cdots, \mathbf{x}_k$ of the reflection may be implicit and suppressed in the notation: $R_k {=} R_k(\mathbf{x}_0, \cdots, \mathbf{x}_k)$. 
Note that an order-$k$ reflection is also an order-$j$ reflection for any $j\geq k$. 
%We analyze the \sublibrary.
In the following lemma, let $a^{(L)}, b^{(L)} {\in} \mathbb{R}, \mathbf{c}^{(L)} {\in} \mathbb{R}^N, j_1, \cdots, j_{L} {\in} [N]$. For $l {\in} [L{-}1]$, recursively define
    $a^{(l)}{=}\left|a^{(l{+}1)}{-}c^{(l{+}1)}_{j_{l{+}1}} \right|,b^{(l)}{=}\left|b^{(l{+}1)}{-}c^{(l{+}1)}_{j_{l{+}1}} \right|, c_{j_l}^{(l)}{=}\left|c_{j_l}^{(l{+}1)}-c_{j_{l{+}1}}^{(l{+}1)}\right|$. Let $\mathcal{S}^{(l)}{=}\left\{a^{(l)},b^{(l)}\right\}\cup\left\{c^{(l)}_{l'}:l'\leq l\right\}$.  We will see that this recursively models the structure of a feature function in a \sublibrary{}.

\begin{lemma}\label{lemma:abrecursion}
    Let $l {\in} [L]$. Let $R^{(l,{+})}_0{=}R^{(l,{-})}_0 {=} \reflect{a^{(l)}}{b^{(l)}}$. For $l' {\in} \{0,{\cdots},L{-}l{-}1\}$, there exist $l{+}l'{+}1$-order reflections $R_{2(l'{+}1)}^{(l{+}l'{+}1,{+})},R_{2(l'{+}1)}^{(l{+}l'{+}1,{-})}$ of $\mathcal{S}^{(l{+}l'{+}1)}$ such that 
    %and $R^{(n)}_k$ denote a order-$k$ reflection of $a^{(l)},b^{(l)},c^{(l)}$. 
    \begin{equation}\label{eq:general_recursebias}
        \begin{aligned}
            R_{2l'}^{(l{+}l',{\pm})} &\in \left\{ R_{2(l'{+}1)}^{(l{+}l'{+}1,{+})}-c^{(l{+}l'{+}1)}_{l{+}l'{+}1},c^{(l{+}l'{+}1)}_{l{+}l'{+}1} - R_{2(l'{+}1)}^{(l{+}l'{+}1,{-})}\right\}.
        \end{aligned}
    \end{equation}
   % for some $l{+}l'{+}1$-order reflections $R_{(2(l'+1)}^{(l+l'+1,+)},R_{(2(l'+1)}^{(l+l'+1,-)}$ 
   for $ R_{2l'}^{(l{+}l',{\pm})} \in \left\{ R_{2l'}^{(l{+}l',{+})}, R_{2l'}^{(l{+}l',{-})}\right\}.$
\end{lemma}

\begin{proof}[Proof of \Cref{lemma:abrecursion}]
    Observe that for any $l$,

\begin{equation}
\scalemath{1}{
    \begin{aligned}
        \reflect{a^{(l)}}{b^{(l)}}&{=}\reflect{\left|a^{(l+1)}{-}c^{(l+1)}_{j_{l{+}1}}\right|}{\left|b^{(l+1)}{-}c^{(l+1)}_{j_{l{+}1}}\right|}\\
        &{\in}{\pm}\left\{\reflect{a^{(l{+}1)}{-}c^{(l{+}1)}_{j_{l{+}1}}}{b^{(l{+}1)}{-}c^{(l{+}1)}_{j_{l{+}1}}}, \reflect{a^{(l{+}1)}{-}c^{(l{+}1)}_{j_{l{+}1}}}{c^{(l{+}1)}_{j_{l{+}1}}{-}b^{(l{+}1)}} \right\}\\
        &{=}{\pm}\left\{{-}a^{(l{+}1)} {+}2b^{(l{+}1)} {-}c^{(l{+}1)}_{j_{l{+}1}}  , {-}a^{(l{+}1)}{-}2b^{(l{+}1)}{+}3c^{(l{+}1)} \right\}\\
        &{=}\pm \left\{\reflect{a^{(l{+}1)}}{b^{(l{+}1})}{-}c^{(l{+}1)}_{j_{l{+}1}}{=}c^{(l{+}1)}_{j_{l{+}1}} {-}\reflect{\reflect{a^{(l{+}1)}}{b^{(l{+}1)}}}{c^{(l{+}1)}_{j_{l{+}1}}}, \right. \\ 
        & \quad \left. \reflect{a^{(l{+}1)}}{\reflect{b^{(l{+}1)}}{c^{(l{+}1)}_{j_{l{+}1}}}}{-}c^{(l{+}1)}_{j_{l{+}1}}{=}c^{(l{+}1)}_{j_{l{+}1}}{-}\reflect{\reflect{a^{(l{+}1)}}{c^{(l{+}1)}_{j_{l{+}1}}}}{b^{(l{+}1)}} \right \}.
        %&
        %\Bigg\{\reflect{a^{(l{+}1)}{-}c^{(l{+}1)}}{b^{(l{+}1)}{-}c^{(l{+}1)}} {=}  {-}a^{(l{+}1)} {+}2b^{(l{+}1)} {-}c^{(l{+}1)}  {=} \reflect{a^{(l{+}1)}}{b^{(l{+}1})}{-}c^{(l{+}1)} {=} c^{(l{+}1)} {-}\reflect{\reflect{a^{(l{+}1)}}{b^{(l{+}1)}}}{c^{(l{+}1)}}, \\
        %&\reflect{a^{(l{+}1)}{-}c^{(l{+}1)}}{c^{(l{+}1)}{-}b^{(l{+}1)}} {=}{-}a^{(l{+}1)}{-}2b^{(l{+}1)}{+}3c^{(l{+}1)}{=} \reflect{a^{(l{+}1)}}{\reflect{b^{(l{+}1)}}{c^{(l{+}1)}}}{-}c^{(l{+}1)} {=} c^{(l{+}1)}{-}\reflect{\reflect{a^{(l{+}1)}}{c^{(l{+}1)}}}{b^{(l{+}1)}}\Bigg\}\\
    \end{aligned}
    }
\end{equation}
Therefore for some $a^{(l+1)'},b^{(l+1)'},c^{(l+1)'}\in\left\{a^{(l+1)},b^{(l+1)},c^{(l+1)}_{j_{l{+}1}}\right\}$ and
\begin{equation}\label{eq:dblereflection}
    R^{(l{+}1,{+})}_2, R^{(l{+}1,{-})}_2 \in \left\{\reflect{\reflect{a^{(l+1)'}}{b^{(l+1)'}}}{c^{(l+1)'}},\reflect{c^{(l+1)'}}{\reflect{a^{(l+1)'}}{b^{(l+1)'}}}  \right\}
\end{equation}
which are
order-$2$ reflections of $\left\{a^{(l+1)},b^{(l+1)},c^{(l+1)}_{j_{l{+}1}}\right\}$, we have
\begin{equation}\label{eq:recurse}
    \reflect{a^{(l)}}{b^{(l)}}{=}\left\{c^{(l{+}1)}_{j_{l{+}1}}-{R}^{(l+1,-)}_2={R}^{(l+1,+)}_2-c^{(l{+}1)}_{j_{l{+}1}}\right\}.
\end{equation}

Applying \eqref{eq:recurse} for $l'=l+1$ gives 
\begin{equation}\label{eq:recursel+1}
    \reflect{a^{(l{+}1)}}{b^{(l{+}1)}}\in\left\{c^{(l{+}2)}_{j_{l{+}2}}-{R}^{(l+2,-)}_2 = {R}^{(l+2,+)}_2-c^{(l{+}2)}_{j_{l{+}2}}\right\}
\end{equation}
 for some $R^{(l{+}2,+)}_2,R^{(l{+}2,-)}_2$ that are order-$2$ reflections of $\left\{a^{(l{+}2)},b^{(l{+}2)},c^{(l{+}2)}_{j_{l{+}2}}\right\}$. Plugging \eqref{eq:recursel+1} into \eqref{eq:dblereflection} shows that for $R_2^{(l{+}1,{\pm})}\in\left\{R_2^{(l{+}1,{+})},R_2^{(l{+}1,-)} \right\}$ we have

\begin{equation}\label{eq:R2l+1pos}
\scalemath{1}{
\begin{aligned}
R_2^{(l+1,\pm)}&=\reflect{\reflect{a^{(l+1)'}}{b^{(l+1)'}}}{c^{(l+1)'}}\\
    &=2\left|c^{(l+2)'}-c^{(l+2)}_{j_{l{+}2}}\right|-\left\{c^{(l{+}2)}_{j_{l{+}2}}-{R}^{(l+2,-)}_2 = {R}^{(l+2,+)}_2-c^{(l{+}2)}_{j_{l{+}2}}\right\}\\
    &\in \left\{-c^{(l+2)}_{j_{l{+}2}}+2c^{(l+2)'}-R^{(l+2,+)}_2, c^{(l+2)}_{j_{l{+}2}}-2c^{(l+2)'}+R^{(l+2,-)}_2 \right\}\\
    &=\left\{\reflect{R^{(l+2,+)}_2}{c^{(l+2)'}}-c^{(l+2)}_{j_{l{+}2}}{=}c^{(l+2)}_{j_{l{+}2}}-
    \reflect{\reflect{R^{(l+2,+)}_2}{c^{(l+2)'}}}{c^{(l+2)}_{j_{l{+}2}}}, \right. \\
    &=\left. c^{(l+2)}_{j_{l{+}2}}-\reflect{R^{(l+2,-)}_2}{c^{(l+2)'}}{=} \reflect{
              \reflect{R^{(l+2,-)}_2}{c^{(l+2)'}}
              }{c^{(l+2)}_{j_{l{+}2}}}-c^{(l+2)}_{j_{l{+}2}} \right\}\\
   % &=\Bigg\{
    %\reflect{R^{(l+2,+)}_2}{c^{(l+2)'}}-c^{(l+2)}= c^{(l+2)}-
   % \reflect{\reflect{R^{(l+2,+)}_2}{c^{(l+2)'}}}{c^{(l+2)}}, 
   % c^{(l+2)}-\reflect{R^{(l+2,-)}_2}{c^{(l+2)'}}= 
    %          \reflect{
    %          \reflect{R^{(l+2,-)}_2}{c^{(l+2)'}}
    %          }{c^{(l+2)}}-c^{(l+2)} 
    %\Bigg\}%\\
    %&=\left\{R^{(l+2,+)}_4-c^{(l+2)},c^{(l+2)}- R^{(l+2,-)}_4\right\}
\end{aligned}
}
\end{equation}
 or alternatively,

\begin{equation}\label{eq:R2l+1neg}
    \begin{aligned}
R_2^{(l+1,\pm)}&=\reflect{c^{(l+1)'}}{\reflect{a^{(l+1)'}}{b^{(l+1)'}}}\\  
&=2\left\{c^{(l{+}2)}_{j_{l{+}2}}-{R}^{(l+2,-)}_2 = {R}^{(l+2,+)}_2-c^{(l{+}2)}_{j_{l{+}2}}\right\}-\left|c^{(l+2)'}-c^{(l+2)}_{j_{l{+}2}}\right|\\
  &\in \left\{-c^{(l+2)}_{j_{l{+}2}}+2R^{(l+2,+)}_2-c^{(l+2)'}, c^{(l+2)}_{j_{l{+}2}}-2R^{(l+2,-)}_2+c^{(l+2)'} \right\}\\
  &=\left\{\reflect{c^{(l{+}2)'}}{R^{(l{+}2,{+})}_2}-c^{(l{+}2)}_{j_{l{+}2}}{=}c^{(l{+}2)}_{j_{l{+}2}}- \reflect{\reflect{c^{(l{+}2)'}}{R^{(l{+}2,{+})}_2}}{c^{(l{+}2)}_{j_{l{+}2}}}, \right.  \\
  &\quad \left. c^{(l{+}2)}_{j_{l{+}2}}-\reflect{c^{(l{+}2)'}}{R^{(l{+}2,{-})}_2} {=}
  \reflect{              \reflect{c^{(l{+}2)'}}{R^{(l{+}2,{-})}_2}}{c^{(l{+}2)}_{j_{l{+}2}}}-c^{(l{+}2)}_{j_{l{+}2}}  \right\}
    \end{aligned}
\end{equation}
for some $c^{(l+2)'}{\in}\left\{a^{(l+2)},b^{(l+2)},c^{(l+2)}_{j_{l{+}1}}\right\}$. Therefore \eqref{eq:R2l+1pos} and \eqref{eq:R2l+1neg} show that in all cases, 

\begin{equation}\label{eq:R2l+1} 
    \begin{aligned}
        R_2^{(l+1,\pm)} \in \left\{R^{(l+2,+)}_4-c^{(l+2)}_{j_{l{+}2}},c^{(l+2)}_{j_{l{+}2}}- R^{(l+2,-)}_4\right\}
    \end{aligned}
\end{equation}
for some order-$4$ reflections $R^{(l+2,+)}_4,R^{(l+2,-)}_4$ of $\left\{ a^{(l+2)}, b^{(l+2)}, c^{(l+2)}_{j_{l{+}1}}, c^{(l+2)}_{j_{l{+}2}}\right\}$.

Applying a similar argument used to reach \eqref{eq:R2l+1} for $l'=l+2$ shows that 

\begin{equation}\label{eq:R2l+2} 
    \begin{aligned}
        R_2^{(l+2,\pm)} \in \left\{R^{(l+3,+)}_4-c^{(l+3)}_{j_{l{+}3}},c^{(l+3)}_{j_{l{+}3}}- R^{(l+3,-)}_4\right\}
    \end{aligned}
\end{equation}
for some order-$4$ reflections $R^{(l+3,+)}_4,R^{(l+3,-)}_4$ of $ a^{(l+3)}, b^{(l+3)}, c^{(l+3)}_{j_{l+2}}, c^{(l+3)}_{j_{l+3}}$. 
%Now repeat
Now plug \eqref{eq:R2l+2} into 
%$R^{(l+3,\pm)}_4$ as expressed in 
the last lines of \eqref{eq:R2l+1pos} and \eqref{eq:R2l+1neg}. Applying the same argument as \eqref{eq:R2l+1pos} and \eqref{eq:R2l+1neg} but changing the reflection order subscripts from $1$ to $2$ and from $2$ to $4$, and adding $1$ to the layer superscripts $l+1, l+2$ shows that

\begin{equation}\label{eq:R4l+3} 
    \begin{aligned}
        R_4^{(l+2,\pm)} &\in  \left\{R^{(l+3,+)}_6-c^{(l+3)}_{j_{l+3}},c^{(l+3)}_{j_{l+3}}- R^{(l+3,-)}_6\right\}
    \end{aligned}
\end{equation}

for some order-$6$ reflections $R^{(l+3,+)}_6,R^{(l+3,-)}_6$ of $ \left\{a^{(l+3)}, b^{(l+3)},c^{(l+3)}_{j_{l{+}1}}, c^{(l+3)}_{j_{l{+}2}}, c^{(l+3)}_{j_{l{+}3}} \right\}  $. %or more?.

%Repeating the same argument shows that

Comparing \eqref{eq:recurse}, \eqref{eq:R2l+1}, \eqref{eq:R4l+3} and repeating the same argument shows that

\begin{equation}\label{eq:recursegen}
    \begin{aligned}
        R_1^{l}&=\reflect{a^{(l)}}{b^{(l)}} \\ &\in \left\{R_2^{(l+1,+)}-c^{(l+1)}=c^{(l+1)}-R_2^{(l+1,-)}\right\}\\
        R_2^{(l+1,\pm)} &\in \left\{R^{(l+2,+)}_4-c^{(l+2)},c^{(l+2)}- R^{(l+2,-)}_4\right\}\\
        R_4^{(l+2,\pm)} &\in  \left\{R^{(l+3,+)}_6-c^{(l+3)},c^{(l+3)}- R^{(l+3,-)}_6\right\}\\
        &\vdots\\
        R_{2l'}^{(l+l',\pm)} &\in \left\{ R_{(2(l'+1)}^{(l+l'+1,+)}-c^{(l+l'+1)},c^{(l+l'+1)} - R_{(2(l'+1)}^{(l+l'+1,-)}\right\}.
    \end{aligned}
\end{equation}
This completes the proof.
\end{proof}

%Comparing the definition of $a^{(l)}, b^{(l)}$ in \Cref{lemma:abrecursion} with the first two lines of \eqref{eq:finalrecursion}, the notation in \Cref{lemma:abrecursion} represents the following: 

%$a^{(l)}{=}\xl{L{-}l}_{j_{a}}, a^{(l+1)}{=}\xl{L{-}l{-}1}_{j_{a}}, b^{(l)}{=}\xl{L{-}l}_{j_{b}}, b^{(l+1)}{=}\xl{L-l-1}_{j_{b}}$ and $c_{j_l}^{(l)}{=}\xl{L{-}l}_{j_{L{-}l}}, c_{j_l}^{(l{+}1)}{=}\xl{L{-}l{-}1}_{j_{L{-}l{-}1}}$ for indices $j_a, j_b, j_{L{-}l{-}1} {\in} [N]$. 
%Note $\xl{l}_n {=} \mathbf{X}^{(l)}(\mathbf{x}_n).$ In particular if $l{=}L{-}1$ then $a^{(l)}$ and $ b^{(l)}$ represent training data points, while if $l{=}0$ then $a^{(l)}{=}\xl{L}_{j_a}$ is the output of the neural net for training sample $\mathbf{x}_{j_a}$ and analogously for $b^{(l)}$.

\textbf{Proof sketch of \Cref{absvalsublib}:} We can easily show that $2$ and $3$-layer networks can be written using reflections. We extend this to deeper layers by first showing that reflections of $\xl{l}$ can be expressed as higher order reflections of $\xl{k}$ for $k{<}l$ (\Cref{lemma:abrecursion}). Then, we recursively plug the higher-order reflections into the deeper networks \eqref{eq:finalrecursion}. This will show that for $l{>}2$, the neural net is $\xl{L}{=}\left|\xl{L{-}l}{-}R_{2(l{-}3)}^{(L{-}l)}\right|$ for some order-$2(l{-}3)$ reflection of the parallel units (features) at layer $L{-}l$. Finally, plug in $l{=}L{-}1$.

\begin{proof}[Proof of \Cref{absvalsublib}]\label{proof:absvalsublib}
The features $\featurefunc(\mathbf{x})$ \eqref{eq:feature} of the \sublibrary{} have bias parameters \eqref{eq:optbias}. In other words, $\featurefunc(\mathbf{x}){=}\xl{L}(\mathbf{x})$ where for all $l {\in} [L{-}1]$ there exists $j_l {\in} [N]$ such that for all $\mathbf{x}{\in}\mathbb{R}^d$,
\begin{equation}\label{eq:featurerecurse}
    \begin{aligned}
        \xl{l{+}1}(\mathbf{x}) &= \left|\xl{l}(\mathbf{x}){-}\xl{l}(\mathbf{x}_{j_{l}}) \right|.
    \end{aligned}
\end{equation}
For shorthand, denote $\xl{l}{=}\xl{l}(\mathbf{x}), \xl{l}_j{=}\xl{l}\left(\mathbf{x}_{j_l}\right)$. Then
\begin{equation}
    \begin{aligned}
        \xl{L} &= \left|\xl{L-1}-\xl{L-1}_{j_{L-1}}\right|\\
        &= \left| \left|\xl{L-2}-\xl{L-2}_{j_{L-2}}\right|-\left|\xl{L-2}_{j_{L-1}}-\xl{L-2}_{j_{L-2}}\right|\right|\\
        &\in \left\{\left|\xl{L-2}-\xl{L-2}_{j_{L-2}}\right|, \left| \xl{L-2}-\reflect{\xl{L-2}_{j_{L-1}}}{\xl{L-2}_{j_{L-2}}}\right|\right\}.
    \end{aligned}
\end{equation}
Therefore 
\begin{equation}\label{eq:LL-2R}
    \xl{L} = \left|\xl{L{-}2}-R^{(L{-}2)}_1\right|
\end{equation}
for some order-$1$ reflection $R^{(L{-}2)}_1{\in}\left\{\reflect{\xl{L{-}2}_{j_{L{-}2}}}{\xl{L{-}2}_{j_{L{-}2}}},\reflect{\xl{L{-}2}_{j_{L{-}1}}}{\xl{L{-}2}_{j_{L{-}2}}}\right\}$.
%of $\xl{L{-}2}_{j_{L{-}1}},\xl{L{-}2}_{j_{L{-}2}}$.
%Now, we express the neural net using reflections. 
%Let $R_l = R^{(1)}_l$, which we can see is an order-$l$ reflection of data points. Fix $x \in \mathbb{R}$. Applying \eqref{eq:feature} and \eqref{eq:optbias} for $l \in [2]$ and then henceforth 
Expand $\xl{L{-}2}$ in \eqref{eq:LL-2R} with \eqref{eq:featurerecurse} and apply \Cref{lemma:abrecursion} to $R_1^{(L{-}2)}$. Specifically, we set $a^{(l)},b^{(l)}{\in}\left\{\xl{L{-}2}_{j_{L{-}1}},\xl{L{-}2}_{j_{L{-}2}}\right\}$, let $j_a, j_b$ be the subscripts ($j_{L{-}1}$ or $j_{L{-}2}$) of $a^{(l)},b^{(l)}$, and let $a^{(l{+}1)}{=}\xl{L{-}3}_{j_a},b^{(l{+}1)}{=}\xl{L{-}3}_{j_b},c^{(l{+}1)}{=}\xl{L{-}3}_{j_{L{-}3}}$ in \eqref{eq:recurse}. This yields
\begin{equation}\label{eq:finalrecursion}
    \begin{aligned}
        \xl{L} 
        %&= \left|\xl{L-1}-\xl{L-1}_{j_{L-1}}\right|\\
        %&= \left| \left|\xl{L-2}-\xl{L-2}_{j_{L-2}}\right|-\left|\xl{L-2}_{j_{L-1}}-\xl{L-2}_{j_{L-2}}\right|\right|\\
        %&\in \left\{\left|\xl{L-2}-\xl{L-2}_{j_{L-2}}\right|, \left| \xl{L-2}-\reflect{\xl{L-2}_{j_{L-1}}}{\xl{L-2}_{j_{L-2}}}\right|\right\}\\
       % &= \left|\xl{L-2}-R^{(L-2)}_1\right|\\
        &=\left|\left|\xl{L-3}-\xl{L-3}_{j_{L-3}}\right|-\left\{R_2^{(L-3,+)}-\xl{L-3}_{j_{L-3}}=\xl{L-3}_{j_{L-3}}-R_2^{(L-3,-)}\right\}\right|\\
        &=\left|\xl{L-3}-R_2^{(L-3,\pm)}\right|.
       % &\vdots\\
      %  &=\left|\mathbf{X}-R_{2(L-3)}\right|
    \end{aligned}
\end{equation}

Next, repeat the same argument  \eqref{eq:finalrecursion} but use \eqref{eq:R2l+2} instead of \eqref{eq:recurse} to get $\xl{L}{=}\left|\xl{L{-}4}-R_2^{(L{-}4,\pm)}\right|$. Continue repeating the same argument by using \eqref{eq:general_recursebias} for general $l$. %with $l{\to} L{-}l$. and 
Specifically, for fixed $j_a, j_b \in [N]$, for all $l \in [L]$, let $a^{(l)}{=}\xl{L{-}l}_{j_{a}}, b^{(l)}{=}\xl{L{-}l}_{j_{a}}$ and $c^{(l)}_{j_l} {=}\xl{L{-}l}_{j_{L{-}l}}$ in \eqref{eq:R2l+2}. 
We arrive at  $\xl{L}=\left|\mathbf{X}-R_{2(L-3)}\right|$.
Thus for $L>3$, the neural net has \kink{}s at reflections of training data of up to order-$2(L-3)$. The network has \kink{}s at reflections of order $0$ for $L=2$ and order $1$ for $L=3$ \eqref{eq:LL-2R}, respectively.
\end{proof}

%\newpage

%\bibliography{references}
%\section*{References}

%%%%%%%%%%%%%%%%%%%%%%%%%%%%%%%%%%%%%%%%%%%%%%%%%%%%%%%%%%%%

\end{document}